\pdfoutput=1
\documentclass[jair,twoside,11pt,theapa]{article}
\usepackage{notjair, theapa, rawfonts}
\usepackage{latexsym}
\usepackage{amsmath,amssymb,enumerate,bbm,color,alltt}
\usepackage{url}
\usepackage{graphicx}
\usepackage{float}
\usepackage{stfloats}
\usepackage{subfig}
\usepackage{epstopdf}
\usepackage{longtable}
\usepackage{tabularx}
\usepackage{multirow}
\usepackage{booktabs}
\usepackage{subfig}
\usepackage{multicol}
\usepackage{mathtools}

\newcommand{\define}[1]{{\bf{#1}}}

\newcommand{\system}{$\partial$\textsc{ilp}}

\jairheading{61}{2018}{1-64}{02/17}{01/18}
\ShortHeadings{Learning Explanatory Rules from Noisy Data}
{Evans \& Grefenstette}
\firstpageno{1}

\begin{document}

\title{Learning Explanatory Rules from Noisy Data}

\author{\name Richard Evans \email richardevans@google.com \\
       \name Edward Grefenstette \email etg@google.com \\
       \addr DeepMind, London, UK}

\maketitle

\begin{abstract}
Artificial Neural Networks are powerful function approximators capable of modelling solutions to a wide variety of problems, both supervised and unsupervised. As their size and expressivity increases, so too does the variance of the model, yielding a nearly ubiquitous overfitting problem. Although mitigated by a variety of model regularisation methods, the common cure is to seek large amounts of training data---which is not necessarily easily obtained---that sufficiently approximates the data distribution of the domain we wish to test on. In contrast, logic programming methods such as Inductive Logic Programming offer an extremely data-efficient process by which models can be trained to reason on symbolic domains. However, these methods are unable to deal with the variety of domains neural networks can be applied to: they are not robust to noise in or mislabelling of inputs, and perhaps more importantly, cannot be applied to non-symbolic domains where the data is ambiguous, such as operating on raw pixels. In this paper, we propose a Differentiable Inductive Logic framework, which can not only solve tasks which traditional ILP systems are suited for, but shows a robustness to noise and error in the training data which ILP cannot cope with. Furthermore, as it is trained by backpropagation against a likelihood objective, it can be hybridised by connecting it with neural networks over ambiguous data in order to be applied to domains which ILP cannot address, while providing data efficiency and generalisation beyond what neural networks on their own can achieve.
\end{abstract}

\section{Introduction}
\label{sec:introduction}

Inductive Logic Programming (ILP) is a collection of techniques for constructing logic programs from examples.
Given a set of positive examples, and a set of negative examples, an ILP system constructs a logic program that entails all the positive examples but does not entail any of the negative examples. From a machine learning perspective, an ILP system can be interpreted as implementing a rule-based binary classifier over examples, mapping each example to an evaluation of its truth or falsehood according to the axioms provided to the system, alongside new rules inferred by the system during training. 

ILP has a number of appealing features.
First, the learned program is an explicit symbolic structure that can be inspected, understood\footnote{Human readability is a much touted feature of ILP systems, but when the learned programs become large, and include a number of invented auxiliary predicates, the resulting programs become less readable (see \citeR{besolddoes}). But even a complex machine-generated logic program will be easier to understand than a large tensor of floating point numbers.}, and verified. 
Second, ILP systems tend to be impressively data-efficient, able to generalise well from a small handful of examples.
The reason for this data-efficiency is that ILP imposes a strong language bias on the sorts of programs that can be learned: a short general program will be preferred to a program consisting of a large number of special-case ad-hoc rules that happen to cover the training data.
Third, ILP systems support continual and transfer learning. The program learned in one training session, being declarative and free of side-effects, can be copied and pasted into the knowledge base before the next training session, providing an economical way of storing learned knowledge.
  
The main disadvantage of traditional ILP systems is their inability to handle noisy, erroneous, or ambiguous data. If the positive or negative examples contain any mislabelled data, these systems will not be able to learn the intended rule. \citeA{de2008probabilistic} discuss this issue in depth, stressing the importance of building systems capable of applying relational learning to uncertain data.

A key strength of neural networks is that they are robust to noise and ambiguity.
One way to overcome the brittleness of traditional ILP systems is to reimplement them in a robust connectionist framework.
\citeA{garcez2015neural} argue strongly for the importance of integrating robust connectionist learning with symbolic relational learning.

Recently, a different approach to program induction has emerged from the deep learning community \cite{ntm,reed2015neural,neelakantan2015neural,kaiser,andrychowicz2016learning,graves2016hybrid}.
These neural network-based systems do not construct an explicit symbolic representation of a program.
Instead, they learn an implicit procedure (distributed in the weights of the net) that produces the intended results.
These approaches take a relatively low-level model of computation\footnote{Low-level models include a Turing machine \cite{ntm,graves2016hybrid}, a Forth virtual machine \cite{riedel2016programming}, a cellular automaton \cite{kaiser}, and a pushdown automaton \cite{sun1998neural,grefenstette2015learning,joulin2015inferring}.}---a model that is much ``closer to the metal'' than the Horn clauses used in ILP---and produce a differentiable implementation of that low-level model.
The implicit procedure that is learned is a way of operating within that low-level model of computation (by moving the tape head, reading and writing in the case of differentiable Turing machines; by pushing and popping in the case of differentiable pushdown automata).

There are two appealing features of this differentiable approach to program induction. 
First, these systems are robust to noise.
Unlike ILP, a neural system will tolerate some bad (mis-labeled) data.
Second, a neural program induction system can be provided with fuzzy or ambiguous data (from a camera, for example).
Unlike traditional ILP systems (which have to be fed crisp, symbolic input), a differentiable induction system can start with raw, un-preprocessed pixel input.

However, the neural approaches to program induction have two disadvantages when compared to ILP. 
First, the implicit procedure learned by a neural network is not inspectable or human-readable.
It is notoriously hard to understand what it has learned, or to what extent it has generalised beyond the training data.
Second, the performance of these systems tails off sharply when the test data are significantly larger than the training data:
if we train the neural system to add numbers of length 10, they may also be successful when tested on numbers of length 20. 
But if we test them on numbers of length 100, the performance deteriorates \cite{kaiser,reed2015neural}.
General-purpose neural architectures, being universal function approximators, produce solutions with high variance. There is an ever-present danger of over-fitting\footnote{With extra supervision, over-fitting can be avoided, to an extent. \citeR{reed2015neural} use a much richer training signal. Instead of trying to learn rules from mere input output pairs, they learn from explicit traces of desired computations. In their approach, a training instance is an input plus a fine-grained specification of the desired computational operations. For example, when learning addition, a training instance would be the two inputs $x$ and $y$ that we are expected to add, plus a detailed list of all the low-level operations involved in adding $x$ and $y$. With this additional signal, it is possible to learn programs that generalise to larger training instances.} .

This paper proposes a system that addresses the limits of connectionist systems and ILP systems, and attempts to combine the strengths of both.
\define{Differentiable Inductive Logic Programming} ($\partial$ILP) is a reimplementation of ILP in an an end-to-end differentiable architecture.
It attempts to combine the advantages of ILP with the advantages of the neural network-based systems: a data-efficient induction system that can learn explicit human-readable symbolic rules, that is robust to noisy and ambiguous data, and that does not deteriorate when applied to unseen test data.
The central component of this system is a differentiable implementation of deduction through forward chaining on definite clauses.
We reinterpret the ILP task as a binary classification problem, and we minimise cross-entropy loss with regard to ground-truth boolean labels during training.

Our \system{} system is able to solve moderately complex tasks requiring recursion and predicate invention.
For example, it is able to learn ``Fizz-Buzz'' using multiple invented predicates (see Section \ref{subsubsec:fizz-buzz}).
Unlike the MLP described by Grus\footnote{See the darkly humorous \url{http://joelgrus.com/2016/05/23/fizz-buzz-in-tensorflow/}.}, our learned program generalises robustly to test data outside the range of training examples.

Unlike symbolic ILP systems, \system{} is robust to mislabelled data. 
It is able to achieve reasonable performance with up to 20\% of mislabelled training data (see Section \ref{sub:mislabelledexp}).
Unlike symbolic ILP systems, \system{} is also able to handle ambiguous or fuzzy data. 
We tested \system{} by connecting it to a convolutional net trained on MNIST digits, and it was still able to learn effectively (see Section \ref{sub:ambigexp}).

The major limitation of our \system{} system is that it requires significant memory resources. 
This limits the range of benchmark problems on which our system has been tested\footnote{The memory requirements require us to limit ourselves to predicates of arity at most two.}.
We discuss this further in the introduction to Section \ref{sub:ilpbenchmark} and in Section \ref{symbolic-results}, and offer an analysis in Appendix \ref{space-requirements}.

The structure of the paper is as follows. We begin, in Section~\ref{sec:background}, by giving an overview of logic programming as a field, and of the collection of learning methods known as Inductive Logic Programming. In Section~\ref{sec:satisfiability}, we re-cast learning under ILP as a satisfiability problem, and use this formalisation of the problem as the basis for introducing, in Section~\ref{sec:continuous}, a differentiable form of ILP where continuous representations of rules are learned through backpropagation against a likelihood objective. In Section~\ref{sec:experiments}, we evaluate our system against a variety of standard ILP tasks, we measure its robustness to noise by evaluating its performance under conditions where consistent errors exist in the data, and finally compare it to neural network baselines in tasks where logic programs are learned over ambiguous data such as raw pixels. We complete the paper by reviewing and contrasting with related work in Section~\ref{sec:related} before offering our conclusions regarding the framework proposed here and its empirical validation.

\section{Background}
\label{sec:background}

We first review Logic Programming and Inductive Logic Programming (ILP), before focusing on one particular approach to ILP, which turns the induction problem into a satisfiability problem. 

\subsection{Logic Programming}
\label{sub:logicprogramming}

Logic programming is a family of programming languages in which the central component is not a command, or a function, but an if-then rule. These rules are also known as clauses.

A \define{definite clause}\footnote{In this paper, we restrict ourselves to logic programs composed of definite clauses. We do not consider clauses containing negation-as-failure.}  is a rule of the form
\[
\alpha \leftarrow \alpha_1, ..., \alpha_m
\]
composed of a \define{head} atom $\alpha$, and a \define{body} $\alpha_1, ..., \alpha_m$, where $m \geq 0$.
These rules are read from right to left: if all of the atoms on the right are true, then the atom on the left is also true.
If the body is empty, if $m = 0$, we abbreviate $\alpha \leftarrow$ to just $\alpha$.

An \define{atom} $\alpha$ is a tuple $p(t_1, ..., t_n)$, where $p$ is a $n$-ary predicate and $t_1, ..., t_n$ are terms, either variables or constants.
An atom is \define{ground} if it contains no variables.
The set of all atoms, ground and unground, is denoted by $\mathcal{A}$.
The set of ground atoms, also known as the Herbrand base, is $\mathcal{G}$.

In first-order logic, terms can also be constructed from function symbols.
So, for example, if $c$ is a constant, and $f$ is a one place function, then $f(c), f(f(c)), f(f(f(c))), ...$ are all terms.
In this paper, we impose the following restriction, found in systems like \define{Datalog}, on clauses: the only terms allowed are constants and variables, while function symbols are disallowed\footnote{The reason for restricting ourselves to definite Datalog clauses is that this language is decidable and forward chaining inference (described below) always terminates.}.

For example, the following program defines the $\mathsf{connected}$ relation as the transitive closure of the $\mathsf{edge}$ relation:
\begin{eqnarray*}
\mathsf{connected}(X, Y) & \leftarrow & \mathsf{edge}(X, Y) \\
\mathsf{connected}(X, Y) & \leftarrow & \mathsf{edge}(X, Z), \mathsf{connected}(Z, Y)
\end{eqnarray*}
We follow the logic programming convention of using upper case to denote variables and lower case for constants and predicates.

A \define{ground rule} is a clause in which all variables have been substituted by constants.
For example: 
\[
\mathsf{connected}(a, b) \leftarrow \mathsf{edge}(a, b)
\]
is a ground rule generated by applying the substitution $\theta = \{a/X, b/Y\}$ to the clause:
\[
\mathsf{connected}(X, Y) \leftarrow \mathsf{edge}(X, Y)
\]

The \define{consequences} of a set $R$ of clauses is computed by repeatedly applying the rules in $R$ until no more consequences can be derived.
More formally, define $cn_{R}(X)$ as the immediate consequences of rules $R$ applied to ground atoms $X$:
\[
cn_{R}(X) = X \cup \{\gamma \; | \; \gamma \leftarrow \gamma_1, ..., \gamma_m \in \mathsf{ground}(R), \bigwedge_{i=1}^m \gamma_i \in X \}
\]
where $\mathsf{ground}(R)$ are the ground rules of $R$.
In other words, ground atom $\gamma$ is in $cn_{R}(X)$ if $\gamma$ is in $X$ or there exists a ground rule $\gamma \leftarrow \gamma_1, ..., \gamma_m \in \mathsf{ground}(R)$ such that each ground atom $\gamma_1, ..., \gamma_m$ is in $X$.

Alternatively, we can define $cn_{R}(X)$ using substitutions:
\[
cn_{R}(X) = X \cup \{\alpha [\theta] \; | \; \alpha \leftarrow \alpha_1, ..., \alpha_m \in R, \bigwedge_{i=1}^m \alpha_i [\theta] \in X \}
\]
where $\alpha [\theta]$ is the application of substitution $\theta$ to atom $\alpha$.

Now define a series of consequences $C_{R, 0}, C_{R, 1}, C_{R, 2}, ...$
\[C_{R, 0} = \{\} \qquad C_{R, i+1} = cn_{R}(C_{R, i})\]
Now define the consequences after $T$ time-steps as the union of this series:
\[
\mathsf{con}(R) = \bigcup_{i \geq 0}^T C_{R, i}
\]
Consider, for example, the program $R$:
\begin{center}
\begin{tabular}{lcl}
$\mathsf{edge}(a, b)$ & $\quad$ & $\mathsf{connected}(X, Y)  \leftarrow  \mathsf{edge}(X, Y)$ \\
$\mathsf{edge}(b, c)$ & $\quad$ & $\mathsf{connected}(X, Y)  \leftarrow  \mathsf{edge}(X, Z), \mathsf{connected}(Z, Y)$\\
$\mathsf{edge}(c, a)$ & $\quad$ & \\
\end{tabular}
\end{center}
The consequences are:
\begin{eqnarray*}
C_{R, 1} & = & \{ \mathsf{edge}(a, b), \mathsf{edge}(b, c), \mathsf{edge}(c, a) \} \\
C_{R, 2} & = & C_{R, 1} \cup \{ \mathsf{connected}(a, b), \mathsf{connected}(b, c), \mathsf{connected}(c, a) \} \\
C_{R, 3} & = & C_{R, 2} \cup \{ \mathsf{connected}(a, c), \mathsf{connected}(b, a), \mathsf{connected}(c, b) \} \\
C_{R, 4} & = & C_{R, 3} \cup \{ \mathsf{connected}(a, a), \mathsf{connected}(b, b), \mathsf{connected}(c, c) \} \\
C_{R, 5} & = & C_{R, 4} = \mathsf{con}(R)
\end{eqnarray*}

Given a set $R$ of clauses, and a ground atom $\gamma$, we say $R$ entails $\gamma$, written $R \models \gamma$, if every model satisfying $R$ also satisfies $\gamma$.
To test if $R \models \gamma$, we check if $\gamma \in \mathsf{con}(R)$.
This technique is called \define{forward chaining}. 

An alternative approach is \define{backward chaining}. 
To test if $R \models \gamma$, we work backwards from $\gamma$, looking for rules in $R$ whose head unifies with $\gamma$.
For each such rule $\alpha \leftarrow \alpha_1, ..., \alpha_m$, where $\alpha [\theta] = \gamma$, we create a sub-goal to prove the body $\alpha_1 [\theta], ..., \alpha_m [\theta]$.
This procedure constructs a search tree, where nodes are lists of propositions to be proved, in left-to-right order, and edges are applications of rules with substitutions. 
The root of the tree is the node containing $\gamma$, and search terminates when we find a node with no atoms remaining to be proved.

A distinction we will need later is the difference between intensional and extensional predicates.
An \define{extensional} predicate is a predicate that is wholly defined by a set of ground atoms. 
In our example above, $\mathsf{edge}$ is an extensional predicate defined by the set of atoms:
\[
\{\mathsf{edge}(a, b), \mathsf{edge}(b, c), \mathsf{edge}(c, a)\}
\]
An \define{intensional} predicate is defined by a set of clauses. 
In our example, $\mathsf{connected}$ is an intensional predicate defined by the clauses:
\begin{eqnarray*}
\mathsf{connected}(X, Y) & \leftarrow & \mathsf{edge}(X, Y) \\
\mathsf{connected}(X, Y) & \leftarrow & \mathsf{edge}(X, Z), \mathsf{connected}(Z, Y)
\end{eqnarray*}

\subsection{Inductive Logic Programming (ILP)}
\label{sub:ilp}
An \define{ILP problem} is a tuple $(\mathcal{B}, \mathcal{P}, \mathcal{N})$ of ground atoms, where:
\begin{itemize}
\item
$\mathcal{B}$ is a set of background assumptions, a set of ground atoms\footnote{We assume that $\mathcal{B}$ is a set of ground atoms, but not all ILP systems make this restriction. In some ILP systems, the background assumptions are clauses, not atoms.}.
\item
$\mathcal{P}$ is a set of positive instances - examples taken from the extension of the target predicate to be learned
\item
$\mathcal{N}$ is a set of negative instances - examples taken outside the extension of the target predicate
\end{itemize}
Consider, for example, the task of learning which natural numbers are \emph{even}.
We are given a minimal description of the natural numbers: 
\[
\mathcal{B} = \{zero(0), succ(0, 1), succ(1, 2), succ(2, 3), ...\}
\]
Here, $zero(X)$ is the unary predicate true of $X$ if $X = 0$, and $succ$ is the successor relation.
The positive and negative examples of the $even$ predicate are:
\begin{eqnarray*}
\mathcal{P} & = & \{\mathsf{even}(0), \mathsf{even}(2), \mathsf{even}(4), \mathsf{even}(6), ...\} \\
\mathcal{N} & = & \{\mathsf{even}(1), \mathsf{even}(3), \mathsf{even}(5), \mathsf{even}(7), ...\}
\end{eqnarray*}
The aim of ILP is to construct a logic program that explains the positive instances and rejects the negative instances.

Given an ILP problem $(\mathcal{B}, \mathcal{P}, \mathcal{N})$, a \define{solution} is a set $R$ of definite clauses such that\footnote{This is a specific definition of ILP in terms of entailment. For a more general definition of the ILP problem, that contains this formulation as a special case, see the work of \citeA{muggleton1994inductive,de2008probabilistic,de2008probabilistic2}.}:
\begin{itemize}
\item
$\mathcal{B}, R \models \gamma$ for all $\gamma \in \mathcal{P}$
\item
$\mathcal{B}, R \nvDash \gamma$ for all $\gamma \in \mathcal{N}$
\end{itemize}
Induction is finding a set of rules $R$ such that, when they are applied deductively to the background assumptions $\mathcal{B}$, they produce the desired conclusions. Namely: positive examples are entailed, while negative examples are not.

In the example above, one solution is the set $R$:
\begin{eqnarray*}
even(X) & \leftarrow & zero(X) \\
even(X) & \leftarrow & even(Y), succ2(Y, X) \\
succ2(X, Y) & \leftarrow & succ(X, Z), succ(Z, Y)
\end{eqnarray*}
Note that this simple toy problem is not entirely trivial. The solution requires both recursion and \define{predicate invention}: an auxiliary synthesised predicate $succ2$.

We first describe how the ILP problem can be transformed into a satisfiability problem (Section~\ref{sec:satisfiability}), and then provide the main contribution of the paper: a differentiable implementation of the satisfiability solving process (Section~\ref{sec:continuous}).

\section{ILP as a Satisfiability Problem}

\label{sec:satisfiability}

There are, broadly speaking, two families of approaches to ILP.
The bottom-up approaches\footnote{For example, Progol \cite{muggleton1995inverse}.} start by examining features of the examples, extract specific clauses from those examples, and then generalise from those specific clauses.
The top-down approaches\footnote{For example, TopLog \cite{muggleton2008toplog}, TAL \cite{corapi2010inductive}, Metagol \cite{muggleton2014meta,muggleton2015meta,cropper2015meta}.} use generate-and-test: 
they generate clauses from a language definition, and test the generated programs against the positive and negative examples.

Amongst the top-down approaches, one particular method is to transform the problem of searching for clauses into a satisfiability problem.
Amongst these induction-via-satisfiability approaches, some\footnote{See, for example, the work by \citeA{chikarainductive}.} use the Boolean flags to indicate which branches of the search space (defined by the language grammar) to explore.
Others\footnote{See, for example, the work by \citeA{corapi2010inductive}. Note that they are working with Answer-Set Programming, but under the hood, this is propositionalised and compiled into a SAT problem.} generate a large set of clauses according to a program template, and use the Boolean flags to determine which clauses are on and off.

In this paper, we also take a top-down, generate-and-test approach, in which clauses are generated from a language template. 
We assign each generated clause a Boolean flag indicating whether it is on or off. 
Now the induction problem becomes a satisfiability problem: choose an assignment to the Boolean flags such that the turned-on clauses together with the background facts together entail the positive examples and do not entail the negative examples.
Our approach is an extension to this established approach to ILP: our contribution is to provide, via a continuous relaxation of satisfiability, a \emph{differentiable implementation} of this architecture. This allows us to apply gradient descent to learn which clauses to turn on and off, even in the presence of noisy or ambiguous data.

We use the rest of this section (together with Appendix \ref{sat}) to explain ILP-as-satisfiability in detail.



\subsection{Basic Concepts}
\label{basic-concepts}

A \define{language frame} $\mathcal{L}$ is a tuple \[(target, P_e, arity_e, C)\] where:
\begin{itemize}
\item
$target$ is the target predicate, the intensional predicate we are trying to learn
\item
$P_e$ is a set of extensional predicates
\item
$arity_e$ is a map $P_e \cup \{target\} \rightarrow \mathbb{N}$ specifying the arity of each predicate
\item
$C$ is a set of constants
\end{itemize}
An \define{ILP problem} is a tuple $(\mathcal{L}, \mathcal{B}, \mathcal{P}, \mathcal{N})$ where
\begin{itemize}
\item
$\mathcal{L}$ is a language frame
\item
$\mathcal{B}$ is a set of background assumptions, ground atoms formed from the predicates in $P_e$ and the constants in $C$
\item
$\mathcal{P}$ is a set of positive examples, ground atoms formed from the $target$ predicate and the constants in $C$
\item
$\mathcal{N}$ is a set of negative examples, ground atoms formed from the $target$ predicate and the constants in $C$
\end{itemize}
Here, $\mathcal{P}$ are ground atoms where the target predicate holds. For example, in the case of $even$ on natural numbers, $\mathcal{P}$ might be $\{even(0), even(2), even(4), even(6)\}$. The set $\mathcal{N}$ contains ground atoms where the target predicate does not hold, e.g..~$\{even(1), even(3), even(5)\}$.

A \define{rule template} $\tau$ describes a range of clauses that can be generated. It is a pair $(v, int)$ where:
\begin{itemize}
\item
$v \in \mathbb{N}$ specifies the number of existentially quantified variables allowed in the clause
\item
$int \in \{0,1\}$ specifies whether the atoms in the body of the clause can use intensional predicates ($int=1$) or only extensional predicates ($int=0$)
\end{itemize}
A \define{program template} $\Pi$ describes a range of programs that can be generated. It is a tuple $(P_a, arity_a, rules, T)$ where:
\begin{itemize}
\item
$P_a$ is a set of auxiliary (intensional) predicates; these are the additional invented predicates used to help define the target predicate
\item
$arity_a$ is a map $P_a \rightarrow \mathbb{N}$ specifying the arity of each auxiliary predicate
\item
$rules$ is a map from each intensional predicate $p$ to a pair of rule templates $(\tau_p^1, \tau_p^2)$
\item
$T \in \mathbb{N}$ specifies the max number of steps of forward chaining inference
\end{itemize}
Note that $rules$ defines each intensional predicate by a \emph{pair} of rule templates.
In our system, we insist, without loss of generality\footnote{If a predicate $p$ is defined by three clauses $p(X,Y) \leftarrow \beta_1$, $p(X,Y)  \leftarrow \beta_2$, and $p(X,Y) \leftarrow \beta_3$, create a new auxiliary predicate $q$ and replace the above with four clauses: (i) $p(X,Y) \leftarrow \beta_1$; (ii) $p(X,Y) \leftarrow q(X,Y)$; (iii) $q(X,Y) \leftarrow \beta_2$; (iv) $q(X,Y) \leftarrow \beta_3$. We have transformed a program in which one predicate is defined by three clauses into a program in which two predicates are defined by two clauses each.\label{two-clauses}}  that each predicate can be defined by exactly two clauses. 

Our program template is a tuple of hyperparameters constraining the range of programs that are used to solve the ILP problem.
This template plays the same role as a collection of mode declarations (see Appendix \ref{corapi_comparison}) or a set of metarules in Metagol (see Appendix \ref{metagol_comparison}).

Assume, for the moment, that the program template is given to us as part of the ILP problem specification.
Later, in Section~\ref{iterative-deepening}, we shall discuss how to search through the space of program templates using iterative deepening.

We can combine the extensional predicates from the language-frame \[\mathcal{L} = (target, P_e, arity_e, C)\] with the intensional predicates from the program template \[\Pi = (P_a, arity_a, rules, T)\] into a \define{language} $(P_e, P_i, arity, C)$ where 
\begin{itemize}
\item
$P_i = P_a \cup \{target\}$
\item
$arity = arity_e \cup arity_a$
\end{itemize}
Note that the $target$ predicate is always one of the intensional predicates $P_i$.

Let $P$ be the complete set of predicates:
\[
P = P_e \cup P_i
\]
A language determines the set $G$ of all ground atoms.
If we restrict ourselves to nullary, unary, and dyadic predicates, then the set of ground atoms is:
\begin{eqnarray*}
G =\{\gamma_i\}_{i=1}^n & = & \{ p() \; | \; p \in P, \; \mathsf{arity}(p) = 0\} \cup \\
& & \{ p(k) \; | \; p \in P, \; \mathsf{arity}(p) = 1, \; k \in C\} \cup \\
& & \{ p(k_1,k_2) \; | \; p \in P, \; \mathsf{arity}(p) = 2, \; k_1,k_2 \in C \} \cup  \\
& & \{ \perp \}
\end{eqnarray*}
Note that $G$ includes the falsum atom $\perp$, the atom that is always false.

\subsection{Generating Clauses}
\label{generating-clauses}

For each rule template $\tau$, we can generate a set $cl(\tau)$ of clauses that satisfy the template.
To keep the set of generated clauses manageable, we make a number of restrictions.
First, the only clauses we consider are those composed of atoms involving free variables.
\emph{We do not allow any constants in any of our clauses}.
This may seem, initially, to be a severe restriction---but recall that our language has extensional predicates as well as intensional predicates.
If we need a predicate whose meaning depends on particular constants, then we treat it as an extensional predicate, rather than an intensional predicate.
For example, the predicate $zero/1$, which appears in the arithmetic examples later, is treated as an extensional predicate.
We do not treat $zero$ as an intensional predicate defined by a single rule with an empty body:
\[
zero(0) \leftarrow
\]
Rather, we treat $zero$ as an extensional predicate defined by the single background atom:
\[
zero(0)
\]
The second restriction on generated clauses is that we only allow predicates of arity 0, 1, or 2. We do not currently support ternary predicates or higher\footnote{This is because of practical space constraints.
See Appendix~\ref{space-requirements}.}.

Third, we insist that all clauses have exactly two atoms in the body.
This restriction can also be made without loss of generality. For any logic program involving clauses with more than two atoms in the body of some clause, there is an equivalent logic program (with additional intensional predicates used as auxiliary functions) with exactly two atoms in the body of each clause\footnote{This is a separate condition from the restriction above in Section \ref{basic-concepts} that each predicate can be defined by exactly two clauses.}.

There are four additional restrictions on generated clauses. We rule out clauses that are (1) unsafe (a variable used in the head is not used in the body), (2) circular (the head atom appears in the body), (3) duplicated (equivalent to another clause with the body atoms permuted), and (4) those that do not respect the intensional flag $int$ (i.e.~those that contain an intensional predicate in the clause body, even though the $int$ flag was set to 0, i.e.~False).
We provide a worked example in Appendix \ref{sat}.

\subsection{Reducing Induction to Satisfiability}
\label{reducing-to-sat}

Given a program template $\Pi$, let $\tau_p^i$ be the $i$'th rule template for intensional predicate $p$, where $i \in \{1, 2\}$ indicates which of the two rule templates we are considering for $p$.
Let $C_p^{i, j}$ be the $j'th$ clause in $cl(\tau_p^i)$, the set of clauses generated for template $\tau_p^i$ .

To turn the induction problem into a satisfiability problem, we define a set $\Phi$ of Boolean variables (i.e.~atoms with nullary predicates) indicating which of the various clauses in $C_p^{i, j}$ are actually to be used in our program.
Now a SAT solver can be used to find a truth-assignment to the propositions in $\Phi$, and we can extract the induced rules from the subset of propositions in $\Phi$ that are set to True.
The technical details behind this approach are described in Appendix \ref{sat}.

\section{A Differentiable Implementation of ILP}
\label{sec:continuous}

In this section, we describe our core model: a continuous reimplementation of the ILP-as-satisfiability approach described above. 
The discrete operations are replaced by differentiable operators, so the ILP problem can be solved by minimising a loss using stochastic gradient descent. In this case, the loss is the cross entropy of the correct label with regard to the predictions of the system.

Instead of the discrete semantics in which ground atoms are mapped to $\{False, True\}$, we now use a continuous semantics\footnote{For a related approach, see the work of \citeA{logictensornetworks}, discussed in Section \ref{deep-logic} below.} which maps atoms to the real unit interval\footnote{The values in $[0,1]$ are interpreted as \emph{probabilities} rather than fuzzy ``degrees of truth''.} $[0,1]$.
Instead of using Boolean flags to choose a discrete subset of clauses, we now use continuous weights to determine a probability distribution over clauses.

This model, which we call \system{}, implements differentiable deduction over continuous values. The gradient of the loss with regard to the rule weights, which we use to minimise classification loss, implements a continuous form of induction.

\subsection{Valuations}
\label{subsec:valuations}

Given a set $G$ of $n$ ground atoms, a {\bf valuation} is a vector $[0, 1]^n$ mapping each ground atom $\gamma_i \in G$ to the real unit interval. 

Consider, for example, the language $L = (P_e, P_i, \mathsf{arity}, C)$, where
\[
P_e  =  \{ r / 2 \} \qquad
P_i  =  \{ p / 0, q / 1 \} \qquad
C =  \{a, b \}
\]
One possible valuation on the ground atoms $G$ of $L$ is
\begin{center}
\begin{tabular}{lclclcl}
$\perp \ \mapsto  0.0$ & $\quad$ &  $p() \mapsto 0.0$ & $\quad$ & $q(a) \mapsto 0.1$ & $\quad$ & $q(b) \mapsto 0.3$ \\
 $r(a, a) \mapsto 0.7$  & $\quad$ & $r(a, b) \mapsto 0.1$ & $\quad$ & $r(b, a) \mapsto 0.4$ & $\quad$ &  $r(b, b) \mapsto 0.2$
\end{tabular}
\end{center}
We insist that all valuations map $\perp$ to 0. (The reason for including the $\perp$ atom will become clear in Section~\ref{f-function}).

\subsection{Induction by Gradient Descent}
\label{subsec:differentiable-induction}

Given the sets $\mathcal{P}$ and $\mathcal{N}$ of positive and negative examples, we form a set $\Lambda$ of atom-label pairs:
\[
\Lambda = \{ (\gamma, 1) \; | \gamma \in \mathcal{P}\} \cup \{ (\gamma, 0) \; | \gamma \in \mathcal{N}\}
\]
Each pair $(\gamma, \lambda)$ indicates whether atom $\gamma$ is in $\mathcal{P}$ (when $\lambda=1$) or $\mathcal{N}$ (when $\lambda=0$). This can be thought of as a dataset, used to learn a binary classifier that maps atoms $\gamma$ to their truth or falsehood.

Now given an ILP problem $(\mathcal{L}, \mathcal{B}, \mathcal{P}, \mathcal{N})$, a program template $\Pi$ and a set of clause-weights $W$, we construct a differentiable model that implements the conditional probability of $\lambda$ for a ground atom $\alpha$: 
\[
p(\lambda \; | \; \alpha, W, \Pi, \mathcal{L}, \mathcal{B})
\]
We want our predicted label $p(\lambda \; | \; \alpha, W, \Pi, \mathcal{L}, \mathcal{B})$ to match the actual label $\lambda$ in the pair $(\alpha, \lambda)$ we sample from $\Lambda$.
We wish, in other words, to minimise the expected negative log likelihood when sampling uniformly $(\alpha, \lambda)$ pairs from $\Lambda$:
\begin{eqnarray*}
loss = - \mathop{\mathbb{E}}_{(\alpha, \lambda) \sim \Lambda} [ \lambda \cdot \log p(\lambda \; | \; \alpha, W, \Pi, \mathcal{L}, \mathcal{B}) + (1 - \lambda) \cdot \log (1 - p(\lambda \; | \; \alpha, W, \Pi, \mathcal{L}, \mathcal{B})) ] 
\end{eqnarray*}
To calculate the probability of the label $\lambda$ given the atom $\alpha$, we infer the consequences of applying the rules to the background facts (using $T$ steps of forward chaining). In Figure \ref{fig:architecture} below, these consequences are called the ``Conclusion Valuation''. Then, we extract $\lambda$ as the probability of $\alpha$ in this valuation.

The probability  $p(\lambda \; | \; \alpha, W, \Pi, \mathcal{L}, \mathcal{B})$ is defined as:
\[
p(\lambda \; | \; \alpha, W, \Pi, \mathcal{L}, \mathcal{B}) = f_{extract}(f_{infer}(f_{convert}(\mathcal{B}), f_{generate}(\Pi, \mathcal{L}), W, T), \alpha)
\]
Here, $p(\lambda \; | \; \alpha, W, \Pi, \mathcal{L}, \mathcal{B})$ is computed using four auxiliary functions: $f_{extract}$, $f_{infer}$, $f_{convert}$, and $f_{generate}$. (See Figure \ref{fig:architecture}).
$f_{extract}$ and $f_{infer}$ are differentiable operations, while $f_{convert}$, and $f_{generate}$ are non-differentiable.

The function $f_{extract} : [0,1]^n \times G \rightarrow [0,1]$ takes a valuation $\mathbf{x}$ and an atom $\gamma$ and extracts the value for that atom: 
\[
f_{extract}(\mathbf{x}, \gamma) = \mathbf{x}[\mathsf{index}(\gamma)]
\]
where $\mathsf{index} : G \rightarrow \mathbb{N}$ is a function that assigns each ground atom a unique integer index.
The function $f_{convert} : 2^G \rightarrow [0,1]^n$ takes a set of atoms and converts it into a valuation mapping the elements of $\mathcal{B}$ to 1 and all other elements of $G$ to 0:
\[
f_{convert}(\mathcal{B}) = \mathbf{y}
\quad \text{where} \quad 
\mathbf{y}[i] = \begin{cases} 
1 \; \text{if} \; \gamma_i \in \mathcal{B} \\
0 \; \text{otherwise}
\end{cases}
\]
and where $\gamma_i$ is the $i$'th ground atom in $G$ for $i = 1 .. n$.

The function $f_{generate}$ produces a set of clauses from a program template $\Pi$ and a language $\mathcal{L}$:
\[
f_{generate}(\Pi, \mathcal{L}) = \{ cl(\tau_p^i) \; | \; p \in P_i, i \in \{1, 2 \} \}
\]
This uses the $cl$ function defined in Section~\ref{generating-clauses} above.

The $f_{infer} : [0,1]^n \times C \times W \times \mathbb{N} \rightarrow [0,1]^n $ function is where all the heavy-lifting takes place. It performs $T$ steps\footnote{Recall that $T$ is a part of the program template $\Pi$.} of forward-chaining inference using the generated clauses, amalgamating the various conclusions together using the clause weights $W$.
It is described in detail below.

\begin{figure}
\centering
\includegraphics[scale=0.8]{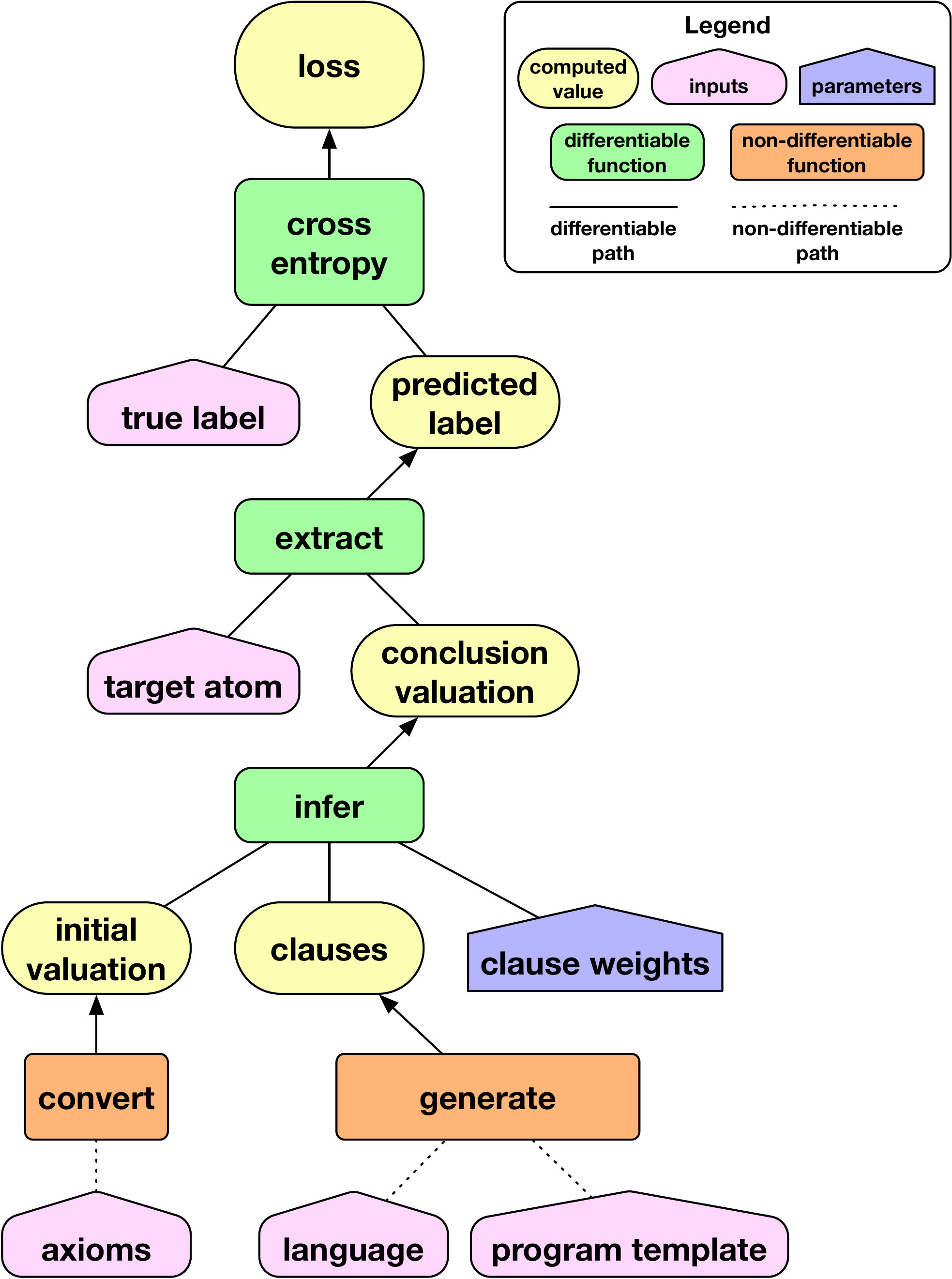}
\caption{The \system{} Architecture.}
\label{fig:architecture}
\end{figure}
Figure~\ref{fig:architecture} shows the architecture.
The inputs to the network, the elements that are fed in every training step, are the atom $\alpha$ and the corresponding label $\lambda$.
These are fed into the boxes labeled ``Sampled Target Atom'' and ``Sampled Label''. 
This input pair $(\alpha, \lambda)$ is sampled from the set $\Lambda$.
The conditional probability $p(\lambda \; | \; \alpha, W, \Pi, \mathcal{L}, \mathcal{B})$ is the value in the ``Predicted Label'' box.
The only trainable component is the set of clause weights $W$, shown in red.
The differentiable operations are shown in green, while the non-differentiable operations are shown in orange.
Note that, even though not all operations are differentiable, the operators between the loss and the clause weights $W$ are all differentiable, so we can compute $\frac{\partial{loss}}{\partial{W}}$, which in turn is used to update the clause weights by stochastic gradient descent (or any related optimisation method).

So far, we have described the high-level picture, but not the details.
In Section~\ref{weights}, we explain how the rule weights are represented.
In Section~\ref{inference}, we show how inference is performed over multiple time steps, by translating each clause $c$ into a function \[\mathcal{F}_c : [0,1]^n \rightarrow [0,1]^n\]
Finally, in Section~\ref{f-function}, we describe how the $\mathcal{F}_c$ functions are computed.

\subsection{Rule Weights}
\label{weights}

The \define{weights} $W$ are a set $\{\mathbf{W}_1, ..., \mathbf{W}_{|P_i|}\}$ of matrices, one matrix for each intensional predicate $p \in P_i$.
The matrix $\mathbf{W}_p$ for predicate $p$ is of shape $(|cl(\tau_p^1)|, |cl(\tau_p^2)|)$, where $|cl(\tau_p^1)|$ is the number of clauses generated by the first rule template $\tau_p^1$, and $|cl(\tau_p^2)|$ is the number of clauses generated by the second rule template $\tau_p^2$.
Note that the various $\mathbf{W_p}$ matrices are typically not of the same size, as the different rule templates produce different sized sets of clauses.

The weight $\mathbf{W}_p[j, k]$ represents how strongly the system believes that the pair of clauses $(C_p^{1,j}, C_p^{2,k})$ is the right way to define the intensional predicate $p$.
(Recall from Section~\ref{reducing-to-sat} that $C_p^{i,j}$ is the $j$'th clause of the $i$'th rule template $\tau_p^i$ for intensional predicate $p$. Here, as each predicate is defined by exactly two clauses, $i \in \{1, 2\}$. Recall from Section~\ref{basic-concepts} that we assume that each intensional predicate is defined by exactly two clauses).
The weight matrix $\mathbf{W}_p \in \mathbb{R}^{|cl(\tau_p^1)| \times |cl(\tau_p^2)|}$ is a matrix of real numbers.
We transform it into a probability distribution $\mathbf{W}^*_p \in [0,1]^{|cl(\tau_p^1)| \times |cl(\tau_p^2)|}$ using $\mathsf{softmax}$:
\[
\mathbf{W}^*_p[j, k] = \frac{e^{\mathbf{W}_p[j,k]}}{\sum_{j',k'} e^{\mathbf{W}_p[j',k']}}
\]
Here, $\mathbf{W}^*_p[j, k]$ represents the probability that the pair of clauses $(C_p^{1,j}, C_p^{2,k})$ is the right way to define the intensional predicate $p$.

Using matrices to store the weights of each \emph{pair} of clauses requires a lot of memory.
See Appendix~\ref{space-requirements}.
An alternative, less memory-hungry approach would be to have a vector of weights for every set of clauses generated by every individual rule template.
Unfortunately, this alternative is much less effective at the ILP tasks. 
See Appendix~\ref{simpler-smaller-model} for a fuller discussion.

\subsection{Inference}
\label{inference}

The central idea behind our differentiable implementation of inference is that each clause $c$ induces a function $\mathcal{F}_c : [0,1]^n \rightarrow [0,1]^n$ on valuations.
Consider, for example, the clause $c$:
\[
p(X) \leftarrow q(X)
\]
Table~\ref{clause-functions} shows the results of applying the corresponding function $\mathcal{F}_c$ to two valuations on the set $G = \{p(a), p(b), q(a), q(b), \perp\}$ of ground atoms.
\begin{table}
\centering
\begin{tabular}{lllll}
\toprule
$G$ & $\mathbf{a}_0$ & $\mathcal{F}_{c}(\mathbf{a_0})$ & $\mathbf{a}_1$ & $\mathcal{F}_{c}(\mathbf{a}_1)$ \\ 
\midrule
$p(a)$ & $0.0$ & $0.1$ & $0.2$  & $0.7$ \\
$p(b)$ & $0.0$ & $0.3$ & $0.9$  & $0.4$ \\
$q(a)$ & $0.1$ & $0.0$ & $0.7$  & $0.0$ \\
$q(b)$ & $0.3$ & $0.0$ & $0.4$  & $0.0$ \\
$\perp$ & $0.0$ & $0.0$ & $0.0$  & $0.0$ \\ \bottomrule
\end{tabular}
\caption{Applying $c = p(X) \leftarrow q(X)$, treated as a function $\mathcal{F}_c$, to valuations $\mathbf{a}_0$ and $\mathbf{a}_1$}
\label{clause-functions}
\end{table}

The details of how the $F_c$ functions are generated is deferred to Section~\ref{f-function} below.
The important point now is that we can automatically generate, from each clause $c$, a differentiable function $\mathcal{F}_c$ on valuations that implements a single step of forward chaining inference using $c$.

Recall that $C$ is an indexed set of generated clauses, where $C_p^{i,j}$ is the $j$'th clause of the $i$'th rule template $\tau_p^i$ for intensional predicate $p$.
Define a corresponding indexed set of functions $\mathcal{F}$ where $\mathcal{F}_p^{i,j}$ is the valuation function corresponding to the clause $C_p^{i,j}$.

Now we define another indexed set of functions $\mathcal{G}_p^{j,k}$ that combines the application of two functions $\mathcal{F}_p^{1,j}$ and $\mathcal{F}_p^{2,k}$.
Recall that each intensional predicate $p$ is defined by two clauses generated from two rule templates $\tau_p^1$ and $\tau_p^2$.
Now $\mathcal{G}_p^{j, k}$ is the result of applying both clauses $C_p^{1,j}$ and $C_p^{2,k}$ and taking the element-wise max:
\[
\mathcal{G}_p^{j, k}(\mathbf{a})  = \mathbf{x} \quad \text{where} \quad
\mathbf{x}[i] =  \max{(\mathcal{F}_p^{1,j}(\mathbf{a})[i], \mathcal{F}_p^{2,k}(\mathbf{a})[i])}
\]
Next, we will define a series of valuations of the form $\mathbf{a}_t$.
A valuation  $\mathbf{a}_t$ represents our conclusions after $t$ time-steps of inference.

The initial value $\mathbf{a}_0$ when $t=0$ is based on our initial set $\mathcal{B} \subseteq G$ of background axioms:
\[
\mathbf{a}_0[x] = \begin{cases}
1 \; \mbox{if} \; \gamma_x \in \mathcal{B} \\
0 \; \mbox{otherwise}
\end{cases}
\]
We now define $\mathbf{c}_t^{p,j,k}$:
\[
\mathbf{c}_t^{p,j,k} = G_p^{j,k}(\mathbf{a}_t)
\]
Intuitively, $\mathbf{c}_t^{p,j,k}$ is the result of applying one step of forward chaining inference to $\mathbf{a}_t$ using clauses $C_p^{1,j}$ and $C_p^{2,k}$.
Note that this only has non-zero values for one particular predicate: the $p'th$ intensional predicate.

We can now define the weighted average of the $\mathbf{c}_t^{p,j,k}$, using the softmax of the weights:
\[
\mathbf{b}_t^{p} = \sum_{j, k} \mathbf{c}_t^{p,j,k} \cdot \frac{e^{\mathbf{W}_p[j,k]}}{\sum_{j',k'} e^{\mathbf{W}_p[j',k']}}
\]
Intuitively, $\mathbf{b}_t^{p}$ is the result of applying all the possible pairs of clauses that can jointly define predicate $p$, and weighting the results by the weights $\mathbf{W}_p$.
Note that $\mathbf{b}_t^{p}$ is also zero everywhere except for the $p$'th intensional predicate.

From this, we define the successor $\mathbf{a}_{t+1}$ of $\mathbf{a}_{t}$:
\[
\mathbf{a}_{t+1} = f_{amalgamate}(\mathbf{a}_t, \sum_{p \in P_i} \mathbf{b}_t^{p})
\]
The successor depends on the previous valuation $\mathbf{a}_t$ and a weighted mixture of the clauses defining the other intensional predicates. Note that the valuations $\mathbf{b}_t^{p}$ are disjoint for different $p$, so we can simply sum these valuations.

When amalgamating the previous valuation, $\mathbf{a}_t$, with the single-step consequences, $\sum_{p \in P_i} \mathbf{b}_t^{p}$, there are various functions we can use for $f_{amalgamate}$.
First we considered:
\[
f_{amalgamate}(\mathbf{x},\mathbf{y}) = \max(\mathbf{x},\mathbf{y})
\]
(Note this is element-wise max over the two valuation vectors).
But the use of $\max$ here adversely affected gradient flow. 
The definition of $f_{amalgamate}$ we actually use is the \define{probabilistic sum}:
\[
f_{amalgamate}(\mathbf{x},\mathbf{y}) = \mathbf{x} + \mathbf{y} - \mathbf{x} \cdot \mathbf{y}
\]
This keeps valuations within the real unit interval $[0,1]$ while allowing gradients to flow through both $\mathbf{x}$ and $\mathbf{y}$.
The two alternative ways of computing $f_{amalgamate}$ are compared in Table~\ref{symbolic-results-table}.

\subsection{Computing the $\mathcal{F}_c$ Functions}
\label{f-function}

We now explain the details of how the various $\mathcal{F}_c$ functions are computed.

Each $\mathcal{F}_c$ function can be computed as follows.
Let $X_c = \{x_k\}_{k=1}^n$ be a set of sets of pairs of indices of ground atoms for clause $c$.
Each $x_k$ contains all the pairs  of indices of atoms that justify atom $\gamma_k$ according to the current clause $c$:
\[
x_k = \{ (a, b) \; | \; \mathsf{satisfies}_c(\gamma_a, \gamma_b) \land \mathsf{head}_c(\gamma_a, \gamma_b) = \gamma_k \}
\]
Note that we can restrict ourselves to pairs only (and don't have to worry about triples, etc) because we are restricting ourselves to rules with two atoms in the body.

Here, $\mathsf{satisfies}_c(\gamma_1, \gamma_2)$ if the pair of ground atoms $(\gamma_1, \gamma_2)$ satisfies the body of clause $c$.
If $c = \alpha \leftarrow \alpha_1, \alpha_2$, then $\mathsf{satisfies}_c(\gamma_1, \gamma_2)$ is true if there is a substitution $\theta$ such that $\alpha_1 [\theta] = \gamma_1$ and $\alpha_2 [\theta ] = \gamma_2$.

Also, $\mathsf{head}_c(\gamma_1, \gamma_2)$ is the head atom produced when applying clause $c$ to the pair of atoms $(\gamma_1, \gamma_2)$.
If $c = \alpha \leftarrow \alpha_1, \alpha_2$ and $\alpha_1 [\theta] = \gamma_1$ and $\alpha_2 [\theta ] = \gamma_2$ then
\[
\mathsf{head}_c(\gamma_1, \gamma_2) = \alpha [\theta]
\]
For example, suppose $P = \{p, q, r\}$ and $C = \{a, b\}$.
Then our ground atoms $G$ are:
\begin{center}
\begin{tabular}{llllllllll}
\toprule
$k$ & 0 & 1 & 2 & 3 & 4 & 5 & 6 & 7 & 8 \\
$\gamma_k$  & $\perp$ & $p(a, a)$ & $p(a, b)$ & $p(b, a)$ & $p(b, b)$ & $q(a, a)$ & $q(a, b)$  & $q(b, a)$ & $q(b, b)$\\
\midrule
$k$ & 9 & 10 & 11 & 12 & & & & &\\
$\gamma_k$ & $r(a, a)$ & $r(a, b)$ & $r(b, a)$ & $r(b, b)$ & & & & &\\
\bottomrule
\end{tabular}
\end{center}

\noindent
Suppose clause $c$ is:
\[
r(X, Y) \leftarrow p(X, Z), q(Z, Y)
\]
Then $X_c = \{x_k\}_{k=1}^n$ is:
\begin{multicols}{3}
\small
\begin{center}
\begin{tabular}{llllll}
\toprule
$k$ & $\gamma_k$ & $x_k$ \\
\midrule
0 & $\perp$ & \{\}\\ 
1 & $p(a, a)$ & \{\}\\
2 & $p(a, b)$ & \{\}\\
3 & $p(b, a)$ & \{\}\\
4 & $p(b, b)$ & \{\}\\
\bottomrule
\end{tabular}
\end{center}
\columnbreak

\begin{tabular}{llllll}
\toprule
$k$ & $\gamma_k$ & $x_k$ \\
\midrule
5 & $q(a, a)$ & \{\}\\
6 & $q(a, b)$ & \{\}\\
7 & $q(b, a)$ & \{\} \\ 
8 & $q(b, b)$ & \{\} \\
\bottomrule
\end{tabular}

\columnbreak
\begin{center}
\begin{tabular}{lll}
\toprule
$k$ & $\gamma_k$ & $x_k$ \\
\midrule
9 & $r(a, a)$ & \{(1,5), (2, 7)\} \\ 
10 & $r(a, b)$ & \{(1, 6), (2, 8)\} \\ 
11 & $r(b, a)$ & \{(3, 5), (4, 7)\} \\ 
12 & $r(b, b)$ & \{(3, 6), (4, 8)\} \\ 
\bottomrule
\end{tabular}
\end{center}
\end{multicols}
\noindent
Focusing on a particular row, the reason why $(2,7)$ is in $x_9$ is that $\gamma_2 = p(a,b)$, $\gamma_7 = q(b,a)$, the pair of atoms $(p(a,b), q(b,a))$ satisfy the body of clause $c$, and the head of the clause $c$ (for this pair of atoms) is $r(a, a)$ which is $\gamma_9$.

We can transform $X_c$, a set of sets of pairs, into a three dimensional tensor: $\mathbf{X} \in \mathbb{N}^{n \times w \times 2}$.
Here, $w$ is the maximum number of pairs for any $k$ in $1 ... n$.
The width $w$ depends on the number of existentially quantified variables $v$ in the rule template. 
Each existentially quantified variable can take $|C|$ values, so $w = |C| ^ v$. $\mathbf{X}$ is constructed from $X_c$, filling in unused space with $(0,0)$ pairs that point to the pair of atoms $(\perp, \perp)$:
\[
\mathbf{X}[k,m] = \begin{cases}
x_k[m] \; \mbox{if} \; m < |x_k| \\
(0,0) \; \mbox{otherwise}
\end{cases}
\]
This is why we needed to include the falsum atom $\perp$ in $G$, so that the null pairs have some atom to point to. In our running example, this yields:
\begin{multicols}{3}
\small
\begin{center}
\begin{tabular}{llllll}
\toprule
$k$ & $\gamma_k$ & $\mathbf{X}[k]$ \\
\midrule
0 & $\perp$   & $\begin{bmatrix} (0,0) \\ (0,0) \end{bmatrix}$\\[2ex] 
1 & $p(a, a)$ & $\begin{bmatrix} (0,0) \\ (0,0) \end{bmatrix}$\\[2ex]
2 & $p(a, b)$ & $\begin{bmatrix} (0,0) \\ (0,0) \end{bmatrix}$\\[2ex]
3 & $p(b, a)$ & $\begin{bmatrix} (0,0) \\ (0,0) \end{bmatrix}$\\[2ex]
4 & $p(b, b)$ & $\begin{bmatrix} (0,0) \\ (0,0) \end{bmatrix}$\\[2ex]
\bottomrule
\end{tabular}
\end{center}
\columnbreak
\begin{center}
\begin{tabular}{llllll}
\toprule
$k$ & $\gamma_k$ & $\mathbf{X}[k]$ \\
\midrule
5 & $q(a, a)$ & $\begin{bmatrix} (0,0) \\ (0,0) \end{bmatrix}$\\[2ex]
6 & $q(a, b)$ & $\begin{bmatrix} (0,0) \\ (0,0) \end{bmatrix}$\\[2ex]
7 & $q(b, a)$ & $\begin{bmatrix} (0,0) \\ (0,0) \end{bmatrix}$\\[2ex]
8 & $q(b, b)$ & $\begin{bmatrix} (0,0) \\ (0,0) \end{bmatrix}$\\[2ex]
\bottomrule
\end{tabular}
\end{center}
\columnbreak
\begin{center}
\begin{tabular}{lll}
\toprule
$k$ & $\gamma_k$ & $\mathbf{X}[k]$ \\
\midrule
9 & $r(a, a)$ & $\begin{bmatrix} (1,5) \\ (2,7) \end{bmatrix}$\\[2ex]
10 & $r(a, b)$ & $\begin{bmatrix} (1,6) \\ (2,8) \end{bmatrix}$\\[2ex]
11 & $r(b, a)$ &$\begin{bmatrix} (3,5) \\ (4,7) \end{bmatrix}$\\[2ex]
12 & $r(b, b)$ & $\begin{bmatrix} (3,6) \\ (4,8) \end{bmatrix}$\\[2ex]
\bottomrule
\end{tabular}
\end{center}
\end{multicols}
\noindent
Let $\mathbf{X}_1, \mathbf{X}_2 \in \mathbb{N}^{n \times w}$ be two slices of $\mathbf{X}$, taking the first and second elements in each pair:
\[
\mathbf{X}_1 = \mathbf{X}[:, :, 0] \qquad
\mathbf{X}_2 = \mathbf{X}[:, :, 1]
\]
We shall use a function $\mathsf{gather_2} : \mathbb{R}^a \times \mathbb{N}^{b \times c} \rightarrow \mathbb{R}^{b \times c}$:
\[
\mathsf{gather_2}(x,y)[i, j] = x[y[i, j]]
\]
Now we are ready to define $F_c(\mathbf{a})$.
Let $\mathbf{Y}_1, \mathbf{Y}_2 \in [0,1]^{n \times w}$ be the results of assembling the elements of $\mathbf{a}$ according to the matrix of indices in $\mathbf{X}_1$ and $\mathbf{X}_2$:
\[
\mathbf{Y}_1 = \mathsf{gather}_2(\mathbf{a}, \mathbf{X}_1) \qquad \mathbf{Y}_2 = \mathsf{gather}_2(\mathbf{a}, \mathbf{X}_2)
\]
Now let $\mathbf{Z} \in [0,1]^{n \times w}$ contain the results of element-wise multiplying the elements of $\mathbf{Y}_1$ and $\mathbf{Y}_2$:
\[
\mathbf{Z} = \mathbf{Y}_1 \odot \mathbf{Y}_2
\]
Here, $\mathbf{Z}[k,:]$ is the vector of fuzzy conjunctions of all the pairs of atoms that contribute to the truth of $\gamma_k$, according to the current clause.
Now we can define $F_c(\mathbf{a})$ by taking the max fuzzy truth values in $\mathbf{Z}$.
Let $F_c(\mathbf{a}) = \mathbf{a}'$ where $\mathbf{a}'[k] = \max(\mathbf{Z}[k, :])$.

The following table shows the calculation of $F_c(\mathbf{a})$ for a particular valuation $\mathbf{a}$, using our running example $c = r(X, Y) \leftarrow p(X, Z), q(Z, Y)$. Here, since there is one existential variable $Z$, $v=1$, and $w = |\{a,b\}|^{v} = 2$.
\begin{table}[H]
\centering
\label{Z[k]}
\begin{tabular}{lllllllll}
\toprule
$k$ & $\gamma_k$ & $\mathbf{a}[k]$ & $\mathbf{X}_1[k]$ & $\mathbf{X}_2[k]$ & $\mathbf{Y}_1[k]$ & $\mathbf{Y}_2[k]$ & $\mathbf{Z}[k]$ & $F_c(\mathbf{a})[k]$ \\ \midrule
0 & $\perp$ & 0.0 & $\begin{bmatrix} 0 & 0 \end{bmatrix}$ & $\begin{bmatrix} 0 & 0 \end{bmatrix}$ & $\begin{bmatrix} 0 & 0 \end{bmatrix}$ & $\begin{bmatrix} 0 & 0 \end{bmatrix}$ & $\begin{bmatrix} 0 & 0 \end{bmatrix}$ & 0.00  \\[.5ex]
1 & $p(a, a)$ & 1.0 & $\begin{bmatrix} 0 & 0 \end{bmatrix}$ & $\begin{bmatrix} 0 & 0 \end{bmatrix}$ & $\begin{bmatrix} 0 & 0 \end{bmatrix}$ & $\begin{bmatrix} 0 & 0 \end{bmatrix}$ & $\begin{bmatrix} 0 & 0 \end{bmatrix}$ & 0.00 \\[.5ex]
2 & $p(a, b)$ & 0.9 & $\begin{bmatrix} 0 & 0 \end{bmatrix}$ & $\begin{bmatrix} 0 & 0 \end{bmatrix}$ & $\begin{bmatrix} 0 & 0 \end{bmatrix}$ & $\begin{bmatrix} 0 & 0 \end{bmatrix}$ & $\begin{bmatrix} 0 & 0 \end{bmatrix}$ & 0.00 \\[.5ex]
3 & $p(b, a)$ & 0.0 & $\begin{bmatrix} 0 & 0 \end{bmatrix}$ & $\begin{bmatrix} 0 & 0 \end{bmatrix}$ & $\begin{bmatrix} 0 & 0 \end{bmatrix}$ & $\begin{bmatrix} 0 & 0 \end{bmatrix}$ & $\begin{bmatrix} 0 & 0 \end{bmatrix}$ & 0.00 \\[.5ex]
4 & $p(b, b)$ & 0.0 & $\begin{bmatrix} 0 & 0 \end{bmatrix}$ & $\begin{bmatrix} 0 & 0 \end{bmatrix}$ & $\begin{bmatrix} 0 & 0 \end{bmatrix}$ & $\begin{bmatrix} 0 & 0 \end{bmatrix}$ & $\begin{bmatrix} 0 & 0 \end{bmatrix}$ & 0.00 \\[.5ex]
5 & $q(a, a)$ & 0.1 & $\begin{bmatrix} 0 & 0 \end{bmatrix}$ & $\begin{bmatrix} 0 & 0 \end{bmatrix}$ & $\begin{bmatrix} 0 & 0 \end{bmatrix}$ & $\begin{bmatrix} 0 & 0 \end{bmatrix}$ & $\begin{bmatrix} 0 & 0 \end{bmatrix}$ & 0.00 \\[.5ex]
6 & $q(a, b)$ & 0.0 & $\begin{bmatrix} 0 & 0 \end{bmatrix}$ & $\begin{bmatrix} 0 & 0 \end{bmatrix}$ & $\begin{bmatrix} 0 & 0 \end{bmatrix}$ & $\begin{bmatrix} 0 & 0 \end{bmatrix}$ & $\begin{bmatrix} 0 & 0 \end{bmatrix}$ & 0.00 \\[.5ex]
7 & $q(b, a)$ & 0.2 & $\begin{bmatrix} 0 & 0 \end{bmatrix}$ & $\begin{bmatrix} 0 & 0 \end{bmatrix}$ & $\begin{bmatrix} 0 & 0 \end{bmatrix}$ & $\begin{bmatrix} 0 & 0 \end{bmatrix}$ & $\begin{bmatrix} 0 & 0 \end{bmatrix}$ & 0.00 \\[.5ex]
8 & $q(b, b)$ & 0.8 & $\begin{bmatrix} 0 & 0 \end{bmatrix}$ & $\begin{bmatrix} 0 & 0 \end{bmatrix}$ & $\begin{bmatrix} 0 & 0 \end{bmatrix}$ & $\begin{bmatrix} 0 & 0 \end{bmatrix}$ & $\begin{bmatrix} 0 & 0 \end{bmatrix}$ & 0.00 \\[.5ex]
9 & $r(a, a)$ & 0.0 & $\begin{bmatrix} 1 & 2 \end{bmatrix}$ & $\begin{bmatrix} 5 & 7 \end{bmatrix}$ & $\begin{bmatrix} 1.0 & 0.9 \end{bmatrix}$ & $\begin{bmatrix} 0.1 & 0.2 \end{bmatrix}$ & $\begin{bmatrix} 0.1 & 0.18 \end{bmatrix}$ & 0.18 \\[.5ex]
10 & $r(a, b)$ & 0.0 & $\begin{bmatrix} 1 & 2 \end{bmatrix}$ & $\begin{bmatrix} 6 & 8 \end{bmatrix}$ & $\begin{bmatrix} 1.0 & 0.9 \end{bmatrix}$ & $\begin{bmatrix} 0 & 0.8 \end{bmatrix}$ & $\begin{bmatrix} 0 & 0.72 \end{bmatrix}$ & 0.72  \\[.5ex]
11 & $r(b, a)$ & 0.0 & $\begin{bmatrix} 3 & 4 \end{bmatrix}$ & $\begin{bmatrix} 5 & 7 \end{bmatrix}$ & $\begin{bmatrix} 0 & 0 \end{bmatrix}$ & $\begin{bmatrix} 0.1 & 0.2 \end{bmatrix}$ & $\begin{bmatrix} 0 & 0 \end{bmatrix}$ & 0.00 \\[.5ex]
12 & $r(b, b)$ & 0.0 & $\begin{bmatrix} 3 & 4 \end{bmatrix}$ & $\begin{bmatrix} 6 & 8 \end{bmatrix}$ & $\begin{bmatrix} 0 & 0 \end{bmatrix}$ & $\begin{bmatrix} 0 & 0.8 \end{bmatrix}$ & $\begin{bmatrix} 0 & 0 \end{bmatrix}$ & 0.00 \\ \bottomrule
\end{tabular}
\end{table}
Please note that the calculation of the indices in $\mathbf{X}$ is not a differentiable operation.
Calculating $\mathbf{X}$ makes use of the discrete operations $\mathsf{satisfies}$ and $\mathsf{head}$.
The matrix $\mathbf{X}$ of indices is created, ahead of time, before the neural net is constructed.
However, the function $F_c$ on valuations \emph{is} differentiable. 
It takes the indices in $\mathbf{X}$ and applies differentiable operations such as $\mathsf{gather}_2$ and element-wise multiplication.

\subsubsection{Defining Fuzzy Conjunction}
\label{t-norms}

Above, when computing $\mathbf{Z}$, we took the element-wise product: 
\[
\mathbf{Z} = \mathbf{Y}_1 \odot \mathbf{Y}_2
\]
Now $\mathbf{Y}_1$ and $\mathbf{Y}_2$ represent the fuzzy truth values of the two atoms satisfying the body of the predicate, and $\mathbf{Z}$ represents their conjunction.

Now there are many different ways of representing fuzzy conjunction. 
At a high level of generality, we need an operator $\ast : [0,1]^2 \rightarrow [0,1]$ satisfying the conditions on a t-norm~\cite{esteva2001monoidal}:
\begin{itemize}
\item
commutativity: $x \ast y = y \ast x$
\item
associativity: $(x \ast y) \ast z = x \ast (y \ast z)$
\item
monotonicity (i): $x_1 \leq x_2$ implies $x_1 \ast y \leq x_2 \ast y$
\item
monotonicity (ii): $y_1 \leq y_2$ implies $x \ast y_1 \leq x \ast y_2$
\item
unit (i): $x \ast 1 = x$
\item
unit (ii): $x \ast 0 = 0$
\end{itemize}
Now there are a number of operators satisfying these conditions:
\begin{itemize}
\item
Godel t-norm: $x \ast y = min(x, y)$
\item
\L{}ukasiewicz t-norm: $x \ast y = max(0, x+y-1)$
\item
Product t-norm: $x \ast y = x \cdot y$
\end{itemize}
The reason we chose the product t-norm over the alternatives proposed by Godel and \L{}ukasiewicz is that, when back-propagating from the loss to the clause weights, the gradients flow evenly between both $\mathbf{Y}_1$ and $\mathbf{Y}_2$.
With Godel's t-norm, there is no gradient information sent to $\mathbf{Y}_1$ when $\mathbf{Y}_1 > \mathbf{Y}_2$.
Similarly, with \L{}ukasiewicz's t-norm, there is no gradient information sent to \emph{either} $\mathbf{Y}_1$ or $\mathbf{Y}_2$ when $\mathbf{Y}_1 + \mathbf{Y}_2 < 1$.

Experimental evaluation bears this out. 
We tested all three t-norms on our dataset of 20 symbolic problems and found that the product t-norm consistently out-performed Godel's and \L{}ukasiewicz's.
See Section~\ref{symbolic-results} below.

\section{Experiments}
\label{sec:experiments}

ILP has a number of excellent features. 
But it has two main areas of weakness: intolerance to mis-labelled data, and inability to cope with fuzzy or ambiguous data.
After first checking that \system{} is capable of learning standard ILP tasks with discrete error-free input, our experiments were largely focused on testing how \system{} compares with standard ILP in these two problem areas.

We implemented our model in TensorFlow \cite{abadi2016tensorflow} and 
tested it with three types of experiment.
First, we used standard symbolic ILP tasks, where \system{} is given discrete error-free input.
Second, we modified the standard symbolic ILP tasks so that a certain proportion of the positive and negative examples are wilfully mis-labelled.
Third, we tested it with fuzzy, ambiguous data, connecting \system{} to the output of a pretrained convolution neural network that classifies MNIST digits.

\subsection{Hyperparameters}

We first ran a grid search to find ``reasonable'' hyperparameters that could solve each of ten tasks in at least 5\% of all random initialisations of weights\footnote{5\% was chosen, arbitrarily, as a cut-off for ``reasonable'' performance.}.  
We tried a range of optimisation algorithms: Stochastic Gradient Descent, Adam, AdaDelta, and RMSProp. We searched across a range of learning rates in $\{0.5, 0.2, 0.1, 0.05, 0.01, 0.001\}$.
Weights were initialised randomly from a normal distribution with mean 0 and a standard deviation that ranged between 0 and 2 (the standard deviation was a hyperparameter but the mean was fixed). For each configuration of hyperparameter settings, we ran 20 trials, each with different random seeds, for each of 10 tasks. 

We found that both RMSProp and Adam gave reasonable results: both were able to solve all 10 tasks for at least 5 \% of random weight initialisations. 
RMSProp was successful on a range of learning rates from 0.5 to 0.01.

Hyperparameters were validated against training data, not held-out development data. Because of \system{}'s strong inductive bias towards universally quantified rules, programs with a low loss on the training data are also biased towards performing well on the test data.

In all the experiments described below, we stuck to a particular fixed set of hyperparameter settings: we used RMSProp with a learning rate of 0.5, and initialised clause weights by sampling from a $\mathcal{N}(0, 1)$ distribution.

\subsection{Experimental Method}

Before training, rule weights are initialised randomly by sampling from a $\mathcal{N}(0, 1)$ distribution.
We train for 6000 steps, adjusting rule weights to minimise cross entropy loss as described above.

During training, \system{} is given \emph{multiple} $(\mathcal{B}, \mathcal{P}, \mathcal{N})$ triples.
Each triple represents a different possible world.
Each training step, \system{} first samples one of these $(\mathcal{B}, \mathcal{P}, \mathcal{N})$ triples, and then samples a mini-batch from $\mathcal{P} \cup \mathcal{N}$.
Providing multiple alternative possible worlds helps the system to generalise correctly.
It is less likely to focus on irrelevant aspects of one particular situation if it is forced to consider multiple situations.

Each step we sample a mini-batch from the positive and negative examples. 
Note that, instead of using the whole set of positive and negative examples each training step, we just take a random subset.
This mini-batching gives the process a stochastic element and helps to escape local minima.

After 6000 steps, \system{} produces a set of rule weights for each rule template. 
To validate this program, we run the model with the learned weights on an entirely new set of background facts.
This is testing the system's ability to generalise to unseen data.
We compute validation error as the sum of mean-squared difference between the actual label $\lambda$ and the predicted label $\hat{\lambda}$:
\[
loss = \sum_{i=1}^k (\lambda - \hat{\lambda}) ^2
\]

Once \system{} has finished training, we extract the rule weights, and take the soft-max.
Interpreting the soft-maxed weights as a probability distribution over clauses, we measure the ``fuzziness'' of the solution by calculating the entropy of the distribution.
On the discrete error-free tasks, \system{} finds solutions with zero entropy, as long as it does not get stuck in a local minimum.
To extract a human-readable logic program, we just take all clauses whose probability is over some constant threshold (currently set, arbitrarily, to 0.1).

\subsection{ILP Benchmark Tasks}
\label{sub:ilpbenchmark}

We tested \system{} on 20 ILP tasks, taken from four domains: arithmetic, lists, group-theory, and family tree relations.
Some of the arithmetic examples appeared in the work of \citeA{cropper2016learning}.
The list examples are used by \citeA{feser2015synthesizing}.
The family tree dataset comes from \citeA{wang2015soft} and is also used by \citeA{yangdifferentiable}.

We emphasize that the 20 tasks we used have the following common feature: in each case, the program can be learned from a small amount of training data.
\system{} is a memory-expensive solution to ILP (see Appendix \ref{space-requirements}), so only problems with small training sets have been tested.
This is why we have not tested \system{} on the larger ILP datasets, such as Mutagenesis, WebKB, or IMDB.
Although the 20 tasks we use are all quite small in the amount of training data needed to learn them, the programs needing to be synthesised in order to solve them are often complex, involving multiple recursive predicates and invented auxiliary predicates.

In this section, we shall focus on three examples in detail.
The complete list of 20 examples is detailed in Appendix \ref{all_experiments}.

\subsubsection{Learning Even/1 on Natural Numbers}
\label{subsubsec:even-succ2}

The task is to learn the $even$ predicate on natural numbers.
The language contains the monadic predicate $zero$ and the successor relation $succ$.
The background knowledge is the set of basic arithmetic facts defining the $zero$ predicate and $succ$ relation on numbers up to 10:
\[
\mathcal{B} = \{zero(0), succ(0, 1), succ(1, 2), succ(2, 3), ..., succ(9, 10)\}
\]
The positive examples $\mathcal{P}$ are:
\begin{eqnarray*}
\mathcal{P} = \{target(0), target(2), target(4), target(6), target(8), target(10)\}
\}
\end{eqnarray*}
In all these examples, $target$ is the name of the target predicate we are trying to learn. In this case, $target = even$.
The negative examples are 
\[
\mathcal{N} = \{ target(1), target(3), target(5), target(7), target(9)\}
\]
For validation and test, we use positive and negative examples of the $even$ predicate on numbers greater than 10.

One possible language template for this task is:
\begin{itemize}
\item
$P_e: \{zero/1, succ/2\}$
\item
$P_i: \{target/1, pred/2\}$
\end{itemize}
Here, $pred$ is an auxiliary binary predicate.
The set $C$ of constants is just $\{0, 1, ..., 10\}$.

One suitable program template\footnote{There are other templates that work equally well.} for this tasks is:
\begin{eqnarray*}
\tau_{target, 1} & = & (h=target, n_\exists=0, int=False) \\
\tau_{target, 2} & = & (h=target, n_\exists=1, int=True) \\
\tau_{pred, 1} & = & (h=pred, n_\exists=1, int=False) \\
\tau_{pred, 2} & = & \mbox{null}
\end{eqnarray*}
This template specifies two clauses for $target$, and one clause for the auxiliary predicate $pred$.
The architecture assumes that every predicate is defined by exactly two clauses specified by two rule templates. Here, we only need one clause to define $pred$, so the second rule template is null.

\system{} can reliably solve this task. One solution\footnote{All clauses have exactly two literals, so the first target clause is actually $target(X) \leftarrow zero(X), zero(X)$. We omit duplicated literals for readability.} found is:
\begin{eqnarray*}
target(X) & \leftarrow & zero(X) \\ 
target(X) & \leftarrow & target(Y), pred(Y, X) \\ 
pred(X, Y) & \leftarrow & succ(X, Z), succ(Z, Y)
\end{eqnarray*}
Here, $pred(X,Y)$ is true if $X+2 = Y$.

One major concern with ILP systems is that the program template makes it easy to ``bake in'' knowledge of the target solution. 
To evaluate this concern, we compared the above program template with the corresponding metarules needed for Metagol to learn the same task.
See Appendix \ref{metagol_comparison} below.
We also experimented with iterative deepening over the space of program templates. See Section \ref{iterative-deepening} below.

\subsubsection{Learning Graph Cyclicity}
\label{subsubsec:cycle}

In this example, \system{} is given various graphs (nodes connected by edges).
The task is to learn the \emph{is-cyclic} predicate.
This predicate is true of a node if there is a path, a sequence of edge connections, from that node back to itself.
(Note that this concept is not definable in first-order logic. 
We need some form of \emph{recursion} to capture the idea that the sequence of edge connections could be indefinitely long).

\begin{figure}[!ht]
  \subfloat[First Training Triple\label{fig:loop-first-training}]{%
    \includegraphics[width=0.45\textwidth]{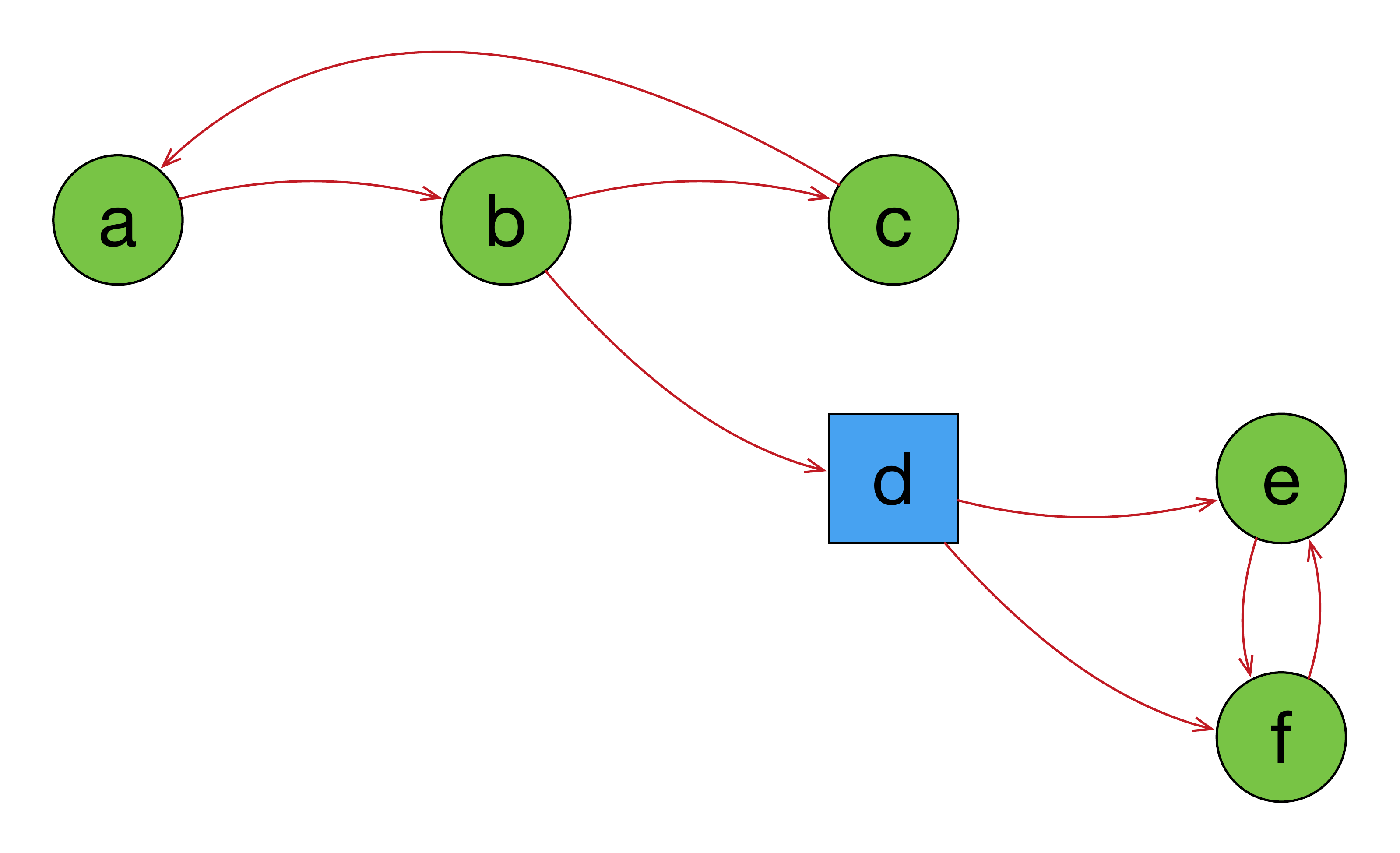}
  }
  \hfill
  \subfloat[Second Training Triple\label{fig:loop-second-training}]{%
    \includegraphics[width=0.35\textwidth]{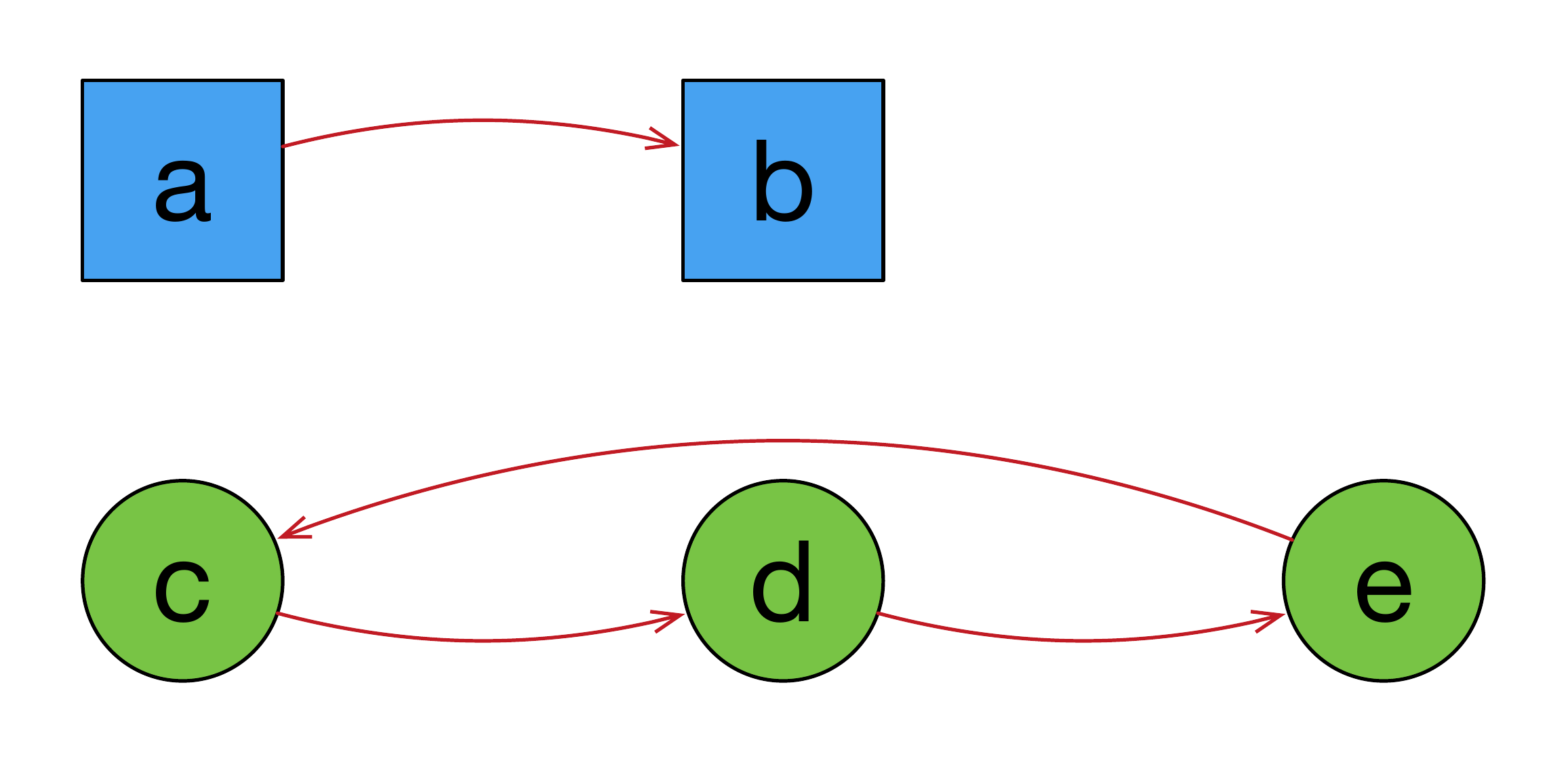}
  }
  
  \subfloat[Validation Triple\label{fig:loop-validation}]{%
    \includegraphics[width=0.45\textwidth]{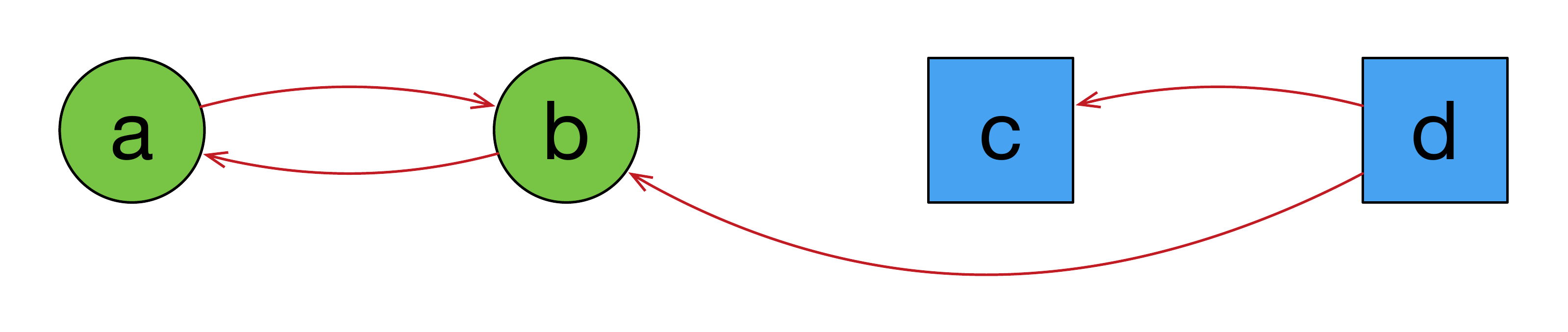}
  }
  
  \caption{Three $(\mathcal{B}, \mathcal{P}, \mathcal{N})$ Triples for the Cyclic Task. Green circular nodes satisfy the \emph{is-cyclic} predicate. Blue square nodes do not.}
  \label{fig:dummy}
\end{figure}

In this task, there are two $(\mathcal{B}, \mathcal{P}, \mathcal{N})$ triples, displayed in Figures \ref{fig:loop-first-training} and \ref{fig:loop-second-training}.
Providing multiple alternative possible worlds helps the system to generalise correctly.

The validation triple is shown in Figure \ref{fig:loop-validation}. 
Again, we are testing generalisation in an unseen situation.

The language for this task is:
\begin{itemize}
\item
$P_e = \{edge/2\}$
\item
$P_i = \{target/1, pred/2\}$
\item
$C = \{a, b, c, d, e, f\}$
\end{itemize}
Here, $target$ is the \emph{is-cyclic} predicate we are trying to learn, and $pred$ is an auxiliary additional predicate.

One of the solutions found is:
\begin{eqnarray*}
target(X) & \leftarrow & pred(X, X) \\
pred(X, Y) & \leftarrow & edge(X, Y) \\
pred(X, Y) & \leftarrow & pred(X, Z), pred(Z, Y)
\end{eqnarray*}
Note that the invented predicate $pred$ is the transitive closure of the edge relation, true of $X$ and $Y$ if there is some sequence of edge connections, of any length, between $X$ and $Y$.
In order to solve this task, \system{} had to use an invented \emph{recursive} predicate.
\system{} is able to solve this problem 100\% of the time, irrespective of initial random weights.

\subsubsection{Learning Fizz-Buzz}
\label{subsubsec:fizz-buzz}

Joel Grus provides\footnote{See the darkly humorous blog post \url{http://joelgrus.com/2016/05/23/fizz-buzz-in-tensorflow/}.} a TensorFlow implementation of the children's game ``Fizz-Buzz''.
In this game, the players must enumerate the natural numbers --- but if the number is divisible by 3, they should say ``Fizz'', and if the number is divisible by 5, they should say ``Buzz''.
 Grus used a standard multilayer perceptron (MLP). 
The training data was integers from 100 to 1024. The integers below 100 were held out as test data.

On all runs, the MLP was unable to generalise correctly to the unseen held-out test data\footnote{``I guess'', he suggests with rueful irony, ``maybe I should have used a deeper network''.}.
This is because a standard MLP can be interpreted as suffering from what \citeA{mccarthy1988epistemological} dubbed the ``propositional fixation'': it remembers particular \emph{instances} but does not capture \emph{generalities}.

We tested \system{} on Fizz-Buzz, after translating it into a curriculum learning task.
We first asked \system{} to learn Fizz, and then, once it had learned the correct rules, we fed the results of those rules, as axioms, into the task of learning Buzz\footnote{Traditionally, Fizz-Buzz has an additional rule, that if a number divides by 3 and by 5, then we say ``Fizz-Buzz'' which is a separate class from either ``Fizz'' or ``Buzz''. We did not model this detail.}.

To learn Fizz, we provided the following language:
\begin{itemize}
\item
$P_e = \{zero/1, succ/2\}$
\item
$P_i = \{target/1, pred1/2, pred2/2 \}$
\item
$C = \{0, 1, 2, 3, 4, 5, 6\}$
\end{itemize}
Here, $zero$ is the monadic predicate that is true of $X$ if $X = 0$, $succ$ is the successor relation, 
and $target(X)$ is the ``Fizz'' predicate we are trying to learn, true of numbers that are divisible by 3.
In this case, we need two auxiliary predicates $pred1$ and $pred2$ to solve the task.
The background axioms are
\[
\{zero(0)\} \cup \{ succ(n, n+1) \; | \; n \in \{0, ..., 5\} \}
\]
Note the small number of integers that \system{} is trained on.
The positive and negative examples are:
\begin{eqnarray*}
\mathcal{P} & = & \{target(0), target(3), target(6)\} \\
\mathcal{N} & = & \{target(1), target(2), target(4), target(5)\} 
\end{eqnarray*}
Note that there are three intensional predicates in the language: the $target$ predicate and two auxiliary predicates, $pred1$ and $pred2$.
The validation data is an expansion of the earlier setup to larger integers: we test if 7, 8, and 9 satisfy the Fizz property.

\system{} can find a solution to this problem. 
One of the solutions that it finds is:
\begin{eqnarray*}
target(X) & \leftarrow& zero(X) \\
target(X) & \leftarrow& target(Y), pred1(Y, X) \\
pred1(X, Y) & \leftarrow & succ(X, Z), pred2(Z, Y) \\
pred2(X, Y) & \leftarrow & succ(X, Z), succ(Z, Y)
\end{eqnarray*}
In order to solve this task, it had to construct two auxiliary invented predicates: $pred1(X,Y)$ is true if $X$ is three greater than $Y$, and 
$pred2(X,Y)$ is true if $X$ is two greater than $Y$.

The rules that \system{} has learned are robust to test data of any size: although it was trained on integers from 0 to 6, the learned program works effectively on test integers of any size.
There is no chance of the learned program suddenly failing when the integers reach a certain size. 
There is no worry that \system{}'s performance will suddenly fall off a cliff when the integers exceed 1000, say. 

Grus' MLP failed to solve the Fizz task because it did not capture the general, universally quantified rules needed to understand this task.
\system{}, by contrast, is entirely incapable of doing anything other than producing general universally quantified rules.
The language-bias in this model is very strong. The only rules that \system{} understands are rules involving free variables. There are no constants allowed whatsoever.
While a MLP struggles to generalise, \system{} is \emph{doomed} to generalise. 

Now \system{} does not always find the correct solution.
All existing neural program induction systems are susceptible to getting stuck in local minima when they are started with an unfortunate random initialisation of initial weights.
(This is a result of the ``spikiness'' of the loss landscape in program induction tasks).
\system{} is no exception, and on complex problems it only finds a correct solution a certain proportion of the time.
However, this dependence on initial random weights is not a serious problem in this particular system, because we can model-select based on the training data.
After a certain number (e.g. 100) of runs, we look at the loss on the training data, and choose the model with the lowest loss.
Even though we are model-selecting on training data, the top-scoring model will generalise robustly to unseen test data because generalisation has been baked into the system from the beginning. The only possible rules it can learn are universally quantified rules. Furthermore, while this was not explored in the current work, one could automate the process of finding the right intialisations by designing a search procedure over clause weight intialisations based on how well the entropy of the distribution over clauses is minimised by a fixed number of training steps.

One of the attractive features of ILP systems is that knowledge gained from one task can be packaged up, in a compressed and readable format, into a set of clauses, and this knowledge can be transferred directly to new tasks.
We took advantage of this transfer learning feature in the Fizz-Buzz task:
when learning Buzz, we gave \system{} the results of the invented predicates $pred1$ and $pred2$ from the Fizz task.
We compiled the learned rules for $pred1$ and $pred2$ into a set of background atoms:
\begin{eqnarray*}
pred1(n, n+3) \\
pred2(n, n+2) \\
\end{eqnarray*}
The language for the Buzz task is:
\begin{itemize}
\item
$P_e = \{ zero/1, succ/2, pred1/2, pred2/2\}$
\item
$P_i = \{pred3/2, target/1 \}$
\item
$C = \{0, 1, 2, 3, 4, 5, 6, 7, 8, 9\}$
\end{itemize}
The background information $\mathcal{B}$ is:
\[
\{zero(0)\} \cup \{succ(n, n+1)\} \cup \{pred1(n, n+3)\} \cup \{pred2(n, n+2)\}
\]
The positive and negative examples are:
\begin{eqnarray*}
\mathcal{P} & = & \{target(0), target(5)\} \\
\mathcal{N} & = & \{target(1), target(2), target(3), target(4), target(6), target(7), target(8), target(9)\} \\
\end{eqnarray*}
\system{} is sometimes (30\% of weight initialisations) able to find the correct solution to this problem.
One solution it finds is:
\begin{eqnarray*}
target(X) & \leftarrow& zero(X) \\
target(X) & \leftarrow& target(Y), pred3(Y, X) \\
pred3(X, Y) & \leftarrow & pred1(X, Z), pred2(Z, Y)
\end{eqnarray*}
Note that the invented predicate $pred3$ is true of $X$ and $Y$ if $X$ is 5 greater than $Y$.

This is, to our knowledge, the first neural program induction system that learns Fizz-Buzz and generalises robustly to unseen test data.

\subsubsection{Results on All Symbolic Tasks}
\label{symbolic-results}

The results of our experiments on all 20 discrete error-free symbolic tasks are shown in Table \ref{symbolic-results-table}.
We compare \system{} with {\bf Metagol}, a state of the art ILP system \cite{cropper2016learning,cropper2015meta,muggleton2015meta,muggleton2014meta}.
For each task, we show the number of intensional predicates $|P_i|$;
this is usually a good estimate of the hardness of the task.
We also indicate whether each task requires at least one recursive clause. 
This is another factor in the hardness of the task.
The hardest tasks involve synthesising multiple intensional predicates, some of which are recursive.

In the few cases where Metagol was unable to find the correct program, this was because it ran out of time.
We gave Metagol and \system{} the same fixed time limit (24 hours running on a standard workstation).
After correspondence with Andrew Cropper, one of the authors of Metagol, we discovered that in each of the problem cases, Metagol was stuck in an infinite loop due to the presence of a recursive metarule interacting with a cycle in the data.

Conversely, there are some problems that Metagol can solve but that \system{} is unable to.
The main limitation of \system{} is that it requires a large amount of memory (see Appendix \ref{space-requirements}).
Because of its memory consumption, we currently restrict ourselves to nullary, unary, and binary predicates.
So problems that require ternary predicates (for example, the chess strategy problem in \citeR{cropper2016learning}) are currently outside the scope of \system{}.
Similarly, the robot waiter example described by \citeA{cropper2016learning} is also too large for \system{} to handle in its present form.

We emphasize that a direct apples-to-apples comparison between Metagol and \system{} on these 20 tasks is hard because the two systems require different forms of program template. See Appendix \ref{metagol_comparison}.

\begin{table}[htpb]
\centering
{\small
\begin{tabular}{llllllll}
\toprule
{\bf Domain} & {\bf Task} & $|P_i|$ & {\bf Recursive} & {\bf \shortstack[l]{Metagol\\ Performance}} & {\bf \shortstack[l]{$\partial$ILP\\ Performance}}\\ \midrule
Arithmetic & Predecessor & 1 & No & $\checkmark$ & $\checkmark$ \\
Arithmetic & Even / odd & 2 & Yes &  $\checkmark$ & $\checkmark$ \\
Arithmetic & Even / succ2 & 2 & Yes &  $\checkmark$ & $\checkmark$ \\
Arithmetic & Less than & 1 & Yes &  $\checkmark$ & $\checkmark$ \\
Arithmetic & Fizz & 3 & Yes & $\checkmark$ & $\checkmark$ \\
Arithmetic & Buzz & 2 & Yes & $\checkmark$ & $\checkmark$ \\
Lists & Member & 1 & Yes & $\checkmark$ & $\checkmark$ \\
Lists & Length & 2 & Yes & $\checkmark$ & $\checkmark$ \\
Family Tree & Son & 2 & No & $\checkmark$ & $\checkmark$ \\
Family Tree & Grandparent & 2 & No & $\checkmark$ & $\checkmark$ \\
Family Tree & Husband & 2 & No & $\checkmark$ & $\checkmark$ \\
Family Tree & Uncle & 2 & No & $\checkmark$ & $\checkmark$ \\
Family Tree & Relatedness & 1 & No & $\times$ & {\bf $\checkmark$} \\
Family Tree & Father & 1 & No & $\checkmark$ & $\checkmark$ \\
Graphs & Undirected Edge & 1 & No & $\checkmark$ & $\checkmark$ \\
Graphs & Adjacent to Red & 2 & No & $\checkmark$ & $\checkmark$ \\
Graphs & Two Children & 2 & No & $\checkmark$ & $\checkmark$ \\
Graphs & Graph Colouring & 2 & Yes &$\checkmark$ & $\checkmark$ \\
Graphs & Connectedness & 1 & Yes & $\times$ & {\bf $\checkmark$} \\
Graphs & Cyclic & 2 & Yes & $\times$ & {\bf $\checkmark$} \\
\bottomrule
\end{tabular}}
\caption{A Comparison Between \system{} and Metagol on 20 Symbolic Tasks}
\label{symbolic-results-table}
\end{table}

A breakdown of \system{}'s performance on the 20 symbolic tasks is shown in Table \ref{symbolic-results-breakdown}. 
We compare four different methods: the core \system{} method, a modification of the \system{} method that uses Godel's t-norm instead of the Product t-norm (see Section \ref{t-norms} above), a modification that uses L{}ukasiewicz's t-norm, and a variant that uses $f_{amalgamate}(x,y) = \max(x,y)$, described in Section \ref{inference}.
For each method, we ran 200 times, with different random initialisations of the weights sampled from a $\mathcal{N}(0, 1)$ distribution.

The numbers in the table represent the percentage of runs that got less than $1\mathrm{e}{-4}$ mean squared error on the validation set (i.e. the percentage of runs that generalised correctly to unseen data).
All existing neural program induction systems are sensitive to the particular values of the initial random weights, and are prone to getting stuck in local minima in some proportion of the runs \cite{neelakantan2015neural,kaiser,bunel2016adaptive,gaunt2016terpret,feser2016neural}.
So it is important to be transparent and make explicit what proportion of random weight initialisations are successful. 
(We stress, again, that the system's susceptibility to random weight initialisation is not actually a \emph{problem} in our particular case, since we can model-select based on training data, without danger of overfitting on test data, because of the architectural constraints enforcing generality of the rules learned. See Section \ref{subsubsec:fizz-buzz} for a fuller discussion).

\begin{table}[htpb]
\centering
{\small
\begin{tabular}{llllllll}
\toprule
{\bf Domain} & {\bf Task} & $|P_i|$ & {\bf Recursive} & {\bf $\partial$ILP} & {\bf Godel} & {\bf \L{}ukasiewicz} & {\bf Max}\\
\midrule
Arithmetic & Predecessor & 1 & No & {\bf 100.0} & {\bf 100.0} & {\bf 100.0} & {\bf 100.0} \\
Arithmetic & Even / odd & 2 & Yes &  {\bf 100.0} & 44.0 & {\bf 52.0} & 34.0 \\
Arithmetic & Even / succ2 & 2 & Yes &  {\bf 48.5} & 28.0 & 6.0 & 20.5 \\
Arithmetic & Less than & 1 & Yes &  {\bf 100.0} & {\bf 100.0} & {\bf 100.0} & {\bf 100.0} \\
Arithmetic & Fizz & 3 & Yes &  {\bf 10.0} & 1.5 & 0.0 & 5.5\\
Arithmetic & Buzz & 2 & Yes &  14.0 & {\bf 35.0} & 3.5 & 5.5 \\
Lists & Member & 1 & Yes &  {\bf 100.0} & {\bf 100.0} & {\bf 100.0} & {\bf 100.0} \\
Lists & Length & 2 & Yes &  {\bf 92.5} & 59.0 & 6.0 & 82.0\\
Family Tree & Son & 2 & No &  {\bf 100.0} & 94.5 & 0.0 & 99.5 \\
Family Tree & Grandparent & 2 & No &  {\bf 96.5} & 61.0 & 0.0 & {\bf 96.5} \\
Family Tree & Husband & 2 & No &  {\bf 100.0} & {\bf 100.0} & {\bf 100.0} & {\bf 100.0} \\
Family Tree & Uncle & 2 & No &  {\bf 70.0} & 60.5 & 0.0 & 68.0 \\
Family Tree & Relatedness & 1 & No &  {\bf 100.0} & {\bf 100.0} & {\bf 100.0} & {\bf 100.0} \\
Family Tree & Father & 1 & No &  {\bf 100.0} & {\bf 100.0} & {\bf 100.0} & {\bf 100.0} \\
Graphs & Undirected Edge & 1 & No &  {\bf 100.0} & {\bf 100.0} & {\bf 100.0} & {\bf 100.0}\\
Graphs & Adjacent to Red & 2 & No &  {\bf 50.5} & 40.0 & 1.0 & 42.0 \\
Graphs & Two Children & 2 & No &  {\bf 95.0} & 74.0 & 53.0 & {\bf 95.0} \\
Graphs & Graph Colouring & 2 & Yes &  {\bf 94.5} & 81.0 & 2.5 & 90.0  \\
Graphs & Connectedness & 1 & Yes &  {\bf 100.0} & {\bf 100.0} & {\bf 100.0} & {\bf 100.0} \\
Graphs & Cyclic & 2 & Yes &  {\bf 100.0} & {\bf 100.0} & 0.0 & {\bf 100.0} \\
\bottomrule
\end{tabular}}
\caption{The Percentage of Runs that Achieve Less Than $1\mathrm{e}{-4}$ Mean Squared Test Error}
\label{symbolic-results-breakdown}
\end{table}

\subsection{Dealing with Mislabelled Data}
\label{sub:mislabelledexp}

Standard ILP finds a set $R$ of clauses such that:
\begin{eqnarray*}
\mathcal{B}, R \models \gamma \; \text{for all} \; \gamma \in \mathcal{P} \\
\mathcal{B}, R \nvDash \gamma\; \text{for all} \; \gamma \in \mathcal{N} 
\end{eqnarray*}
These requirements are strict and there is no room for error.
If one of the positive examples $\mathcal{P}$ or negative examples $\mathcal{N}$ is mislabelled, then standard ILP cannot find the intended program.

\system{}, by contrast, is minimising a loss rather than trying to satisfy a strict requirement, so is capable of handling mislabelled data.
We tested this by adding a global variable $\rho$, the proportion of mislabelled data, and varying it between 0 and 1 in small increments.
At the beginning of each experiment, a proportion $\rho$ of the atoms  from $\mathcal{P}$ and $\mathcal{N}$ are sampled, without replacement, and transferred to the other group.
Note that the algorithm sees the same consistently mislabelled examples over all training steps, rather than letting the mis-labelling vary between training steps.

\begin{figure}[ht]
  \subfloat[Predecessor\label{fig:predecessor}]{%
    \includegraphics[scale=0.25]{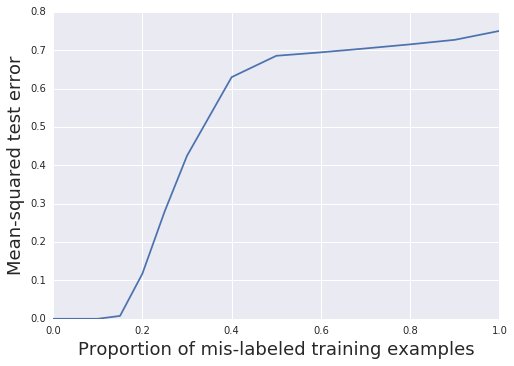}
  }
  \hfill
  \subfloat[Less-Than\label{fig:less_than}]{%
    \includegraphics[scale=0.25]{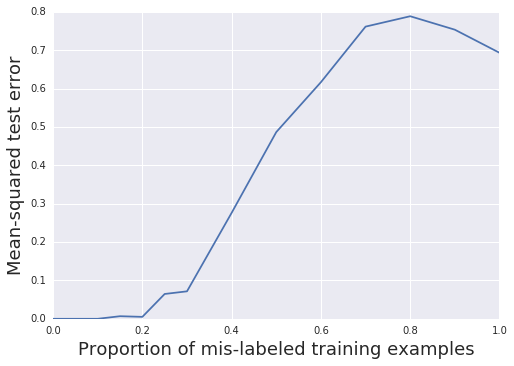}
  }
  \hfill
  \subfloat[List: Member\label{fig:member}]{%
    \includegraphics[scale=0.25]{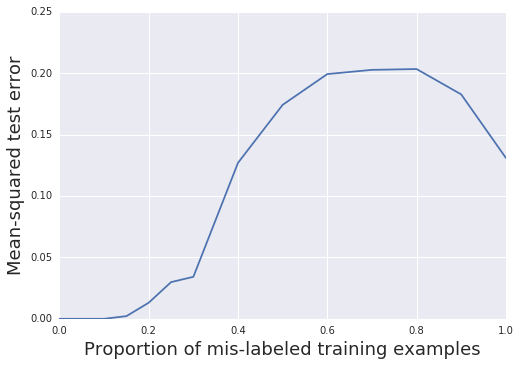}
  }
  
  \subfloat[Family Tree: Son\label{fig:son}]{%
    \includegraphics[scale=0.25]{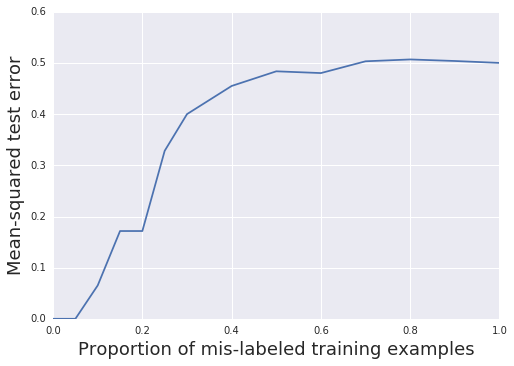}
  }
  \hfill
  \subfloat[Graph: Connectedness\label{fig:connectedness}]{%
    \includegraphics[scale=0.25]{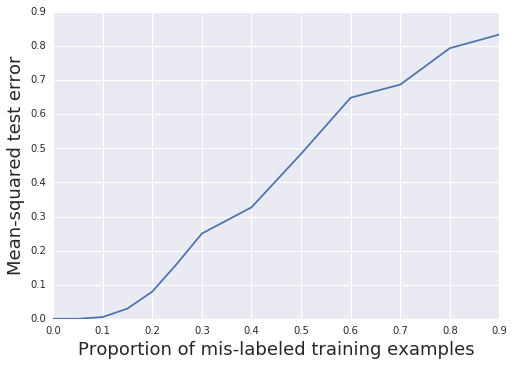}
  }
  \hfill
  \subfloat[Graph: Undirected Edge\label{fig:undirected}]{%
    \includegraphics[scale=0.25]{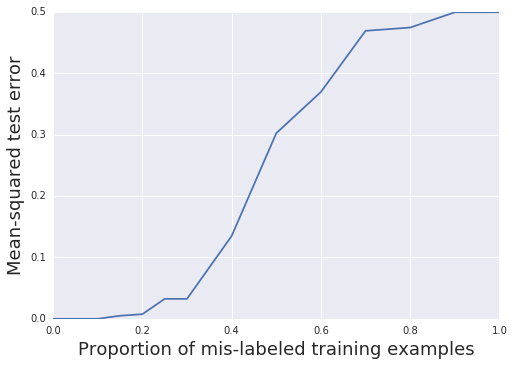}
  }
  
  \caption{Mean squared test error as a function of the proportion of mislabelled data}
  \label{fig:mislabelled-graph}
\end{figure}

The results are shown in Table \ref{mislabelled-results} and Figure \ref{fig:mislabelled-graph}.
Each cell shows the mean squared test error for a particular task with a particular proportion of mislabelled data.
The results show that \system{} is robust to mislabelled data.
The mean-squared error degrades gracefully as the proportion of mislabelled data increases. 
This is in stark contrast to a traditional ILP system, where test error increases sharply as soon as there is a single piece of mislabelled training data.
At 10\% mislabelled data, \system{} still finds a perfect answer in 5 out of 6 of the tasks.
In some of the tasks, \system{} still finds good answers with 20\% or 30\% mislabelled data.

\begin{table}[htpb]
\centering
{\small
\begin{tabular}{lllllll}
\toprule
$\rho$ & {\bf Predecessor}  & {\bf Less Than} & {\bf Member}  & {\bf Son} & {\bf Connectedness}  & {\bf \shortstack[l]{Undirected\\Edge}} \\ 
\midrule
0.00 & 0.000 & 0.000 & 0.000 & 0.000 & 0.000 & 0.000  \\
0.05 & 0.000 & 0.000 & 0.000 & 0.000 & 0.000 & 0.000  \\
0.10 & 0.000 & 0.000 & 0.000 & 0.065 & 0.005 & 0.000  \\
0.15 & 0.007 & 0.006 & 0.002 & 0.171 & 0.030 & 0.005  \\
0.20 & 0.117 & 0.005 & 0.013 & 0.171 & 0.079 & 0.007  \\
0.25 & 0.279 & 0.064 & 0.029 & 0.328 & 0.161 & 0.032  \\
0.30 & 0.425 & 0.071 & 0.034 & 0.400 & 0.250 & 0.032  \\
0.40 & 0.629 & 0.276 & 0.127 & 0.455 & 0.326 & 0.134  \\
0.50 & 0.686 & 0.486 & 0.174 & 0.483 & 0.483 & 0.302  \\
0.60 & 0.695 & 0.616 & 0.199 & 0.480 & 0.648 & 0.369  \\
0.70 & 0.705 & 0.761 & 0.202 & 0.503 & 0.686 & 0.469  \\
0.80 & 0.715 & 0.788 & 0.203 & 0.507 & 0.793 & 0.474  \\
0.90 & 0.727 & 0.753 & 0.182 & 0.504 & 0.833 & 0.499  \\
\bottomrule
\end{tabular}}
\caption{Mean squared test error as a function of the proportion $\rho$ of mislabelled data}
\label{mislabelled-results}
\end{table}

\begin{figure}[!ht]
  \subfloat[Predecessor\label{fig:entropy-predecessor}]{%
    \includegraphics[scale=0.25]{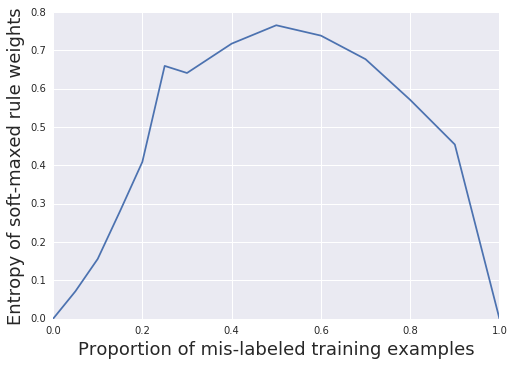}
  }
  \hfill
  \subfloat[Less-Than\label{fig:entropy-less-than}]{%
    \includegraphics[scale=0.25]{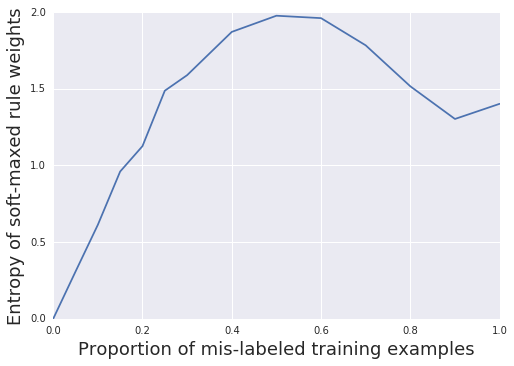}
  }
  \hfill
  \subfloat[List: Member\label{fig:entropy-member}]{%
    \includegraphics[scale=0.25]{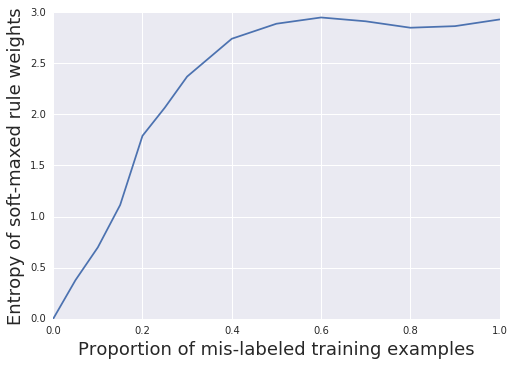}
  }
  
  \subfloat[Family Tree: Son\label{fig:entropy-son}]{%
    \includegraphics[scale=0.25]{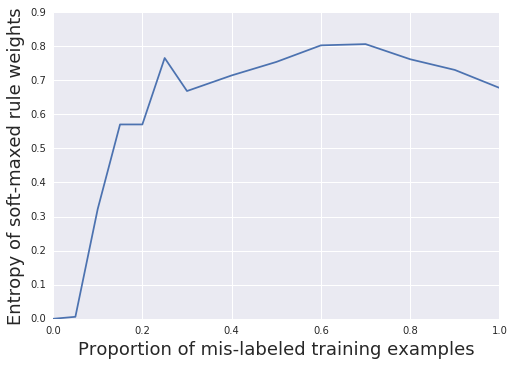}
  }
  \hfill
  \subfloat[Graph: Connectedness\label{fig:entropy-connectedness}]{%
    \includegraphics[scale=0.25]{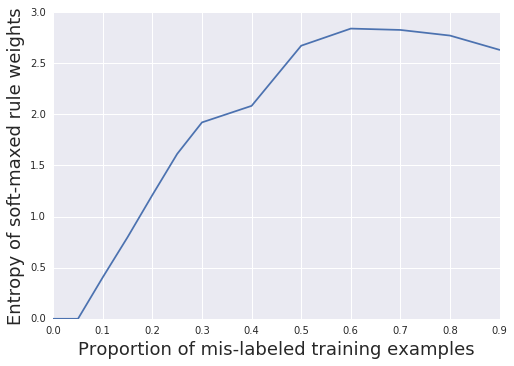}
  }
  \hfill
  \subfloat[Graph: Undirected Edge\label{fig:entropy-undirected}]{%
    \includegraphics[scale=0.25]{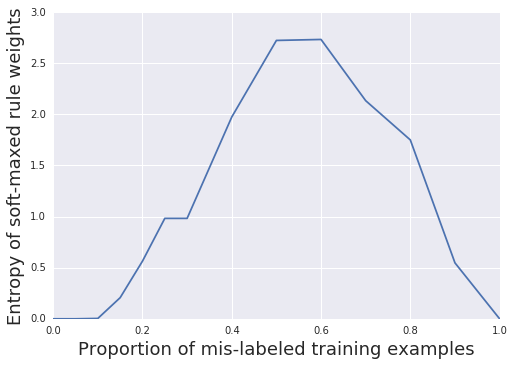}
  }
  
  \caption{Entropy of Rule Weights as the Proportion of Mislabelled Examples Increases}
  \label{fig:entropy}
\end{figure}

Figure \ref{fig:entropy} shows the entropy of the rules as the proportion of mislabelled examples increases.
The rule weights, when soft-maxed, represent a probability distribution over clauses.
The entropy of that probability distribution measures the fuzziness of the learned rules.
Note that, in some cases, entropy goes down as the proportion of mislabelled data approaches 1. 
Sometimes, it is easier to find a clear rule when all the data is mislabelled than when half of it is is mislabelled. 
For example, if we are learning the $even$ predicate on natural numbers, and we switch all the positive and negative examples around, then we will learn a rule for the $odd$ numbers, and this rule will have zero entropy.

\subsection{Dealing with Ambiguous Data}
\label{sub:ambigexp}

Standard ILP assumes that atoms are either true or false. 
There is no room for vagueness or indecision.
\system{}, by contrast, uses a continuous semantics which maps atoms to the real unit interval $[0,1]$.
It was designed to reason with cases where we are unsure if a proposition is true or false. (See Section \ref{subsec:valuations}).

We tested \system{}'s ability to handle ambiguous input by connecting it to the results of a pre-trained convolution neural networks (a ``convnet'') in the style of \citeA{krizhevsky2012imagenet}, trained on MNIST classification, following \citeA{lecun1998gradient}.
Of course, MNIST is no longer a seriously challenging image-classification problem. 
The aim here is not to advance the state of the art in image-recognition, but to hook up a well-understood and easily-trainable image-recognition system to \system{}, to test how \system{} fares with ambiguous data.

We ran tests on 10 separate tasks. In the next few sections, we describe four of those tasks in detail.

\subsubsection{Learning Even Numbers from Raw Pixel Images}

In this task, the system is given a sequence of $28 \times 28$ MNIST images.
The training signal is a single binary value, indicating whether the MNIST image represents a number that is even (1) or odd (0).

\begin{figure}
\includegraphics[width=\textwidth]{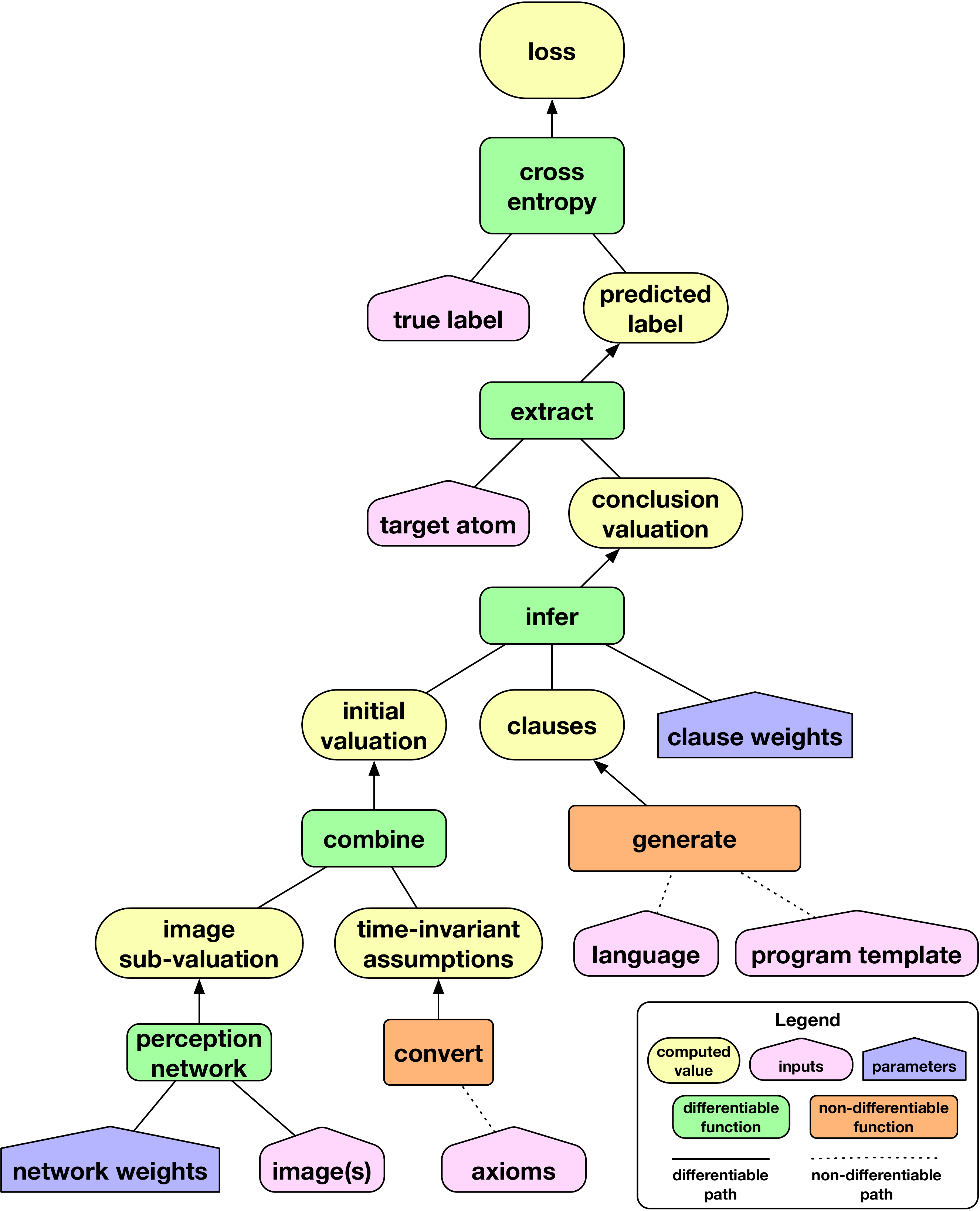}
\caption{\system{} from Raw Pixel Images}
\label{mnist-even}
\end{figure}

Figure \ref{mnist-even} shows how the convnet is connected to \system{}.
The image is fed into the pre-trained convnet.
The logits from the convnet are transformed, via soft-max, into a probability distribution over a set of atoms.
For example, if the image represents a ``2'', the sub-valuation\footnote{A sub-valuation is part of a valuation that is restricted to a single predicate.} produced by the convnet might be:
\begin{align*}
image(0) \mapsto 0.0 & \quad & image(3) \mapsto 0.0 \\
image(1) \mapsto 0.0 & \quad & image(4) \mapsto 0.1 \\
image(2) \mapsto 0.8 & \quad & image(5) \mapsto 0.1 \\
\end{align*}
We only consider integers 0-5 to keep the example simple.

We also have a set of background atoms $\mathcal{B}$. In this case:
\[
\mathcal{B} = \{zero(0)\} \cup \{succ(n, n+1) \; | \; n \in \{0, ..., 4\} \}
\]
$\mathcal{B}$ is transformed into a background valuation, and this valuation is merged with the sub-valuation for $image$ to produce the initial valuation.

The $target$ predicate in this case is a nullary\footnote{A predicate is nullary if it does not take any arguments. \system{} currently handles nullary, unary, and binary predicates only.} predicate: $target$ is true if the image I can currently see is even. 

We emphasise that there is an extra element in Figure \ref{mnist-even} that is not present in Figure \ref{fig:architecture} above.
In Figure \ref{mnist-even}, the initial valuation is generated by combining two elements: the axiom valuation and the image sub-valuation.
The initial valuation represents everything that I know to currently be the case.
This contains both what is timelessly true (the example-invariant axioms) and what is currently true (the example-varying image sub-valuation, representing the image I am currently seeing). 
In Figure \ref{fig:architecture}, there is no equivalent of the image sub-valuation. 

The language $L$ for this task is:
\begin{itemize}
\item
$P_e: \{zero/1, succ/2, image/1\}$
\item
$P_i: \{target/0, pred1/1, pred2/1\}$
\item
$C = \{0, ..., 5\}$
\end{itemize}
Note that this task requires synthesising two auxiliary predicates, $pred1$ and $pred2$.

One of the solutions \system{} finds for this task is:
\begin{eqnarray*}
target() & \leftarrow & image(X), pred1(X) \\
pred1(X) & \leftarrow & zero(X) \\
pred1(X) & \leftarrow & succ(Y, X), pred2(Y) \\
pred2(Y) & \leftarrow & succ(Y, X), pred1(Y)
\end{eqnarray*}
The top rule says: ``the target proposition is true if the image I am currently seeing has numeral $X$, and $X$ satisfies pred1''.
The invented $pred1$ predicate is recognisably the $even$ predicate, with a base case for zero and a step case.
The $pred2$ predicate is the $odd$ predicate.
Solving this task involved synthesising two mutually recursive invented predicates. 

\subsubsection{Learning if the Left Image is Exactly Two Less than the Right Image}

In this task, the system is given a \emph{pair} of images (left and right) each training step.
The training label is a 1 if the number represented by the left image is exactly two less than the number represented by the right image.
The language $L$ for this task is:
\begin{itemize}
\item
$P_e: \{zero/0, succ/2, image1/1, image2/1\}$
\item
$P_i: \{target/0, pred1/1, pred2/2\}$
\item
$C = \{0, ..., 5\}$
\end{itemize}
Here, $image1(X)$ is true if the left image represents the integer $X$, and $image2(X)$ is true if the right image represents $X$.
Note that again there are two auxiliary predicates to synthesise as well as the target predicate.

One of the solutions that \system{} finds is:
\begin{eqnarray*}
target() & \leftarrow & image2(X), pred1(X) \\
pred1(X) & \leftarrow & image1(Y), pred2(X, Y) \\
pred2(X, Y) & \leftarrow & succ(Y, Z), succ(Z, X)
\end{eqnarray*}
Here, $target$ is true if the right image $X$ satisfies the $pred1$ predicate, and $pred1(X)$ is true if the left image $Y$ is two less than $X$.

\subsubsection{Learning if the Right Image Equals 1}

In this task, the system is given a pair of images (left and right) each training step.
The training label is a 1 if the number represented by the right image is exactly 1.
The system must learn to ignore the label of the left image.
The language for this task is:
\begin{itemize}
\item
$P_e: \{zero/0, succ/2, image1/1, image2/1\}$
\item
$P_i: \{target/0, pred1/1\}$
\item
$C = \{0, ..., 5\}$
\end{itemize}
One of the solutions that \system{} finds is:
\begin{eqnarray*}
target() & \leftarrow & image2(X), pred1(X) \\
pred1(X) & \leftarrow & zero(Y), succ(Y, X)
\end{eqnarray*}
It has learned to correctly ignore the content of the left image, $image1$.
Note that the invented predicate $pred1$ is true of $X$ if and only if $X = 1$.

\subsubsection{Learning the Less-Than Relation}

\begin{figure}[htb]
\centering
\includegraphics[width=0.4\textwidth]{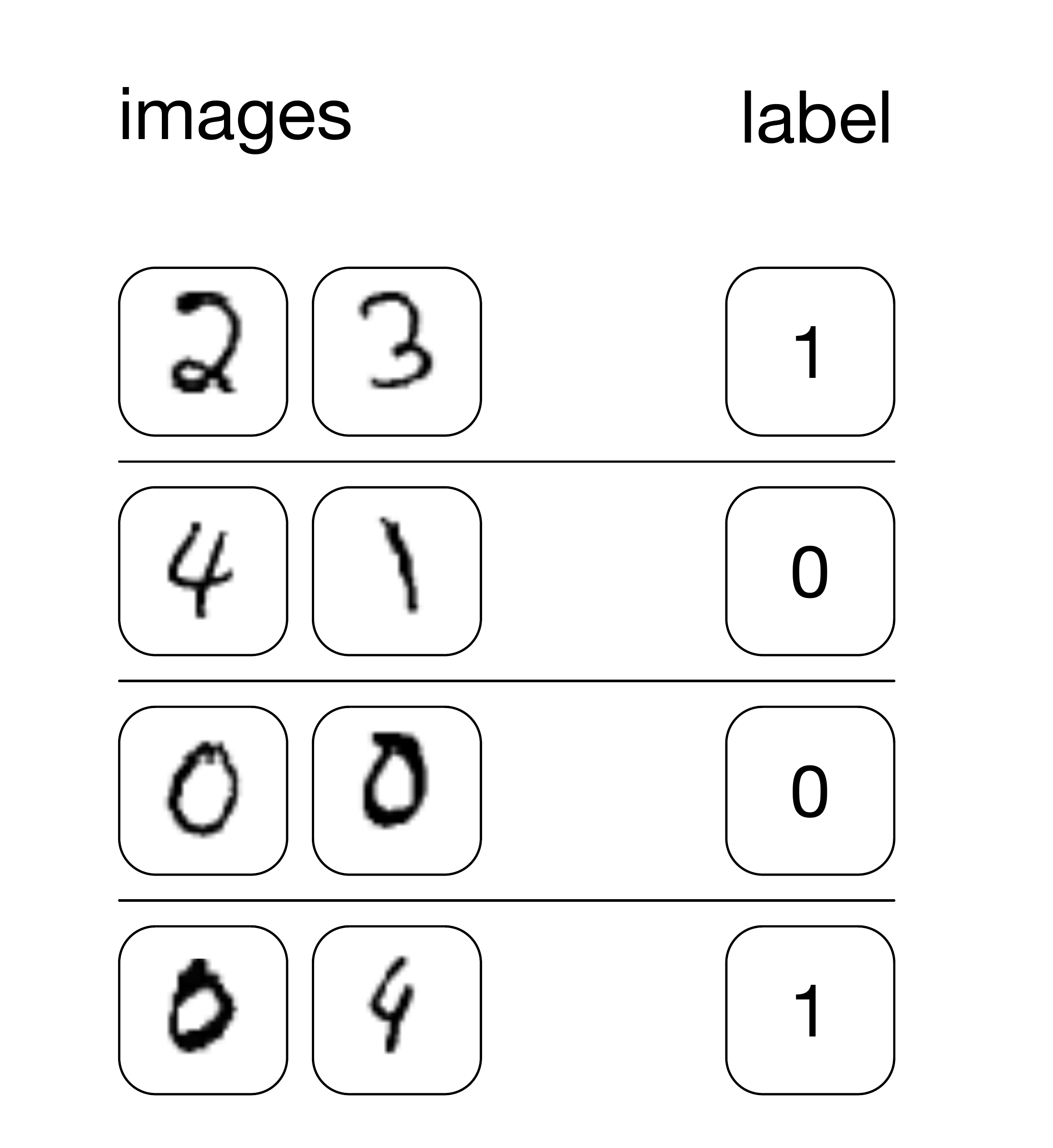}
\caption{Learning Less-Than from Raw Pixel Images}
\label{mnist-image-labels}
\end{figure}

In this task, the system is again given a pair of images each training step.
The training label is a 1 if the number represented by the left image is less than the number represented by the right image.
See Figure \ref{mnist-image-labels}.
The language for this task is:
\begin{itemize}
\item
$P_e: \{zero/1, succ/2, image1/1, image2/1\}$
\item
$P_i: \{target/0, pred1/1, pred2/2\}$
\item
$C = \{0, ..., 5\}$
\end{itemize}
This is another hard task that requires synthesising two auxiliary predicates, one of which is recursive.
One of the solutions that \system{} found is:
\begin{eqnarray*}
target() & \leftarrow & image2(X), pred1(X) \\
pred1(X) & \leftarrow & image1(Y), pred2(Y, X) \\
pred2(X, Y) & \leftarrow & succ(X, Y) \\
pred2(X, Y) & \leftarrow & pred2(Z, Y), pred2(X, Z)
\end{eqnarray*}
Here $pred2$ is the less-than relation.
\system{} found a number of other solutions, some of which are less human-readable, e.g.
\begin{eqnarray*}
target() & \leftarrow & image2(X), pred1(X) \\
pred1(X) & \leftarrow & succ(Y, X), pred2(X, Y) \\
pred2(X, Y) & \leftarrow & succ(Y, X), image1(Y) \\
pred2(X, Y) & \leftarrow & succ(Y, X), pred2(Y, Z)
\end{eqnarray*}
Here, $pred2$ is true of pairs of integers $(n+1, n)$ for $n$ greater than or equal to the left image.
It creates a ``ladder'' that starts at the left image and keeps going up, and then $target$ tests if the right image intersects with that ladder.

Finally, to test the sample-complexity of this approach, we ran the less-than experiment while holding out certain pairs of integers.
Please note that we are not just holding out pairs of \emph{images}. 
Rather, we are holding out pairs of \emph{integers}, and removing from training \emph{every} pair of images whose labels match that pair.

We built a simple MLP baseline to compare \system{} against.
In the baseline model, the logits for the two images are concatenated together, creating a vector of size 20.
This vector is then passed to a linear layer with ReLU activations.
Again, we train by minimising cross-entropy loss.
The MLP baseline is able to solve this straightforward task reliably when it is given every pair of images.
But as the number of hold-out pairs increases, the baseline performs increasingly poorly.
If the MLP model has seen that $2 < 3$ and that $3 < 4$, but has never seen any pair of images with labels $(2, 4)$, then it does not know what to do.

\begin{figure}[!ht]
  \subfloat[Proportion of successful runs\label{fig:hold-out-1}]{%
    \includegraphics[width=0.45\textwidth]{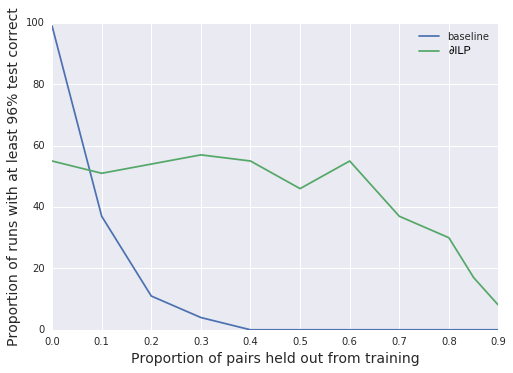}
  }
  \hfill
  \subfloat[Mean Error\label{fig:hold-out-2}]{%
    \includegraphics[width=0.45\textwidth]{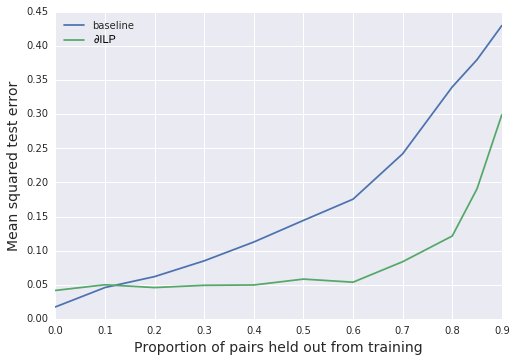}
  }
  
    \caption{Comparing \system{} with the MLP Baseline}
    \label{hold-out}
\end{figure}

We compare the baseline and \system{}'s performance in Figure \ref{hold-out}.
In the left figure, we plot the number of runs that get at least 96\% correct on the test data.
The reason we cap at 96\% is that our pre-trained MNIST classifier has a test accuracy of 98\%, and we are running it twice to classify two images, so the chance of it classifying both images correctly is $0.98 * 0.98$.
Because \system{}'s performance depends, in part, on the values of the initial random weights, it never achieves 100\% performance.
But please note that its performance is robust when holding out 70\% of the data.
Because of its strong inductive bias, \system{} is able to learn the correct generalisation from much fewer pairs of data.

In the right image, we plot the mean test error as we increase the hold-out proportion.
Note that \system{} maintains a respectable mean error even when 70\% of the pairs of integers are never seen.

\subsubsection{Discussion}

Our aim is to connect \system{} to a perceptual system, with the view of potentially training the whole system end-to-end: learning rules and perceptual classifiers \emph{at once}.
The initial experiments described above do not quite reach that goal, for two reasons.
First, we do not train the convnet at the same time as \system{}.
Rather we pre-train the convnet on MNIST classification, and then freeze the convnet weights.
Second, we provide a custom mapping from the convnet's logits to the initial sub-valuation. 
We applied our knowledge that there are two images, each of which describes a single digit, to bake in a particular mapping from the convnet's logits to the initial sub-valuation:
since we know that there is exactly one $X$ such that $image1(X)$, we apply soft-max to the logits when transforming to the sub-valuation.
While this does not provide information about the \emph{specific tasks} the system will deal with, this is nonetheless knowledge about the \emph{domain} the tasks form a part of. Ideally, this is knowledge which should be \emph{learned} by the system rather than baked in by the designers, and it is important to note that the framework provides no obstacles to this being achieved. In future work, we plan to develop an end-to-end differentiable system that learns rules and classifiers simultaneously, focussing on good initialisation through pretraining of the perceptual network.

In a further series of experiments, we no longer pass the logits from the pre-trained convnet directly to the sub-valuation (the mapping from $\{image(i)\}_{i=0}^5$ to probabilities).
Rather, we extract the final hidden layer of the convnet, and \emph{learn} how to convert it into a sub-valuation.
If the final hidden layer of the convnet is $\bf{h}$, then we compute the sub-valuation $\mathbf{v}$ as follows:
\[
\mathbf{v} = \mathsf{softmax} (A \mathbf{h} + B)
\]
where $A$ and $B$ are learned weights.
This way, the network learns the mapping from perceptual discriminations to symbolic assignments as part of the end-to-end learning process.
The model learns to discriminate the images of integers \emph{just as much as it needs to for the task at hand}.
For example when learning the $zero$ task (whether the image I am currently seeing represents the integer zero), it reliably maps images of zero to the integer 0, but all other images are mapped to a high entropy distribution. It ``blurs'' the difference between the other integers, as it does not need them.
In another example, when learning the $identity$ relation, it creates \emph{some} bijective mapping from classes of images to integers, but may map images of ``1'' to 2, and images of ``2'' to 1, etc.
In a third example, when learning the $predecessor$ relation on integers, it learned to map the image of $n$ to $9 - n$, so e.g. $0 \mapsto 9$ and $1 \mapsto 8$, and then it used $succ$ to implement predecessor directly.
We find these initial results intriguing, in that they serve to illustrate the arbitrariness of symbols in standing for concepts, and believe they show the promise of end-to-end learning that combines perception and reasoning.

\section{Related Work}
\label{sec:related}

We survey some recent related work involving neural or differentiable implementations of logic programming in general, and inductive logic programming in particular.
Early work in this area involved translating propositional logic programs into neural nets \cite{steinbach2002neural}, or learning a propositional logic program from a set of examples \cite{holldobler1991towards}.
More recently, nonclassical logics such as modal logics, non-monotonic propositional logics, and argumentation frameworks have been encoded in neural nets \cite{garcez2007abductive,garcez2008neural,d2009connectionist,garcez2014neural}.

The related work discussed below focuses on learning first-order rules.

\subsection{``Approximating the Semantics of Logic Programs by Recurrent Neural Networks''}
\label{approximating}

\citeA{holldobler1999approximating} show that, for every logic program in a certain class, there exists a recurrent net that can compute the forward chaining consequences of that program to any desired degree of precision.
Given a set $G$ of ground atoms and an injective function $|| \cdot || : G \rightarrow \mathbb{N}$ mapping atoms to integer indices, they define a mapping $R : 2^G \rightarrow \mathbb{R}$ from sets of ground atoms to the reals\footnote{There is nothing particularly special about the use of base $4$ in this expression. Any base $b > 2$ could have been used. To see that base 2 is not acceptable, consider an infinite set $\{g_1, g_2, ...\} $ of ground atoms. Now $R(\{g_1\}) = 0.1 = R(\{g_2, ...\}) = 0.0\overline{1}$, so our $R$ function fails to be injective. In any base $b$ greater than 2,  $0.1_b \neq 0.0\overline{1}_b$, so the function $R$ is injective, as desired.}:
\[
R(M) = \sum_{g \in M} 4^{- ||g||}
\]
Since  $|| \cdot ||$ is injective, $R$ has an inverse $R^{-1}$.
They define a single-step consequence operator: 
\[
T_P(M) = \{g \; | \; g \leftarrow g_1, ..., g_m \in \mathsf{ground}(P), \bigwedge_{i=1}^m g_i \in M\}
\]
Note that this is not quite the same as our $cn_P$, defined in Section \ref{sub:logicprogramming}, since $T_P(M)$ does not include the input atoms $M$.

Now, given a logic program $P$, they define a function $f_P : \mathbb{R} \rightarrow \mathbb{R}$ on reals that mirrors the forward chaining consequence function, as $T_P : 2^G \rightarrow 2^G$ on sets of ground atoms:
\[
f_P(x) = R (T_P(R^{-1}(x)))
\]
The above function is defined only on the range of $R$. They extend $f_P$ to a function on $\mathbb{R}$ by linear interpolation.
Now, given that $f_p$ as defined is continuous, they apply the well-known result that any continuous function on the reals can be approximated, to any degree of precision, by a three layer network.
Therefore, for every acyclic program, there exists a neural net that can compute forward chaining inference on that program to any desired degree of precision.

However, as the authors acknowledge, this instructive and revealing theoretical result does not give us a practical method for constructing neural nets for particular logic programs.
The particular encoding $R$ they use, mapping sets of atoms to individual real numbers, is not practical, as it assumes arbitrary precision floating point numbers. 

\subsection{``Connectionist Model Generation: A First-Order Approach''}

\citeA{bader2008connectionist} built upon the work described above by \citeA{holldobler1999approximating}.
Here, the authors describe an actual implementation based on the theory described above.

The impractical aspect of the work of \citeA{holldobler1999approximating} is the use of a \emph{single real number} to represent an entire set of ground atoms.
Now, to make this approach practical, they use a vector, rather than a single number, to represent a set of ground atoms.
They modify $|| . ||$ to be a 2-dimensional level mapping, assigning each ground atom a pair $(L, i)$ where $L$ is a level, as before, and $i$ is a position index in the distributed vector representation.

Similarly, they redefine $R : 2^G \rightarrow \mathbb{R}^k$ from sets of ground atoms to vectors of length $k$:
\begin{eqnarray*}
\iota(A) & = & (\iota_1(A), ..., \iota_k(A)) \\
\iota_i(A) & = & \begin{cases}
4^{-l} \; \mbox{if} \; ||A|| = (l,d) \; \mbox{and} \; i = d \\
0 \; \mbox{otherwise}
\end{cases} \\
R(M) & = & \sum_{A \in M} \iota(A) 
\end{eqnarray*}

The paper describes two key technical contributions.
The first is a program that converts an acyclic logic program $P$ into a neural net that computes $f_P$ for any desired level $n$.
To convert a logic program into a multi layer neural net, for a particular desired level $n$, they enumerate all possible models containing atoms that are of level less than or equal to $n$.
For each such model $M$, they compute $T_P(M)$, add a corresponding hidden unit $h$ and set the weights from $h$ to the output layer to $R(T_P(M))$.
In other words, there is a new hidden unit for every possible model $M$. This means there are $2^n$ hidden units added. 
Each hidden unit represents a particular combination of proposition atoms. 
The neural net effectively memorises how to map particular sets of propositional atoms to their immediate consequents.

The second major contribution is a technique for learning an acyclic program from examples.
To learn a set of weights that computes $f_P$, we sample models $M$ containing atoms of level $\leq n$.
Set the input layer of the net to the vector representation of the model:
\[
\mathsf{input} = R(M)
\]
Set the desired output layer $\mathbf{y}$ to the vector representation of the single-step consequence of $M$:
\[
\mathbf{y} = R(T_P(M))
\]
They update the weights from hidden units to output using the weight update rule with learning rate $\eta$:
\[
\mathbf{w}_{out} \leftarrow \eta \cdot \mathbf{y} + (1 - \eta) \cdot \mathbf{w}_{out}
\]

Their approach is similar to the approach taken in the current paper in that it involves a differentiable implementation of forward chaining.
But it is different in a number of respects.
First, they consider acyclic logic programs, while we consider Datalog programs. 
Acyclic programs can include function symbols, and hence infinite Herbrand universes, but do not include programs involving mutual recursion.
Datalog programs do not support function symbols, have finite Herbrand universes,  but do allow mutual recursion.
Second, once a net has learned to compute $f_P$, the program is hidden in the weights. 
It is not at all obvious how to extract an explicit symbolic logic program from these weights.
How to do so, if indeed it is possible, is an open research question.
Third, there is a fundamental difference in the way the clause is represented in the neural net.
In their system, each hidden unit in their system represents a hyper-square around a vector representation of a particular set of ground atoms.
Each hidden unit has a ``propositional fixation''.
In our system, by contrast, a universally quantified clause is applied simultaneously to every pair of atoms. 
The hidden units in our net encode the relative weight of a general rule that is applied repeatedly, while each hidden unit in their network encodes a particular set of ground atoms.

\subsection{``Logic Tensor Networks: Deep Learning and Logical Reasoning from Data and Knowledge''}
\label{deep-logic}
\citeA{logictensornetworks} provide a generalisation of the semantics of first-order logic, replacing the standard Boolean values with real values in $[0, 1]$.
In their ``Real Logic'', a constant $c$ is interpreted as a vector $G(c) \in \mathbb{R}^n$ of reals, a function $f$ is interpreted as a function $G(f) \in \mathbb{R}^{\mathsf{arity}(f) \times n} \rightarrow \mathbb{R}^n$ from tuples of vectors to vectors of reals, and a predicate $p$ is interpreted as a function $G(p) \in \mathbb{R}^{\mathsf{arity}(f) \times n} \rightarrow [0,1]$ from tuples of vectors to $[0,1]$.
Given these interpretations, they define satisfaction for complex ground formulae:
\begin{eqnarray*}
G(f(t_1, ..., t_m) & = & G(f)(G(t_1), ..., G(t_m)) \\
G(p(t_1, ..., t_m) & = & G(p)(G(t_1), ..., G(t_m)) \\
G(\neg \phi) & = & 1 - G(\phi) \\
G(\phi_1 \lor ... \lor phi_m) & = & \mu(G(\phi_1), ..., G(\phi_m)) 
\end{eqnarray*}
where $\mu$ is the dual of some t-norm operator. (See Section \ref{t-norms} above).

Note that this continuous semantics works for full first-order logic. They are not restricted to Horn clauses in general or to function-free Datalog clauses in particular. 

The major difference between this approach and ours is that theirs is a form of continuous abduction, rather than induction. 
Given an assignment $\hat{G}$ from atoms to $[0,1]$, and a set $R$ of first-order statements, they find an extension $G \supseteq \hat{G}$ such that $G$ minimises the discrepancy between the given (continuous) truth-values assigned to the ground instances of the first-order statements in $R$ and the calculated truth-values assigned to the ground instances of $R$ using the continuous satisfaction rules above.
This way, they can ``fill in'' facts that were not given, so as to maximize the truth-values of the first order rules.
This is a form of \emph{knowledge completion} over fuzzy valuations.

\subsection{``Computing First-Order Logic Programs by Fibring Artificial Neural Networks''}

\citeA{bader2005computing} describe a way of implementing the single-step operator $T_P$ (described in Section \ref{approximating} above) using fibred neural networks.
This approach can represent recursive programs, and even mutually recursive programs
(e.g. the mutual definition of even and odd described in Appendix \ref{even-odd}).
However, this system cannot represent programs with free variables: variables that are existentially quantified in the body of a rule that do not appear in the head of the rule.
 
\subsection{``Logical Hidden Markov Models''}

The state transitions of a Hidden Markov Model (HMM) can be viewed as encoding a set of propositional rules, each weighted with a probability.
\citeA{kersting2006logical} lift HMMs from a propositional to a first-order representation, and call the resulting structure a Logical HMM (LOHMM).

A rule in LOHMM is of the form:
\[
p: h(X,Y) \overset{a(X)}{\longleftarrow} b(X)
\]
where $p$ is a probability; $h$, $a$, and $b$ are predicates; $X$ is a vector of universally quantified variables and constants, and $Y$ is a vector of existentially quantified variables appearing only in the head.
This rule reads as: for all $X$, if $b(X)$ holds then, with probability $p$, emit atom $a(X)$ and transition to a state $h(X,Y)$ for some value of $Y$.

In their paper, they show how the rule probabilities $p$ can be learned from sets of atom sequences, using a modified version of the Baum-Welch algorithm.
Using this technique, they are able to learn first-order rules from sequences of data.
Experimental results show that this approach produces smaller models and better generalisation than traditional HMMs.

This approach is very similar to ours in its aim: we both want to learn first-order rules from noisy data by maximising a conditional probability.
One difference is that they are learning rules from \emph{sequences} of atoms.
The background assumption behind their approach is that exactly one atom is true at a time:
a set of hidden states emit a sequence of atoms, and we wish to learn the hidden states, or the rules that transition between the states.
In our \system{} approach, by contrast, the state of the world at any time is modelled by a \emph{set} of atoms, not by a single atom.

Another difference between our approaches is that LOHMM rules have a single atom in the body, while \system{} rules have multiple atoms in the body. 
LOHMM rules are equivalent to the rules of linear $Datalog^{\exists}$ \cite{cali2012general}.

\subsection{``Probabilistic Inductive Logic Programming''}

\citeA{de2008probabilistic} make a strong case for the importance of applying relational learning to uncertain noisy data. 
They provide a very general formulation of ILP using a $covers$ relation. 
This general formulation encompasses ILP learning from entailment (this is the formulation of ILP used in our paper), learning from interpretations, and learning from proofs.
They generalise this $covers$ relation to the probabilistic setting by reinterpreting it as a conditional probability.
If $e$ is an example, $R$ is a set of rules and $\mathcal{B}$ is a background theory, they define $covers(e,R,\mathcal{B})$ as
\[
covers(e, R, \mathcal{B}) = P(e | R, \mathcal{B})
\]
This is similar to our probabilistic formulation in Section \ref{subsec:differentiable-induction} above.

They describe nFOIL \cite{landwehr2005nfoil}, a probabilistic variant of FOIL \cite{quinlan1995induction} using naive Bayes. 
This system is able to learn individual clauses in the presence of noisy data, and outperformed mFoil and Aleph on the benchmark Mutagenesis dataset.
nFOIL is certainly similar in approach to \system{}.
One difference between their system and ours is that \system{} learns multiple clauses at once, including multiple recursive clauses, while FOIL and its variants are not able to learn recursive clauses\footnote{But see the work of \citeA{natarajan2012gradient} and \citeA{khot2015gradient} for systems that can learn multiple clauses at once. See the work of \citeA{raedt2016statistical} for an overview of Statistical Relational Learning.}. 

\subsection{``First-order Logic Learning in Artificial Neural Networks''}

\citeA{guillame2010first} describe an approach to learning definite clauses which is similar, in various ways, to our approach. 
In their approach, just like ours, the architecture of the network is designed to implement forward chaining on definite clauses.

Although our approach is similar to theirs in overall approach, there are a number of differences in the details.
First, the type of clause they learn is different.
They do not make the Datalog restriction, and consider definite clauses with function symbols. 
Second, their system is only able to learn a single clause, rather than a set of clauses.
Third, they do not support recursive clauses.

\subsection{``Fast Relational Learning Using Bottom Clause Propositionalization with Artificial Neural Networks''}

\citeA{francca2014fast} describe a powerful approach for learning first-order logic programs efficiently.
They use the bottom clauses $\bot_e$ introduced by \citeA{muggleton1995inverse}.
For each positive example $e \in \mathcal{P}$, the bottom clause $\bot_e$ is the most specific clause whose head unifies with $e$.
Since the clauses form a lattice (with a partial ordering defined by theta-subsumption), there is always a most specific clause for a particular language bias. 
For example, if $target$ is a binary predicate, the bottom clause $\bot_e$ for $target(a,b)$ might be
\[
target(X, Y) \leftarrow p(X), q(Y), r(X, X), s(Y), t(Y, X)
\]
For each negative example $e \in \mathcal{N}$, we find the most specific clause whose head unifies with $\neg e$ (where $\neg$ is classical negation).
For example, the bottom clause for a negative example $target(c,d) \in \mathcal{N}$ might be:
\[
\neg target(X, Y) \leftarrow q(X), p(Y), r(Y, X), t(X, X)
\]
A bottom clause is typically rather long: it is, after all, the most specific clause (in the finite set of clauses specified by the language bias) that captures the facts about the example
\footnote{For a thorough discussion, see the work of \citeA{tamaddoni2009lattice}.}.

\citeA{francca2014fast}  use the set of bottom clauses generated from the examples as a set of features: each clause is applied to every tuple of individuals, generating a list of features for each tuple.

\citeA{francca2014fast} take the bottom clauses from the positive and negative examples, ground them for all tuples, and then pass the ground bottom clauses to CILP++. 
Then, they use the ground clauses to generate a neural net, and learn the appropriate weights.

This system has a number of appealing features.
First, it learns first-order rules rather than just propositional rules.
Second, it is very efficient.

Comparing CILP++ with \system{}, both systems have their advantages and disadvantages. 
We note first that CILP++ is unable to learn recursive clauses, since the predicates in the body of the bottom clauses $\bot_e$ are always extensional predicates.
\system{}, by contrast, is able to learn complex recursive programs, sometimes involving mutual recursion. See Appendix \ref{all_experiments}.
On the other hand, \system{} is currently restricted to small datasets, because of its memory requirements. 
So it cannot be tested on large datasets such as Mutagenesis, KRK, UW-CSE. 
\system{} and CILP++ occupy different positions in the space of design tradeoffs: \system{} can learn more complex programs, while CILP++ can handle larger amounts of data.

\subsection{``Learning Knowledge Base Inference with Neural Theorem Provers''}

\citeA{rocktaschel2016learning} describe a differentiable neural net implementation of a backward chaining inference procedure.
From the examples they give, it looks like they are also operating on the Datalog subset of logic programs, although they do not explicitly say this.

The Neural Theorem Prover is handed a set of predefined \define{clause templates}: clauses in which some elements (including predicates) are wild cards, e.g.
\[
\#1(X, Z) \leftarrow \#2(X, Y ) \land \#2(Y, Z)
\]
This template has already specified the variable bindings.
Notice also the wild-card predicate matching in the above template. 
It insists that the same predicate ($\#2$) is used for both atoms in the body of the rule.
These templates are similar to the mode declarations used by \citeA{corapi2011inductive}, and are a more restrictive form of rule template than the $(v, int)$ templates used in \system{}.

Given a set of clause templates, and a goal, they produce a neural net model that performs backwards chaining, starting with the goal, working backwards using the rules to find a set of substitutions that makes the goal true.
They keep track of substitutions as vectors of vectors, and keep track of the plausibility of a particular substitution as a real number in $[0,1]$.
The Neural Theorem Prover uses a distributed representation of atoms: each atom is represented by a triple of vectors. 
The vector for each symbol is a distributed latent representation.  
One thing that is particularly interesting about this system is that they learn clauses and vector embeddings \emph{simultaneously}. 

At a high level, the Neural Theorem Prover is similar to our work: they provide a differentiable model of inference on Datalog clauses, using a distributed representation of atoms.
The main difference between their approach and our is that their model uses backward chaining inference, keeping track of the various proof paths and substitutions, while our approach uses forward chaining.

\subsection{``A Differentiable Approach to Inductive Logic Programming''}

\citeA{yangdifferentiable} outline a system that learns clauses by a differentiable implementation of forward-chaining. 
They apply their model to the  European royal family dataset \cite{wang2015soft}, and learn rules.
Some of our baseline ILP tasks are borrowed from this model.
There are two differences between their approach and the one described here.
First, they only learn individual clauses, not sets of clauses.
Second, their system does not learn recursive clauses.

\subsection{``Dyna: A Declarative Language for Implementing Dynamic Programs''}

\citeA{eisner2004dyna} describe a high level declarative first-order language for specifying Natural Language Processing systems.
Starting with an initial assignment of probabilities to axioms, the system runs forward-chaining inference to deduce further probabilities,
concluding with the probability of a certain parse of a sentence.
The system can also be run in reverse, to maximise the probability of the input sentence.
In this reverse mode, the weights of the axioms are updated (using back-propagation) to maximise the probability of the training data.

The central difference between \system{} and Dyna is the difference between induction and abduction.
In \system{}, we are given a set $\mathcal{B}$ of background rules, and we use gradient descent to learn the weights of the rules $R$.
In Dyna, by contrast, the set of rules $R$ is fixed, and they use gradient descent to learn the weights of the background assumptions $\mathcal{B}$.

\section{Discussion}

We conclude by discussing the strengths and weaknesses of our approach.

\subsection{ Program Templates and Language Bias}
\label{iterative-deepening}

A common complaint levelled at Inductive Logic Programming, and at Inductive Program Synthesis more generally, is that the system needs to be given a program template (also known as ``language-bias'', ``mode declarations'', or ``metarules''), carefully hand-engineered by a human to match the task at hand, in order for the system to have any chance of finding the desired program.
A lot of the difficult work has already been done by the choice of program template, and all the program synthesis system needs to do is fill in the blanks.

The strength of this complaint depends on the amount of information in the program template. 
If the program template specifies exactly the predicates or functions to be called, then it is very restrictive indeed.
The rule templates used in \system{} are not so restrictive; we specify the number of existential variables and whether the clause may or must use intensional predicates, but we do not specify the form of the recursive call. 
See Appendix \ref{corapi_comparison} and Appendix \ref{metagol_comparison} for a comparison with other forms of language bias.

Nevertheless, it would be clearly preferable to avoid the need for program templates at all.
We believe that the need for program templates is the key weakness in existing ILP systems, ours included.
We experimented with a brute-force way
 to avoid hand-engineering rule templates by using \emph{iterative deepening}.
Suppose we have a way of enumerating program templates, starting with the most minimal, in such a way that all program templates will eventually be produced.
To do this, we used Cantor's pairing function, applied recursively, generating sequences of program templates.
Given such an enumeration, we searched through the space of programs without needing any hand-engineered rule templates, by running the different program templates in batch over a large cluster of machines. 
In this brute-force approach, we have, of course, only avoided hand-engineering rule templates by resorting to using a massive amount of computation.

\subsection{Conclusion}
\label{sec:conclusion}

Our main contribution is a differentiable implementation of Inductive Logic Programming.
This model is able to learn moderately complex programs: programs requiring multiple invented predicates and recursive clauses.
Unlike traditional symbolic ILP systems, our model is robust to mislabelled training data.
Futhermore, unlike traditional symbolic ILP systems, our model can handle ambiguous data in domains where artificial neural networks are traditionally the tool of choice. Unlike neural networks on their own, it offers the same data efficiency and generalisation guarantees as traditional ILP systems, extended to new domains. In doing so, this work opens the door to an end-to-end differentiable system that is capable of learning perceptual and inferential rules \emph{simultaneously}.

We evaluated this model on 20 symbolic ILP tasks, and showed that it can consistently solve problems traditional ILP systems excel on. 
We tested its tolerance to mislabelled data, and demonstrated it was able to learn good models even with 20\% mislabelled training examples.
We also tested it with ambiguous data, connecting the output of a pretrained convnet to the inputs of the induction system, and it was able to learn effectively and consistently generalise.

There are two main limitations of our framework that we wish to address in future work.
First, we currently use program templates to constrain the set of programs that are searched through.
We have experimented with avoiding the use of hand-engineered program templates,
by searching through the space of program templates using iterative deepening, as outlined in Section~\ref{iterative-deepening}.
But this approach is hugely computationally expensive.
In future work, we plan to adopt a less brute-force solution to this problem.

Second, the experiments on ambiguous data relied on pre-trained neural network classifiers.
We ``baked in'' the knowledge that the MNIST images were clustered into ten classes corresponding to the ten digits. Preliminary experiments showed that pre-training only the convolutional layers of a vision network, and jointly learning the projection into symbols and reasoning over such symbols, yielded satisfactory and interpretable results for our MNIST-based tasks, but full end-to-end joint training of vision modules and \system{} requires further research. Future work will focus on sensible pretraining strategies for these neural networks, which do not obviously bias them towards the specific problems into which they will be integrated.

\acks{We thank Nando de Freitas, Shane Legg, Demis Hassabis, Phil Blunsom, Chris Dyer, Karl Moritz Hermann, David Pfau, Chrisantha Fernando, Joel Veness, and Marek Sergot for their helpful comments and suggestions. We are grateful to Andrew Cropper and Stephen Muggleton for making Metagol available online; thanks to Andrew for his patience in explaining how to use it. We thank the reviewers for their thoughtful comments. }

\appendix

\section{Notational Conventions}
\label{notation}

The notational conventions used throughout this paper are summarised in Tables~\ref{fig:noteconv} and~\ref{fig:noteconv2}. 
We use $\mathcal{A}, \mathcal{B}, ...$ to denote sets of ground atoms e.g. $\mathcal{A} = \{p(a,b), p(b, c)\}$.
We use $p, q, ...$ to denote predicates, $a, b, ...$ to denote constants, and $X, Y, ...$ to denote variables.
Italic lowercase e.g. $a, b, ...$ denote scalars.
Boldface lowercase e.g. $\mathbf{a}, \mathbf{b}, ...$ denote vectors.
Boldface uppercase e.g. $\mathbf{A}, \mathbf{B}, ...$ denote matrices and tensors.
The $i$'th element of a vector $\mathbf{x}$ is $\mathbf{x}[i]$.
The element at row $i$ and column $j$ in a matrix $\mathbf{A}$ is $\mathbf{A}[i, j]$.
The $i$'th row of a matrix $\mathbf{A}$ is $\mathbf{A}[i, :]$. The $j$'th column is $\mathbf{A}[:, j]$. If $\mathbf{X}$ is a tensor of shape $(k, m, n)$ then $\mathbf{X}[i, :, :]$ is the $i$'th matrix of shape $(m, n)$.
We use $\times$ for scalar multiplication (and also for cross-product), $\odot$ for component-wise multiplication, and $\cdot$ for dot product.

\begin{table}[htbp]
\centering
\small
\begin{tabular}{lp{11cm}}
\toprule
{\bf Term} & {\bf Explanation} \\ \midrule
$p(X) \leftarrow q(X,Y)$ & A definite clause. This clause states that, for all $X$, if there is a $Y$ such that $q(X,Y)$, then $p(X)$ \\
$\mathcal{L}$ & A language frame: a tuple consisting of a target, a set of extensional predicates $P_e$ with associated arities, and a set $C$ of constants \\
$target$ & The target predicate that the ILP system is trying to learn \\
$P_e$ & A set of extensional predicates \\
$arity$ & A map $P \rightarrow \mathbb{N}$ from predicates to arities \\
$C$ & A set of constants \\
$\mathcal{A}_L$ &  The set of all atoms (both ground and unground) that can be generated from language $L$ \\ 
$\alpha$ & A particular atom (ground or unground) \\
$\mathcal{G}$ & The set of all ground atoms \\
$n$ &The size of G \\
$\gamma$ &  A particular ground atom \\
$\perp$ & Falsum, the atom that is always false \\
$\mathcal{B}$ &  The set of background assumptions \\
$\mathcal{P}$ &  The set of positive examples of the target predicate \\
$\mathcal{N}$ &  The set of negative examples of the target predicate \\
$\tau$ & A rule template, a tuple $(v, int)$ \\
$v$ & The number of existentially quantified variables allowed by a rule template \\
$int$ & Whether intensional predicates are allowed in clauses \\ 
$\Pi$ & A program template: a tuple $(P_a, arity_a, rules, T)$ \\ 
$P_a$ & A set of auxiliary (invented) predicates \\ 
$P_i$ & The set of intensional predicates: $P_i = \{target\} \cup P_a$ \\ 
$rules$ & A map from intensional predicates $p \in P_i$ to pairs of rule templates $\tau_p^1, \tau_p^2$ \\ 
$P$ & The set of all predicates in the language $P_e \cup P_i$ \\ 
$T$ & The max number of time-steps of forward chaining inference permitted by the program template \\ 
$L$ & A language $(P_e, P_i, arity, C)$ combining the extensional predicates $P_e$ from a language frame with the intensional predicates $P_a$ from the program template \\ 
$\tau_p^i$ & The $i$'th rule template for intensional predicate $p$, $i \in \{1,2\}$ \\ 
$C_L$ &  The set of all definite clauses that can be generated from language $L$ \\ 
$R$ &  A set of definite clauses, a program, a candidate solution to an ILP problem\\ 
$cl(\tau)$ & The set of clauses that can be generated from rule template $\tau$ \\ 
$C_p^{i,j}$ & The $j$'th clause generated by the $i$'th rule template for predicate $p$ \\ 
$\Phi$ & The set $\Phi = \{f_p^{i,j}\}_{p \in P_i}^{i \in \{1,2\}, j = 1 .. |cl(\tau_p^i)|}$ of Boolean variables indicating whether to use the $j$'th clause in the $i$'th rule template for $p$ \\ 
$cl^*(\tau_p^i)$ & The set of generated clauses for $\tau_p^i$, modified by adding the corresponding $f_p^{i,j}$ flags, for $j = 1 .. |cl(\tau_p^i)|$ \\ 
$F \subset \Phi$ & A subset of flags specifying a particular set of clauses \\ 
\bottomrule
\end{tabular}
\caption{Notational Conventions used throughout this paper.}
\label{fig:noteconv}
\end{table}

\begin{table}[htbp]
\centering
\small
\begin{tabular}{lp{10cm}}
\toprule
{\bf Term} & {\bf Explanation} \\ \midrule
$\Lambda$ & A set of atom-label pairs \\
$\lambda$ & A label for a particular ground atom $\gamma$ \\
$W$ & A set of weight matrices $W = \{\mathbf{W}_p | p \in P_i\}$ \\
$\mathbf{W}_p \in \mathbb{R}^{|cl(\tau_p^1)| \times |cl(\tau_p^2)|}$ & The weight matrix for a particular intensional predicate $p$, determining the probability of every pair $c_1, c_2$ of clauses satisfying rule templates $\tau_p^1$ and $\tau_p^2$ \\
$\mathbf{a}, \mathbf{b}, \mathbf{c}, \mathbf{x}$ & Valuation vectors in $[0,1]^n$ \\
$\mathcal{F}_c: [0,1]^n \rightarrow [0,1]^n$ & The valuation function performing one step of forward inference corresponding to clause $c$ \\
$\mathcal{F}_p^{i,j}: [0,1]^n \rightarrow [0,1]^n$ & The valuation function corresponding to clause $C_p^{i,j}$ \\
$\mathcal{G}_p^{j,k}: [0,1]^n \rightarrow [0,1]^n$ & The valuation function corresponding to one step of forward inference using both clause $C_p^{1,j}$ \emph{and} clause $C_p^{2,k}$ \\
$\theta$ & A substitution mapping free variables to constants \\ 
$\alpha [\theta]$ & The result of applying substitution $\theta$ to atom $\alpha$ \\
\bottomrule
\end{tabular}
\caption{Notational Conventions used throughout this paper.}
\label{fig:noteconv2}
\end{table}

\section{ILP as Satisfiability}
\label{sat}

This elaborates the discussion in Section \ref{reducing-to-sat} above.
We shall consider the following simple example throughout this section.
Let our language-frame be \[\mathcal{L} = (target, P_e, arity_e, C)\] where
\begin{itemize}
\item
$target = q/2$
\item
$P_e = \{p/2\}$
\item
$C = \{a,b,c,d\})$
\end{itemize}
\noindent
We follow the logic programming convention of placing the arity of the predicate after the predicate name, so $q/2$ states that $\mathsf{arity}(q) = 2$.

We shall consider the following ILP problem $(\mathcal{L}, \mathcal{B}, \mathcal{P}, \mathcal{N})$, with background facts:
\[
\mathcal{B} = \{p(a, b), p(b, c), p(c, d)\}
\]
The positive examples $\mathcal{P}$ are 
\[
\mathcal{P} = \{q(a, b), q(a, c), q(a, d), q(b, c), q(b, d), q(c, d)\}
\]
The negative examples $\mathcal{N}$ are
\[
\mathcal{N} = \{q(X, Y) \; | \; (X,Y) \in \{a, b, c, d\}^2, q(X,Y) \notin \mathcal{P} \}
\]
We shall consider two rule templates for $q$. 
Template $\tau_q^1$ is $(v=0, int=0)$, specifying no existentially quantified variables and disallowing intensional predicates in the body.
Template $\tau_q^2$ is $(v=1, int=1)$, specifying one existentially quantified variable and requiring an intensional predicate in the body.

For this simple problem, we use the program template $\Pi = (P_a, arity_a, rules, T)$ where:
\begin{itemize}
\item
$P_a = \{\}$
\item
$arity_a = \{\}$
\item
$rules = \{\tau_q^1, \tau_q^2\}$
\item
$T = 3$ specifying 3 steps of forward chaining
\end{itemize}
Here, there are no auxiliary predicates needed. 
This ILP problem is sufficiently simple that it can be solved without additional invented predicates.
Most of the benchmark ILP problems described below in Section~\ref{sub:ilpbenchmark} require at least one invented predicate.
When there are no auxiliary predicates, the only intensional predicate that needs to be assigned rule templates is the target predicate, $q$.

Suppose template $\tau_q^1$ for predicate $q$ is $(v=0, int=0)$, specifying no existentially quantified variables and disallowing intensional predicates in the body.
There are only 8 generated clauses for $\tau_q^1$ remaining after pruning:
\begin{multicols}{2}
\begin{enumerate}
\item
$q(X, Y) \leftarrow p(X, X), p(X, Y)$
\item
$q(X, Y) \leftarrow p(X, X), p(Y, X)$
\item
$q(X, Y) \leftarrow p(X, X), p(Y, Y)$
\item
$q(X, Y) \leftarrow p(X, Y), p(X, Y)$
\item
$q(X, Y) \leftarrow p(X, Y), p(Y, X)$
\item
$q(X, Y) \leftarrow p(X, Y), p(Y, Y)$
\item
$q(X, Y) \leftarrow p(Y, X), p(Y, X)$
\item
$q(X, Y) \leftarrow p(Y, X), p(Y, Y)$
\end{enumerate}
\end{multicols}
Note that some of the clauses have the same atom repeated twice in the body. 
For example, clause 4:
\[
q(X, Y) \leftarrow p(X, Y), p(X, Y)
\]
This clause is equivalent to:
\[
q(X, Y) \leftarrow p(X, Y)
\]
The repetition of atoms is not a mistake. The clauses are generated automatically, and we insist that all clauses have exactly two atoms in the body.
This insistence on a uniform clause size will come in useful when we transfer to the matrix-based approach in Section~\ref{sec:continuous}.

Suppose template $\tau_q^2$ for $q$ is $(v=1, int=1)$, specifying one existentially quantified variable and requiring an intensional predicate in the body.
There are 58 generated clauses for $\tau_q^2$ remaining after pruning, of which the first 16 are:
\begin{multicols}{2}
\begin{enumerate}
\item
$q(X, Y) \leftarrow p(X, X), q(Y, X)$
\item
$q(X, Y) \leftarrow p(X, X), q(Y, Y)$
\item
$q(X, Y) \leftarrow p(X, X), q(Y, Z)$
\item
$q(X, Y) \leftarrow p(X, X), q(Z, Y)$
\item
$q(X, Y) \leftarrow p(X, Y), q(X, X)$
\item
$q(X, Y) \leftarrow p(X, Y), q(X, Z)$
\item
$q(X, Y) \leftarrow p(X, Y), q(Y, X)$
\item
$q(X, Y) \leftarrow p(X, Y), q(Y, Y)$
\item
$q(X, Y) \leftarrow p(X, Y), q(Y, Z)$
\item
$q(X, Y) \leftarrow p(X, Y), q(Z, X)$
\item
$q(X, Y) \leftarrow p(X, Y), q(Z, Y)$
\item
$q(X, Y) \leftarrow p(X, Y), q(Z, Z)$
\item
$q(X, Y) \leftarrow p(X, Z), q(Y, X)$
\item
$q(X, Y) \leftarrow p(X, Z), q(Y, Y)$
\item
$q(X, Y) \leftarrow p(X, Z), q(Y, Z)$
\item
$q(X, Y) \leftarrow p(X, Z), q(Z, Y)$
\end{enumerate}
\end{multicols}

Given a program template $\Pi$, let $\tau_p^i$ be the $i$'th rule template for intensional predicate $p$, where $i \in \{1, 2\}$ indicates which of the two rule templates we are considering for $p$.
Let $C_p^{i, j}$ be the $j'th$ clause in $cl(\tau_p^i)$, the set of clauses generated for template $\tau_p^i$ .

To turn the induction problem into a satisfiability problem, we define a set $\Phi$ of Boolean variables (i.e. atoms with nullary predicates) indicating whether the various generated clauses are actually to be used in our program $R$:
\[
\Phi = \{f_p^{i,j}\}_{p \in P_i}^{i \in \{1,2\}, j = 1 .. |cl(\tau_p^i)|}
\]
Here, $f_p^{i, j}$ indicates whether the $j$'th clause of $C_p^i$ is used in our program $R$:
\[
f_p^{i,j} \; \text{iff} \; C_p^{i,j} \in R
\]
A candidate solution to an ILP problem is a subset $F \subset \Phi$ of these atoms indicating which clauses to turn on.
We insist that exactly one flag is turned on for each rule template $\tau_p^i$:
\begin{eqnarray*}
\forall p \in P_i \; \forall i \in \{1,2\} \; \exists ! j \in \{1 .. |cl(\tau_p^i)|\} \; f_p^{i,j} \in F
\end{eqnarray*}

To transform the induction problem into a satisfiability problem, we modify our generated clauses by adding a Boolean flag to each clause, so the $j$'th clause of $cl(\tau_p^i)$ now tests whether $f_p^{i,j}$ is true. 
Let $cl^*(\tau_p^i)$ be the modified set of clauses that include the appropriate flags.

Returning to our example, the Boolean variables (i.e. atoms using nullary predicates) are $f_q^{1,1}, ..., f_q^{1,8}$ for $\tau_q^1$ and $f_q^{2,1}, ..., f_q^{2,58}$ for $\tau_q^2$.
\noindent
The modified clauses $cl^*(\tau_q^1)$ are:
\begin{multicols}{2}
\begin{enumerate}
\item
$q(X, Y) \leftarrow p(X, X), p(X, Y), f_q^{1,1}$
\item
$q(X, Y) \leftarrow p(X, X), p(Y, X), f_q^{1,2}$
\item
$q(X, Y) \leftarrow p(X, X), p(Y, Y), f_q^{1,3}$
\item
$q(X, Y) \leftarrow p(X, Y), p(X, Y), f_q^{1,4}$
\item
$q(X, Y) \leftarrow p(X, Y), p(Y, X), f_q^{1,5}$
\item
$q(X, Y) \leftarrow p(X, Y), p(Y, Y), f_q^{1,6}$
\item
$q(X, Y) \leftarrow p(Y, X), p(Y, X), f_q^{1,7}$
\item
$q(X, Y) \leftarrow p(Y, X), p(Y, Y), f_q^{1,8}$
\end{enumerate}
\end{multicols}
The modified clauses $cl^*(\tau_q^2)$ are:
\begin{multicols}{2}
\begin{enumerate}
\item
$q(X, Y) \leftarrow p(X, X), q(Y, X), f_q^{2,1}$
\item
$q(X, Y) \leftarrow p(X, X), q(Y, Y), f_q^{2,2}$
\item
$q(X, Y) \leftarrow p(X, X), q(Y, Z), f_q^{2,3}$
\item
$q(X, Y) \leftarrow p(X, X), q(Z, Y), f_q^{2,4}$
\item
$q(X, Y) \leftarrow p(X, Y), q(X, X), f_q^{2,5}$
\item
$q(X, Y) \leftarrow p(X, Y), q(X, Z), f_q^{2,6}$
\item
$q(X, Y) \leftarrow p(X, Y), q(Y, X), f_q^{2,7}$
\item
$q(X, Y) \leftarrow p(X, Y), q(Y, Y), f_q^{2,8}$
\item
$q(X, Y) \leftarrow p(X, Y), q(Y, Z), f_q^{2,9}$
\item
$q(X, Y) \leftarrow p(X, Y), q(Z, X), f_q^{2,10}$
\item
$q(X, Y) \leftarrow p(X, Y), q(Z, Y), f_q^{2,11}$
\item
$q(X, Y) \leftarrow p(X, Y), q(Z, Z), f_q^{2,12}$
\item
$q(X, Y) \leftarrow p(X, Z), q(Y, X), f_q^{2,13}$
\item
$q(X, Y) \leftarrow p(X, Z), q(Y, Y), f_q^{2,14}$
\item
$q(X, Y) \leftarrow p(X, Z), q(Y, Z), f_q^{2,15}$
\item
$q(X, Y) \leftarrow p(X, Z), q(Z, Y), f_q^{2,16}$
\end{enumerate}
\end{multicols}
Recall the ILP problem. Given a set $\mathcal{B}$ of background assumptions, and sets $\mathcal{P}$ and $\mathcal{N}$ of positive and negative examples, we wish to find a set $R$ of definite clauses such that
\begin{eqnarray*}
\mathcal{B}, R \models \gamma \; \text{for all} \; \gamma \in \mathcal{P} \\
\mathcal{B}, R \nvDash \gamma\; \text{for all} \; \gamma \in \mathcal{N} 
\end{eqnarray*}
We recast this as a satisfiability problem.
Find a set $F \subset \Phi$ of Booleans such that:
\begin{eqnarray*}
\forall p \in P_i \; \forall i \in \{1,2\} \; \exists ! j \in \{1 .. |cl(\tau_p^i)|\} \; f_p^{i,j} \in F \\
\mathcal{B} \cup F, \bigcup_{p \in P_i, i \in \{1,2\}} cl^*(\tau_p^i) \models \gamma \; \text{for all} \; \gamma \in \mathcal{P} \\
\mathcal{B} \cup F, \bigcup_{p \in P_i, i \in \{1,2\}} cl^*(\tau_p^i) \nvDash \gamma \; \text{for all} \; \gamma \in \mathcal{N} \\
\end{eqnarray*}
This approach transforms the induction problem (find a set $R$ of rules) into an satisfiability problem (find a subset $F \subseteq \Phi$ of atoms).

If we find a set $F$ of flags satisfying these constraints, then we can extract the logic program $R$:
\[
R = \{C_p^{i,j} \; | \; f_p^{i,j} \in F\}
\]
Returning to the running example from Section \ref{reducing-to-sat}, suppose our background facts $\mathcal{B}$ are
\[
\mathcal{B} = \{p(a, b), p(b, c), p(c, d)\}
\]
Suppose our positive examples $\mathcal{P}$ are 
\[
\mathcal{P} = \{q(a, b), q(a, c), q(a, d), q(b, c), q(b, d), q(c, d)\}
\]
and our negative examples $\mathcal{N}$ are
\[
\mathcal{N} = \{q(X, Y) \; | \; (X,Y) \in \{a, b, c, d\}^2, q(X,Y) \notin \mathcal{P} \}
\]
Then one set $F$ satisfying the constraints is
\[
F = \{f_q^{1,4}, f_q^{2,16}\}
\]
We can extract a logic program $R$ from these clauses:
\[
R = \{C_q^{1,4}, C_q^{2,16}\}
\]
In other words, $R$ is:
\begin{eqnarray*}
q(X, Y) \leftarrow p(X, Y), p(X, Y) \\
q(X, Y) \leftarrow p(X, Z), q(Z, Y) 
\end{eqnarray*}
This solution defines $q$ as the transitive closure of $p$.

\section{Comparing Language Bias in $\partial$ILP with Mode Declarations}
\label{corapi_comparison}

Many ILP systems \footnote{E.g., Progol \cite{muggleton1995inverse} or ASPAL \cite{corapi2011inductive}.} specify a language bias (a range of admissible programs) by using a set of mode declarations. 
A \define{mode declaration} is an atom $p(t_1, ..., t_n)$ where each $t_i$ is a placemarker indicating the type and grounding of the term that can appear in this position.
Each mode declaration is either a head declaration ($modeh$) specifying the form of the head of a rule, or a body declaration ($modeb$) specifying the form of one of the literals in the body of a rule.

For example, the mode declarations for the $even/succ2$ example in Section \ref{subsubsec:even-succ2} are:
\begin{eqnarray*}
m1 & : & modeh(even(+nat)) \\
m2 & : & modeh(pred(+nat, +nat)) \\
m3 & : & modeb(zero(+nat)) \\
m4 & : & modeb(even(-nat)) \\
m5 & : & modeb(pred(-nat, +nat)) \\
m6 & : & modeb(succ(+nat, -nat)) \\
m7 & : & modeb(succ(-nat, +nat))
\end{eqnarray*}
The first two declarations $m1$ and $m2$ specify the form of the head of the rule: it must be either $even(X)$ or $pred(X, Y)$.
The $modeb$ declarations specify the sorts of atoms that can appear in the body of the rule.

Let us compare this with the program templates in Section \ref{subsubsec:even-succ2}:
\begin{eqnarray*}
\tau_{target, 1} & = & (h=target, n_\exists=0, int=False) \\
\tau_{target, 2} & = & (h=target, n_\exists=1, int=True) \\
\tau_{pred, 1} & = & (h=pred, n_\exists=1, int=False)
\end{eqnarray*}
In comparison with \system{}'s rule templates, the mode declarations are rather restrictive.
While our rule templates only specify the number of existentially quantified variables, and the \emph{type} of predicate (intensional or extensional) that can appear in the body of the rule, the mode declarations specify precisely the \emph{exact predicates} allowed in the body of the rule.
Rule $m5$ is particularly restrictive:
\[
m5 : modeb(pred(-nat, +nat))
\]
Not only do they specify the precise predicate to be used in the atom (it must be $pred$), but the declaration also specifies exactly where the new existentially quantified variable may appear in the atom: it must be in the first argument of \verb|pred|.

\section{Comparing Language Bias in $\partial$ILP and in Metagol}
\label{metagol_comparison}

Recall when learning $even$ in Section \ref{subsubsec:even-succ2} above, we used the following program template:
\begin{eqnarray*}
\tau_{target, 1} & = & (h=target, n_\exists=0, int=False) \\
\tau_{target, 2} & = & (h=target, n_\exists=1, int=True) \\
\tau_{pred, 1} & = & (h=pred, n_\exists=1, int=False)
\end{eqnarray*}
Here, we compare the above program template with the corresponding metarules needed for Metagol to learn the same task:
\begin{verbatim}
metarule(base,[P,Q],([P,A]:-[[Q,A]])).
metarule(recursive,[P,Q],([P,A]:-[[Q,B,A],[P,B]])).
metarule(invented, [P,Q], ([P,A,B]:-[[Q,A,C],[Q,C,B]])).
\end{verbatim}
Comparing the two program templates, \system{} specifies less information about the terms used in the rules: the Metagol metarules specify the \emph{exact variable bindings} that must occur, while the \system{} template only specifies the number of existential variables.
The \system{} template also specifies less information about the predicates used in the rules. 
The Metagol template specifies the exact form of recursion to appear in the rule:
\begin{verbatim}
metarule(recursive,[P,Q],([P,A]:-[[Q,B,A],[P,B]])).
\end{verbatim}
This is more restrictive than the \system{} template, which simply says that at least one intensional predicate is allowed in the body of the rule.

\section{Space Requirements}

\label{space-requirements}

\system{} is undoubtedly a memory intensive solution to logic program induction.

One major source of memory consumption is the collection of vectors $\mathbf{a}_t^{p,j,k}$ and $\mathbf{c}_t^{p,j,k}$.
Together, these require the following number of floats:
\[
2 \cdot n \cdot t \cdot \sum_{i=1}^{|P_i|} |cl(\tau_i^1)| \cdot |cl(\tau_i^2)|
\]
where $P_i$ is the set of intensional predicates, $t$ is the number of time steps of forward inference, and $cl(\tau)$ is the clauses generated for rule template $\tau$.

Here, $n$ is the size of the Herbrand base $G$:
\[
n = |G| \leq |P| \cdot |C|^2 + 1
\]
where $P$ and $C$ are the sets of predicates and constants.

The other major source of memory consumption is the calculation of the intermediate vectors $\mathbf{Y}_1$, $\mathbf{Y}_2$, and $\mathbf{Z}$.
These require the following number of floats:
\[
3 \cdot n \cdot |C| \cdot \sum_{i=1}^{|P_i|} |cl(\tau_i^1)| \cdot |cl(\tau_i^2)|
\]

\section{Why Not Use a Simpler, Smaller Model?}

\label{simpler-smaller-model}

A natural question to ask about the model presented here is: why make everything so complicated?
Instead of having a matrix of weights $\mathbf{W}_p \in \mathbb{R}^{|cl(\tau_p^1)| \times |cl(\tau_p^2)|}$ for every \emph{pair} of rule templates $\tau_p^1, \tau_p^2$, why not have a vector of weights $\mathbf{W}_\tau \in \mathbb{R}^{|cl(\tau)|}$ for each individual rule template $\tau$?
Let $\mathcal{F}_\tau^j$ be the valuation function for the $j'th$ clause of $cl(\tau)$.
Let $c_t^{\tau,j}$ be the result of applying $\mathcal{F}_\tau^j$ to $\mathbf{a}_t$:
\[
c_t^{\tau,j} = \mathcal{F}_\tau^j(\mathbf{a}_t)
\]
Now we generate weighted sums $\mathbf{b}_t^\tau$ for each time-step $t$ and template $\tau$ in the program template $\Pi$:
\[
\mathbf{b}_t^{\tau} = \sum_{j} \mathbf{c}_t^{\tau, j} \cdot \frac{e^{\mathbf{W}_p[j]}}{\sum_{j'} \mathbf{W}_p[j']}
\]
Finally, define $\mathbf{a}_{t+1}$ in terms of $\mathbf{a}_t$ and the $\mathbf{b}_t^\tau$'s:
\[
\mathbf{a}_{t+1} = f_{amalgamate}(\mathbf{a}_t, \max_{\tau \in \Pi} \mathbf{b}_t^{\tau})
\]
In the memory-intensive \system{} model, the $\mathbf{b}_t^{p}$ were disjoint for different predicates $p$.
But now, in this simpler model, some of the $\mathbf{b}_t^{\tau}$ may produce conclusions involving the same predicate head.
The various $\mathbf{b}_t^{\tau}$ are no longer disjoint.
So we can no longer add them together.
Instead, we must take the elementwise-max.

This simpler model was, in fact, the first model we tried. 
We only moved away from it, with reluctance, when we found it incapable of escaping local minima on harder tasks.
The problem with this simpler, smaller model is that if there are two rule templates both defining the same intensional predicate, then the conclusions of the two rules will over-write each other when they are combined together.
In the memory-intensive \system{} model, the results of applying the clause-pairs is \emph{disjoint} from the results of applying clause-pairs whose head is a different predicate, so the results can be simply \emph{added}.
This means no information is lost when amalgamating the results from different clauses. 
In the simpler model, information is lost when taking the element-wise max, impacting the system's ability to learn programs involving multiple clauses.

\section{The 20 Benchmark ILP Experiments}
\label{all_experiments}

We describe our twenty benchmark experiments in detail. 

\subsection{Predecessor}

The aim of this task is to learn the \emph{predecessor} relation from examples.
The language contains the monadic predicate $zero$ and the successor relation $succ$.
The background knowledge is the set of basic arithmetic facts defining the $zero$ predicate and $succ$ relation:
\[
\mathcal{B} = \{zero(0), succ(0, 1), succ(1, 2), succ(2, 3), ...\}
\]
The positive examples $\mathcal{P}$ are:
\begin{eqnarray*}
\mathcal{P} = \{target(1, 0), target(2, 1), target(3, 2), ...
\}
\end{eqnarray*}
In all these examples, $target$ is the name of the target predicate we are trying to learn. In this case, $target = predecessor$.
The negative examples are 
\[
\mathcal{N} = \{ target(X,Y) \; | \; (X,Y) \in \{0, ..., 9\}^2 \} - \mathcal{P}
\]
The solution found by \system{} is:
\begin{eqnarray*}
target(X, Y) & \leftarrow succ(Y, X) 
\end{eqnarray*}

\subsection{Even / Odd}
\label{even-odd} 

The aim of this task is to learn the \emph{even} predicate from examples.
Again, the language contains the monadic predicate $zero$ and the successor relation $succ$, and the same background facts as in the previous example above.
This task requires an additional auxiliary monadic predicate $pred1$.
The positive examples are:
\begin{eqnarray*}
\mathcal{P} = \{ target(0), target(2), target(4), ... \}
\end{eqnarray*}
Negative examples are:
\begin{eqnarray*}
\mathcal{N} = \{ target(1), target(3), target(5), ... \}
\end{eqnarray*}
One solution found by \system{} is:
\begin{eqnarray*}
target(X) & \leftarrow & zero(X) \\ 
target(X) & \leftarrow & pred1(Y), succ(Y, X) \\ 
pred1(X) & \leftarrow & target(Y), succ(Y, X)
\end{eqnarray*}
Here, $target$ and $pred1$ are mutually recursive.
Note that the invented predicate $pred1(X)$ is true if $X$ is odd.

\subsection{Even / Succ2}

This example is described in Section \ref{subsubsec:even-succ2} above.

\subsection{Less-Than}

The aim of this task is to learn the \emph{less than} relation.
The language and background axioms are the same as the previous examples.
All integers are between 0 and 9.
Positive examples of the $target$ predicate are:
\[
\mathcal{P} = \{target(X,Y) \; | \; X < Y \}
\]
Negative examples are:
\[
\mathcal{N} = \{target(X,Y) \; | \; X \geq Y \}
\]
One solution found by \system{} is:
\begin{eqnarray*}
target(X, Y) & \leftarrow & succ(X, Y) \\
target(X, Y) & \leftarrow & target(X, Z), target(Z, Y) \\ 
\end{eqnarray*}

\subsection{Fizz}

This experiment is described in Section~\ref{subsubsec:fizz-buzz}.

\subsection{Buzz}

This experiment is described in Section~\ref{subsubsec:fizz-buzz}.

\subsection{Member}

The task is to learn the $member$ relation on lists, where $member(X,Y)$ if $X$ is an element in list $Y$.
The elements in a list are encoded using two relations:
\begin{itemize}
\item
$cons(X, Y)$ if the node after $X$ is node $Y$; we terminate lists with the null node $0$
\item
$value(X, Y)$ if the value of node $X$ is $Y$
\end{itemize}
We train on two separate $(\mathcal{B}, \mathcal{P}, \mathcal{N})$ triples. 
In the first, the list is $[4,3,2,1]$ and the positive examples are:
\begin{center}
\begin{tabular}{ll}
$target(4, [4,3,2,1])$ & $target(3, [4,3,2,1])$ \\
$target(2, [4,3,2,1])$ & $target(1, [4,3,2,1])$ \\
$target(3, [3,2,1])$ & $target(2, [3,2,1])$ \\
... & ... \\
\end{tabular}
\end{center}
The negative examples are all other ground atoms involving $target$.
In the second $(\mathcal{B}, \mathcal{P}, \mathcal{N})$ triple, the list is $[2,3,2,4]$ and the positive examples are:
\begin{center}
\begin{tabular}{ll}
$target(2, [2,3,2,4])$ & $target(3, [2,3,2,4])$ \\
$target(4, [2,3,2,4])$ & $target(3, [3,2,4])$ \\
$target(2, [3,2,4])$ & $target(4, [3,2,4])$ \\
... & ... \\
\end{tabular}
\end{center}
One solution found by \system{} is:
\begin{eqnarray*}
member(X, Y) & \leftarrow & value(Y, X) \\
member(X, Y) & \leftarrow & cons(Y, Z), member(X, Z)
\end{eqnarray*}

\subsection{Length}

The task is to learn the \emph{length} relation, where $length(X,Y)$ is true if the length of list $X$ is $Y$.
The representation of lists is the same as in the experiment above.
To solve this task, we need at least one auxiliary relation $pred1$.

Again, we train on two separate $(\mathcal{B}, \mathcal{P}, \mathcal{N})$ triples. 
In the first, the list is $[1,2,3]$ and the positive examples are:
\begin{eqnarray*}
& & target([1,2,3], 3]) \\
& & target([2,3], 2) \\
& & target([3], 1) \\
\end{eqnarray*}
The negative examples are all ground atoms involving $target$ except those in the positive set.

In the second triple, the list is $[3,2,1]$ and the positive examples are:
\begin{eqnarray*}
& & target([3,2,1], 3]) \\
& & target([2,1], 2) \\
& & target([1], 1) \\
\end{eqnarray*}
One of the solutions found by \system{} is:
\begin{eqnarray*}
target(X, X) & \leftarrow & zero(X) \\
target(X, Y) & \leftarrow & cons(X, Z), pred1(Z,Y) \\
pred1(X, Y) & \leftarrow & target(X, Z), succ(Z,Y)
\end{eqnarray*}

\subsection{Son}

The task here is to learn the \emph{son-of} relation from various facts about a family tree involving the relations \emph{father-of}, \emph{brother-of}, and \emph{sister-of}.
The task requires at least one auxiliary monadic predicate $pred1$.
There are a set of constants representing people: $a, b, c, ...$
The background facts are 
\begin{eqnarray*}
\{father(a, b), father(a, c), father(d, e), father(d, f), father(g, h), father(g, i), \\
brother(b, c), brother(c, b), brother(e, f), sister(f, e), sister(h, i), sister(i, h)\}
\end{eqnarray*}
Note that the facts about \emph{sister-of} are irrelevant to this problem.
The positive examples are:
\begin{eqnarray*}
\mathcal{P} = \{target(b, a), target(c, a), target(e, d)\}
\end{eqnarray*}
The negative examples are a subset of all ground atoms involving the target predicate that are not in $\mathcal{P}$.

One of the solutions found by \system{} is:
\begin{eqnarray*}
target(X, Y) & \leftarrow & father(Y, X), pred1(X) \\
pred1(X) & \leftarrow & brother(X, Y) \\
pred1(X) & \leftarrow & father(X, Y)
\end{eqnarray*}
Here, the invented predicate $pred1$ here defines the \emph{is-male} property: $X$ is male if $X$ is the brother of someone, or if $X$ is the father of someone. The clause for $target$ reads: $X$ is the son of $Y$ if $Y$ is the father of $X$ and $X$ is male. 

\subsection{Grandparent}

The task here is to learn the \emph{grandparent} relation from various facts involving the father-of and mother-of relations.
The problem uses an auxiliary relation $pred1$.
There are a set of constants representing people: $a, b, c, ...$
The background facts are:
\begin{eqnarray*}
&& \{mother(i, a), father(a, b), father(a, c), father(b, d), father(b, e), \\
&& mother(c, f), mother(c, g), mother(f, h)\}
\end{eqnarray*}
Positive examples are:
\begin{eqnarray*}
\mathcal{P} = \{target(i, b), target(i, c), target(a, d), target(a, e), target(a, f), target(a, g), target(c, h)\}
\end{eqnarray*}
Again, the negative examples are all ground atoms involving $target$ except those in the positive set. 

One solution found by \system{} is:
\begin{eqnarray*}
target(X,Y) & \leftarrow & pred1(X, Z), pred1(Z, Y) \\
pred1(X, Y) & \leftarrow & father(X, Y) \\
pred1(X, Y) & \leftarrow & mother(X, Y)
\end{eqnarray*}
Note that the invented predicate $pred1$ here defines the \emph{parent-of} relation. 
This is the same solution found in \citeA{muggleton2014meta}.

\subsection{Husband}

The task here is to learn the \emph{husband-of} relation from various facts about family relations involving the relations \emph{father-of}, \emph{daughter-of}, and \emph{brother-of}.
The atomic facts were taken from the European family-tree dataset \cite{wang2015soft}.

One of the solutions found by \system{} is:
\begin{eqnarray*}
target(X,Y) & \leftarrow & father(X,Z), daughter(Z, Y)
\end{eqnarray*}

\subsection{Uncle}

The task here is to learn the \emph{uncle-of} relation from various facts about family relations including \emph{father-of}, \emph{mother-of}, and \emph{brother-of}.
Again, the background facts were taken from the European family-tree dataset \cite{wang2015soft}.
The task requires one auxiliary relation $pred1$.

One of the solutions found by \system{} is:
\begin{eqnarray*}
target(X,Y) & \leftarrow & brother(X,Z), pred1(Z, Y) \\
pred1(X,Y) & \leftarrow & father(X, Y) \\
pred1(X,Y) & \leftarrow & mother(X, Y)
\end{eqnarray*}
The invented predicate $pred1$ here defines the \emph{is parent of} relation.

\subsection{Relatedness}

The task here is to learn the \emph{related} relation from facts about family relations involving the \emph{parent-of} relation.
The background facts are
\begin{eqnarray*}
parent(a, b) & parent(a, c) \\
parent(c, e) & parent(c, f) \\ 
parent(d, c) & parent(g, h)
\end{eqnarray*}
The idea is that $target(X,Y)$ is true if there is some (undirected) path between $X$ and $Y$. 
Positive examples are:
\begin{eqnarray*}
\mathcal{P} = \{target(a, b), target(a, c), target(a, e), target(a, f), target(f, a), target(a, a), target(d, b), target(h, g)\}
\end{eqnarray*}
Negative examples are:
\begin{eqnarray*}
\mathcal{N} = \{target(g, a), target(a, h), target(e, g), target(g, b)\}
\end{eqnarray*}
One of the solutions found by \system{} is:
\begin{eqnarray*}
target(X, Y) & \leftarrow & pred1(X, Y) \\
target(X, Y) & \leftarrow & pred1(X, Z), target(Z, Y) \\
pred1(X, Y)  & \leftarrow & parent(X, Y) \\
pred1(X, Y)  & \leftarrow & parent(Y, X)
\end{eqnarray*}

\subsection{Father}
\label{subsubsec:father}

\begin{figure}
\centering
\includegraphics[scale=0.1]{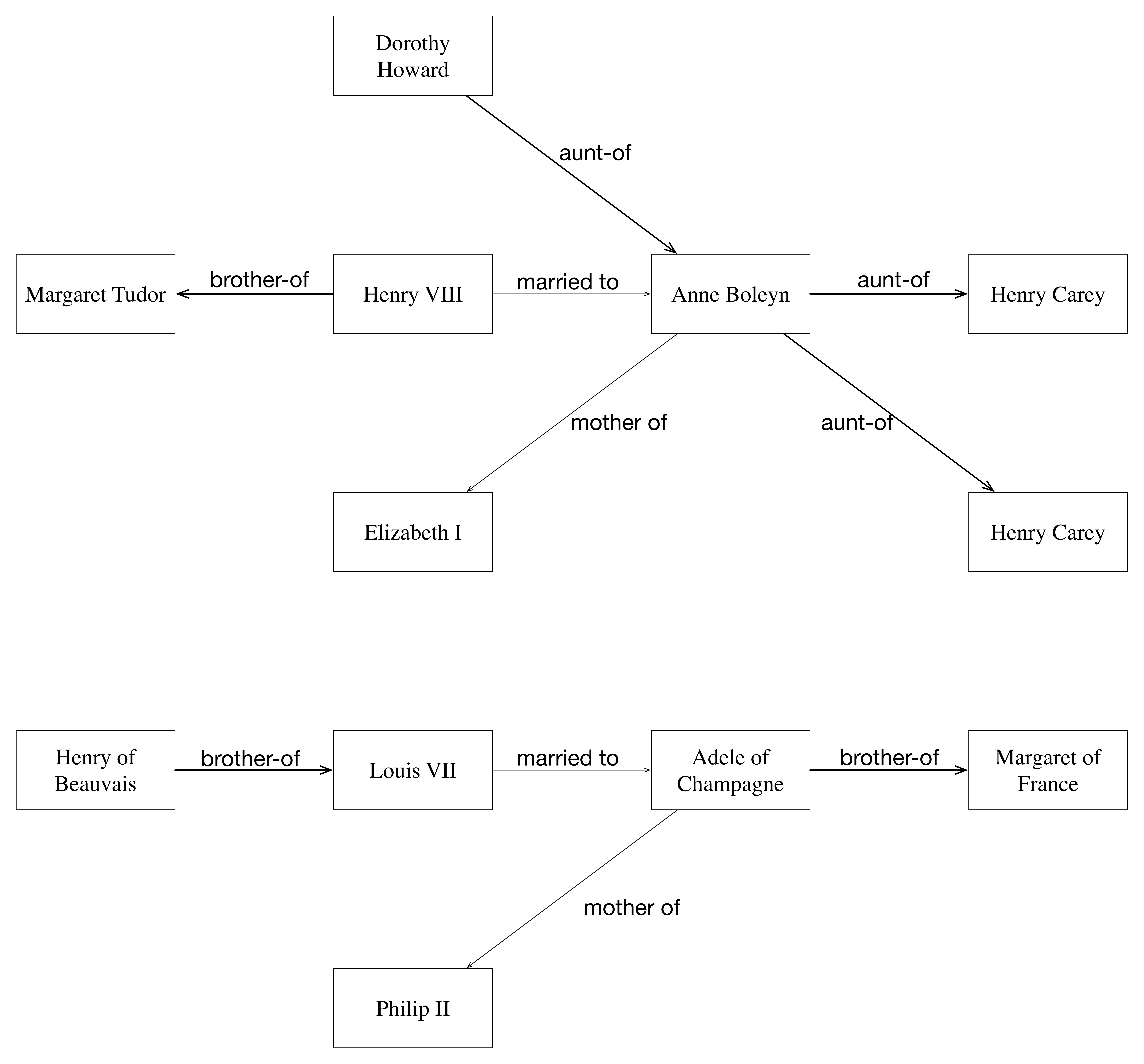}
\caption{The Background $\mathcal{B}$ for the task in Section \ref{subsubsec:father}}
\label{father-training}
\end{figure}

Figure \ref{father-training} shows part of a family tree centered on Louis VII and Henry VIII.
The task is to learn the \emph{father-of}relation given this background data and just two positive examples:
\begin{itemize}
\item Louis VII is the \emph{father of} Phillip II
\item Henry VIII is the \emph{father of} Elizabeth I
\end{itemize}
All other ground atoms involving the \emph{father-of} relation are assumed to be false and are added to the negative examples.

Note that some of the background data is \emph{irrelevant} to the task at hand: we could solve this task without any data about brothers or aunts.
\system{} has to sift through the background data, finding a rule that entails the positive examples without entailing any of the negative examples.

Formally, the language L for this task is:
\begin{itemize}
\item
$P_e: \{husband/2, mother/2, brother/2, aunt/2\}$
\item
$P_i: \{target/2\}$
\end{itemize}
The set $C$ of constants contains Louis VII, Henry VIII, etc.
Here, $target$ is the \emph{father-of} relation we are trying to learn.

For validation, we give the system fictitious family tree data:
\begin{itemize}
\item
George Donatus is married to Cecilie of Mountbatten
\item
Cecilie of Mountbatten is the mother of Louis
\end{itemize}
The positive example for validation is the single atom stating that George Donatus is the \emph{father of} Louis.
The negative examples are all other ground atoms whose predicate is $target$.

In this simple task, \system{} always finds the correct solution.
It finds the program:
\[
target(X, Y) \leftarrow husband(X, Z), mother(Z, Y)
\]
In other words, if $X$ is the husband of some $Z$, and $Z$ is the mother of $Y$, then $X$ is the father of $Y$.

\subsection{Undirected Edge}

A graph is represented by a set of $edge(X,Y)$ atoms, representing that there is an edge from node $X$ to node $Y$.
The task is to learn the \emph{unconnected-edge} relation, which is true of $X$ and $Y$ if there is an edge, in either direction, between $X$ and $Y$.

We train on two separate $(\mathcal{B}, \mathcal{P}, \mathcal{N})$ triples. 
In the first, the background facts are:
\begin{eqnarray*}
\{ 
edge(a, b), edge(b, d), edge(c, c)
\}
\end{eqnarray*}
The positive examples are:
\begin{eqnarray*}
\mathcal{P} = \{
target(a, b), target(b, a), target(b, d), target(d, b), target(c, c)
\}
\end{eqnarray*}
In the second triple, the background facts are:
\begin{eqnarray*}
\{
edge(a, b), edge(c, d)
\}
\end{eqnarray*}
The positive examples are:
\begin{eqnarray*}
\mathcal{P} = \{
target(a, b), target(b, a), target(c, d), target(d, c)
\}
\end{eqnarray*}
The solution found by \system{} is:
\begin{eqnarray*}
target(X, Y) & \leftarrow & edge(X, Y) \\ 
target(X, Y) & \leftarrow & edge(Y, X)
\end{eqnarray*}

\subsection{Adjacent to Red}
In this example, the nodes of the graph are coloured either green or red.
The task is to learn the predicate \emph{is adjacent to a red node}.
As well as the $edge$ relation, we have a $colour$ relation on nodes, where $colour(X, C)$ means node $X$ has colour $C$. 
We also have two monadic predicates $red(C)$ and $green(C)$, true of colour $C$ if it is red or green respectively.
The task requires at least one auxiliary monadic predicate $pred1$.

We train on two separate $(\mathcal{B}, \mathcal{P}, \mathcal{N})$ triples. 
In the first, the background facts are:
\begin{eqnarray*}
&& \{ edge(a, b), edge (b, a), edge(c, d), edge(c, e),
edge(d, e), \\
&& colour(a, red), colour(b, green),
colour(c, red), colour(d, red), colour(e, green) \}
\end{eqnarray*}
The positive examples are:
\begin{eqnarray*}
\mathcal{P} = \{
target(b), target(c)
\}
\end{eqnarray*}
In the second triple, the background facts are:
\begin{eqnarray*}
& & \{ edge(b, c), edge(d, c), colour(a, red), colour(b, green), \\ 
& & colour(c, red), colour(d, red), colour(e, green)\}
\end{eqnarray*}
The positive examples are:
\begin{eqnarray*}
\mathcal{P} = \{
target(b, b), target(d, d)
\}
\end{eqnarray*}
One solution found by \system{} is:
\begin{eqnarray*}
target(X) & \leftarrow & edge(X, Y), pred1(Y) \\
pred1(X) & \leftarrow & colour(X, Y), red(Y)
\end{eqnarray*}

\subsection{Two Children}

The task is to learn the predicate \emph{has at least two children}.
As well as the $edge$ predicate, this task uses the \emph{not-equals} relation $neq$.
In future work, we plan to add negation-as-failure to the language, at which point it will not be necessary to include the $neq$ relation explicitly.
This task requires an auxiliary relation $pred1$.
The background facts are 
\[
\mathcal{B} = \{neq(X, Y) \; | \; X \neq Y \}
\]
We train on two separate $(\mathcal{B}, \mathcal{P}, \mathcal{N})$ triples. 
In the first, the background facts are:
\begin{eqnarray*}
\{
edge(a, b), edge(a, c), edge(b, d), edge(c, d),
edge(c, e), edge(d, e)
\}
\end{eqnarray*}
The positive examples are:
\begin{eqnarray*}
\mathcal{P} = \{
target(a, a), target(c, c)
\}
\end{eqnarray*}
In the second triple, the background facts are:
\begin{eqnarray*}
\{
edge(a, b), edge(b, c), edge(b, d), edge(c, c),
edge(c, e), edge(d, e)
\}
\end{eqnarray*}
The positive examples are:
\begin{eqnarray*}
\mathcal{P} = \{
target(b, b), target(c, c)
\}
\end{eqnarray*}
One solution found by \system{} is:
\begin{eqnarray*}
target(X) & \leftarrow & edge(X, Y), pred1(X, Y) \\
pred1(X, Y) & \leftarrow & edge(X, Z), neq(Z, Y)
\end{eqnarray*}

\subsection{Graph Colouring}

A graph is well-coloured if no two adjacent nodes are the same colour.
The task here is to learn the \emph{is-bad-node} predicate that is true of a node $X$ if $X$ is adjacent to a node of the same colour.
As well as the $edge$ relation, we have a $colour$ relation on nodes, where $colour(X, C)$ means node $X$ has colour $C$. 
The task requires at least one auxiliary relation $pred1$.

We train on two separate $(\mathcal{B}, \mathcal{P}, \mathcal{N})$ triples. 
In the first, the background facts are:
\begin{eqnarray*}
&& \{ edge(a, b), edge(b, c), edge(b, d), edge(c, e), edge(e, f), colour(a, green), \\
& & colour(b, red), colour(c, green), colour(d, green), colour(e, red), colour(f, red)\}
\end{eqnarray*}
The positive example is:
\begin{eqnarray*}
\mathcal{P} = \{
target(e, e)
\}
\end{eqnarray*}
In the second triple, the background facts are:
\begin{eqnarray*}
&& \{edge(a, b), edge(b, a), edge(c, b), edge(f, c), edge(d, c), edge(e, d), colour(a, green), \\
&& colour(b, green), colour(c, red), colour(d, green), colour(e, green), colour(f, red)
\}
\end{eqnarray*}
The positive examples are:
\begin{eqnarray*}
\mathcal{P} = \{
target(a, a), target(b, b), target(e, e), target(f, f)
\}
\end{eqnarray*}
One solution found by \system{} is:
\begin{eqnarray*}
target(X) & \leftarrow & edge(X, Y), pred1(X, Y) \\
pred1(X, Y) & \leftarrow & colour(X, Z), colour(Y, Z)
\end{eqnarray*}
Here, the invented $pred1$ relation is true of $X$ and $Y$ if $X$ and $Y$ are both the same colour.

\subsection{Graph Connectedness}

In this graph task, the problem is to learn the $connected(X,Y)$ relation that is true if there is some sequence of edge transitions connecting $X$ and $Y$.
The background facts are:
\begin{eqnarray*}
\{
edge(a, b), edge(b, c), edge(c, d), edge(b, a)
\}
\end{eqnarray*}
The positive examples are:
\begin{eqnarray*}
\mathcal{P} = 
& & \{ target(a, b), target(b, c), target(c, d),
target(b, a), target(a, c), \\
& & target(a, d),
target(a, a), target(b, d), target(b, a),
target(b, b)
\}
\end{eqnarray*}
The solution \system{} finds is:
\begin{eqnarray*}
target(X, Y) & \leftarrow & edge(X, Y) \\ 
target(X, Y) & \leftarrow & edge(X, Z), target(Z, Y)
\end{eqnarray*}
This is the transitive closure of the $edge$ relation.
This is the same relation that is synthesised as the recursive auxiliary predicate needed for the Graph Cycles task described in Section~\ref{subsubsec:cycle}.

\subsection{Graph Cycles}

This experiment is described in detail in Section~\ref{subsubsec:cycle}.

\bibliography{nilp}
\bibliographystyle{theapa}

\end{document}